\documentclass{article}
\usepackage{arxiv}
\usepackage{authblk}
\usepackage[utf8]{inputenc} % allow utf-8 input
\usepackage[T1]{fontenc}    % use 8-bit T1 fonts
\usepackage{hyperref}       % hyperlinks
\usepackage{url}            % simple URL typesetting
\usepackage{booktabs}       % professional-quality tables
\usepackage{amsmath}
\usepackage{amsfonts}       % blackboard math symbols
\usepackage{nicefrac}       % compact symbols for 1/2, etc.
\usepackage{microtype}      % microtypography
\usepackage{amssymb}
\usepackage{systeme}
\usepackage{bbm}
\usepackage{stmaryrd}
\newcommand{\ssymbol}[1]{^{\@fnsymbol{#1}}}
\usepackage{bm}
\usepackage{graphicx}
\usepackage{natbib}
\usepackage{doi}
\usepackage{subfig}
\usepackage{float}
\usepackage{multirow}
\usepackage{algorithm}
\usepackage{algorithmic}
\floatname{algorithm}{Procedure}
\usepackage{booktabs}
\usepackage{xcolor}

\usepackage{alphalph}
\makeatletter
\renewcommand\AB@authnote[1]{\textsuperscript{\normalfont\bfseries#1}}
\makeatother

\title{Geometry-aware framework for deep energy method: an application to structural mechanics with hyperelastic materials}

\author[1,2,3]{Thi Nguyen Khoa Nguyen}
\author[1,2]{Thibault Dairay}
\author[2]{Raphaël Meunier}
\author[2]{Jean Di Stasio}
\author[1,3]{Christophe Millet}
\author[1,4]{Mathilde Mougeot}
\affil[1]{Universite Paris-Saclay, ENS Paris-Saclay, CNRS, Centre Borelli, Gif-sur-Yvette, 91190 France}
\affil[2]{Michelin, Centre de Recherche de Ladoux, Cébazat, 63118 France}
\affil[3]{CEA, DAM, DIF, F-91297 Arpajon, France}
\affil[4]{ENSIIE, Évry-Courcouronnes, 91000 France}
\begin{document}
\maketitle

\begin{abstract}
Physics-Informed Neural Networks (PINNs) have gained considerable interest in diverse engineering domains thanks to their capacity to integrate physical laws into deep learning models. Recently, geometry-aware PINN-based approaches that employ the strong form of underlying physical system equations have been developed with the aim of integrating geometric information into PINNs. Despite ongoing research, the assessment of PINNs in problems with various geometries remains an active area of investigation. In this work, we introduce a novel physics-informed framework named the Geometry-Aware Deep Energy Method (GADEM) for solving structural mechanics problems on different geometries. As the weak form of the physical system equation (or the energy-based approach) has demonstrated clear advantages compared to the strong form for solving solid mechanics problems, GADEM employs the weak form and aims to infer the solution on multiple shapes of geometries. Integrating a geometry-aware framework into an energy-based method results in an effective physics-informed deep learning model in terms of accuracy and computational cost. Different ways to represent the geometric information and to encode the geometric latent vectors are investigated in this work. We introduce a loss function of GADEM which is minimized based on the potential energy of all considered geometries. An adaptive learning method is also employed for the sampling of collocation points to enhance the performance of GADEM. We present some applications of GADEM to solve solid mechanics problems, including a loading simulation of a toy tire involving contact mechanics and large deformation hyperelasticity. The numerical results of this work demonstrate the remarkable capability of GADEM to infer the solution on various and new shapes of geometries using only one trained model.
\end{abstract}

% \end{frontmatter}

% \linenumbers

\section{Introduction}
Partial differential equations (PDEs) are widely used to model and analyze physical problems in engineering. Due to the difficulty of obtaining analytical solutions, various numerical methods for solving these equations have been proposed and widely used, such as the finite difference method, finite element method, and finite volume method \citep{quarteroni2008numerical, ames2014numerical}. However, these schemes remain expensive in practice due to the large number of degrees of freedom (DOFs) required to accurately solve the PDEs \citep{amsallem2010interpolation}, and a skillful mesh refinement technique is required to enhance the accuracy \citep{de1983posteriori}. In the last few years, advances in deep learning methods have gained much attention in the field of physical modeling. These methods have increasingly been used to study the physics of systems by incorporating physical constraints, which are typically expressed as PDEs, into the models \citep{yu2018deep,raissi2019physics, rackauckas2020universal,samaniego2020energy}. Among these studies, the series of work by Karniadakis’ group on Physics-Informed Neural Networks (PINNs) \citep{raissi2019physics,karniadakis2021physics} stands out as an attractive and remarkable approach to solving forward and inverse PDE problems. PINNs use neural networks as approximators and integrate the physical constraints of the systems into the cost function by minimizing the PDE residuals calculated on a set of collocation points (non-supervised points). Thanks to their simplicity and powerful capability when dealing with problems involving PDE or systems of PDEs, PINNs are gaining more and more attention from researchers in engineering fields. 

Recently, many studies focused on improving the efficiency of PINNs by using adaptive strategies, such as adaptive activation functions \cite{jagtap2020adaptive,jagtap2020locally,jagtap2022deep1}, adaptive weights in the cost function \citep{mcclenny2020self,wang2020understanding, wang2020and,wang2022respecting}, and adaptive collocation points during the training \citep{peng2022rang,daw2022rethinking,wu2023comprehensive,nguyen2023fixed}. Besides that, many extensions of PINNs are gradually being developed and improved, whether to overcome the computational limitation or to deal with spectral bias of the neural networks \citep{kharazmi2019variational,pang2019fpinns,yang2021b,jagtap2021extended,moseley2023finite}. Besides that, the applicability of PINNs has been demonstrated in a wide range of research areas. For example,  \cite{chen2020physics} applied PINNs for inverse problems in nano-optics and metamaterials, \cite{kissas2020machine,arzani2021uncovering} employed PINNs for the modeling of cardiovascular flows and blood flow, \cite{strelow2023physics} applied PINNs to gas transport problems, \cite{shukla2021physics} used PINNs to identify polycrystalline nickel material coefficients, \cite{nguyen2022physics} applied PINNs in a non-Newtonian fluid thermo-mechanical problem which is used in the rubber calendering process, \cite{chen2023tgm} employed a physics informed framework to forecast a tumor growth. 

Current efforts also focus on developing convolutional neural networks (CNN) to efficiently learn large-scale spatiotemporal physical fields \citep{fang2021high,ren2022phycrnet,li2020fourier}. However, these convolutional-based deep learning frameworks require a regular grid for the training points and are not efficient when dealing with irregular geometries. To overcome this challenge, \cite{gao2021phygeonet} introduced a physics-informed architecture using convolutional neural networks, which is trainable on a set of geometries, named PhyGeoNet that uses elliptic mapping to transform an irregular physical domain into a regular reference domain. \cite{li2023fourier} proposed a geometry-aware extension of the Fourier neural operator named Geo-FNO, which deforms the irregular input domain into a uniform latent mesh on which the Fourier transform can be applied. Recently, \cite{serrano2024operator} introduced the coordinate-based model for operator learning framework (CORAL), which is based on implicit neural representations and able to learn mappings between functions sampled on irregular meshes, for solving PDEs on general geometries. 

As far as computational mechanics is concerned, there has been a growing number of studies that focus on the capability of PINNs to solve problems involving solid mechanics. \cite{haghighat2021physics} employed PINNs with the PDE-based approach (strong form of the system) to identify the material parameters in solid mechanics problems. With the same technique, \cite{sahin2023solving} utilized PINNs to solve a contact mechanic in linear elasticity. \cite{samaniego2020energy,nguyen2020deep,nguyen2021parametric} proposed to use the Deep Energy Method (DEM) (a weak form of the system) in the loss function of PINNs to solve linear elasticity and hyperelasticity problems. \cite{fuhg2022mixed} introduced an extension of DEM named mixed Deep Energy Method (mDEM) that enhances the training capability of the network by taking the stress components as additional outputs. However, this approach requires higher computational resources due to additional input and the computation of the first Piola-Kirchhoff stress tensor. \cite{abueidda2023enhanced} introduced a framework that combined the strong form and the weak form of the system, and employed the coefficient of variation weighting scheme to enhance the accuracy of the model. \cite{chadha2022optimizing} used Bayesian optimization algorithms and random search to identify optimal values for hyperparameters when using DEM. \cite{lee2024adversarial} proposed an extension of DEM designed for multiphysics simulation named adversarial Deep Energy Method (adversarial DEM) for solving saddle point problems with electromechanical coupling. \cite{li2021physics2} compared the PDE-based approach and the energy-based approach in the problem of predicting the mechanical responses of elastic plates. Throughout these works, the energy-based approach has been demonstrated to be more efficient than the PDE-based approach for solving solid mechanics problems in terms of accuracy and computational resources, as it only requires first-order automated differentiation. 

% \textcolor{red}{TIBO: si j'étais reviewer, je me demanderais dans la suite pourquoi on n'a pas utilisé mDEM au lieu de DEM. Tu avais testé ? C'était pas mieux ?} \textcolor{blue}{Khoa: Oui j'ai ajouté une phrase pour la raison de ne pas utiliser mDEM, c'est à cause à le temps calcul et les resources computaionelles. Et non, je n'ai pas testé car cette approche car ça marche déjà avec DEM. Pour mDEM, c'est une mixed formulation, i.e. il y a une forte loss et une faible loss. Et personellement, je pense que c'est difficile à ajouter une forte loss car ça ajoute une terme que on ne savait pas comment controller sa poid dans la function court (il faut le temps pour trouver un bon poid je pense). Dans le papier de mDEM, les auteurs l'utilisent parce que DEM marche pas dans leur examples je crois. Ils ne montrent pas une exact gain entre DEM et mDEM dans les cas ces deux marchent.}

However, one drawback of the classical PINNs and DEM is that the model is capable of inferring the solution on only one configuration, that is, one needs to retrain PINNs or DEM in case dealing with a new geometry. To tackle this issue, \cite{kashefi2022physics} proposed a Physics-informed PointNet (PIPN) that used a point-cloud-based neural network to encode the geometric features so that the model can predict the solution on multiple sets of irregular geometries. With the same purpose, \cite{oldenburg2022geometry} introduced geometry-aware PINNs (GAPINNs) that used a Variational Autoencoder (VAE) \citep{kingma2019introduction} to extract geometric latent vector, and then used this latent vector as input of the neural networks. The PhyGeoNet framework of \cite{gao2021phygeonet} is also trainable on a set of geometries and able to infer the solution on new geometries. Recently \cite{cameron2024nonparametric} proposed a neural operator architecture designed to learn PDE solutions based on boundary geometry. This approach does not constrain the geometry to a fixed, finite-dimensional parameterization, and therefore is capable of estimating the PDE solution over a whole new geometry. 
% \textcolor{red}{TIBO: ici à mon avis il faut reformuler le paragraphe. Il ne faut pas mélanger les conditions initiales et aux limites et la géométrie. On peut tout à fait paramétrer PINNs par les conditions aux limites ou initiales en les rajoutant en entrée du domaine. Donc je virerais la partie CL/CI. Ensuite, il faut dire que notre but est d'inférer les solutions sur différentes géométries sans réentrainer. C'est aussi un des objectifs de PhyGeoNet (pas uniquement puisqu'au départ la transformation elliptique sert surtout à passer d'un domaine irrégulier à un domaine régulier sur lequel on peut appliquer un CNN) et peut être aussi d'autres références que tu cites et que je ne connais pas. Attention par contre, à mon avis Geo-FNO n'essaie pas de faire ça, le but est vraiment de mapper une géométrie irrégulière sur une géométrie régulière car ils font du Fourier et ils ont besoin de ça (idem CNN)... mais ils ne peuvent pas à mon avis inférer sur n'importe quelle géométrie sans réentrainer... A vérifier pour chaque article que tu cites: quel est le but, inférer sur plusieurs géométries sans réentrainer ou trouver un espace simple où on peut entrainer le réseau mais pour inférer sur une géométrie complexe donnée ?} \textcolor{blue}{Khoa: Oui tu as raison. Désolée je n'ai pas fait attention à Geo-FNO et CORAL. Je les ai ajouté dans une autre paragraph et corrigé la paragraph avec ce que tu as proposé.}

In this paper, we introduce a novel framework, which 
is an extension of GAPINNs that employs deep energy method in the lost function, named the Geometry-Aware Deep Energy Method (GADEM) to infer the solution of structural mechanical problems using the energy-based approach. In the first training phase, different representation techniques, such as spatial coordinates of the objects' boundaries or images of the objects, are employed to represent the geometries. To encode geometric information, parametric encoding (\textit{i.e.} the parameters that control the shapes of geometries, if available) or dimensionality reduction methods are employed. In particular, we invest and compare a linear dimensionality reduction method (Principal Component Analysis \citep{jolliffe2002principal}), and a non-linear method (Variational Autoencoder) \citep{kingma2019introduction}, to encode the geometric information into latent vectors. In the second phase, the model uses the geometric latent vectors as additional input for the networks. The Deep Energy Method is utilized during the second phase and we minimize the loss (which is the potential energy of the system) over all the training geometries.  We also employ an original adaptive strategy \cite{nguyen2023fixed} to infer the best location for the training points to enhance the performance of GADEM. We first validate our proposed method on an academic test case with various shapes of geometries and the reference solutions are obtained using the Finite Element Method (FEM). We then consider the problem of contact between a deformable object and a rigid obstacle, which involves additional inequality constraints in the loss function. This application consists of a simplified tire loading simulation that involves different irregular geometries. To the best of our knowledge, this is the first time a physics-informed deep learning framework has been successfully applied to solve a contact mechanics problem involving hyperelastic materials on different geometries. All our results are compared to a reference solution obtained with a finite element solver. The code used in this study is available on the GitHub repository \url{https://github.com/nguyenkhoa0209/gadem} upon publication.
% \textcolor{red}{TIBO: il faudrait quand même à un moment dire que GADEM est une extension de GAPINNs qui consiste à coupler GAPINNs et DEM, non ?} \textcolor{blue}{Khoa: Oui j'ai ajouté ça dans la phrase.}

The following of this paper is organized as follows. In section 2, the mathematical formulation of contact mechanics for hyperelastic materials is introduced. In section 3, the framework of vanilla PINNs and the Deep Energy Method for solving forward problems in contact mechanics are presented. Later in this section, we introduce the Geometry-Aware Deep Energy Method (GADEM) and its formulation. An adaptive training strategy for the energy-based methods is also briefly presented in this section. In section 4, we validate our framework in an academic test case of linear elasticity. We then investigate the performance of GADEM in the tire loading simulation in section 5. Finally, the conclusions and perspectives are summarized in section 6.  
% \textcolor{red}{TIBO: Il manque pas l'exemple académique là ?}.\textcolor{blue}{Oui :))}
% The complement performance of our framework on academic test cases is provided in the appendix.

\section{Contact mechanics for hyperelastic materials}
In the following, we consider a contact problem between a deformable body made of hyperelastic material and a rigid obstacle. The body is denoted as $\bm{\Omega}$ in its initial configuration (reference configuration before deformation). Assuming the static case, the function that describes the deformation of the body and maps it from the initial configuration into the actual configuration is denoted by $\Phi$. The actual configuration is then denoted by $\Phi(\bm{\Omega})$. For a point of the deformable body, we denote $\bm{X}$ the position vector in the initial configuration, and $\bm{x}$ the position vector in the actual configuration. Then the relation between two positions is given by $\bm{x} = \Phi(\bm{X})=\bm{X} + \bm{u}(\bm{X})$ where $\bm{u}$ denotes the displacement vector.

% \textcolor{green}{Jean: For a point of the deformable body, } \textcolor{blue}{Khoa: oui j'ai ajouté ta phrase}\textcolor{green}{Jean: Tout bon pour moi } 

% \textcolor{green}{Jean: Pour la description des conditions limites de Neumann, je dirais plutôt : the Neumann boundary where a pressure is applied} \textcolor{blue}{Khoa: Oui j'ai le corrigé dans le context}
The boundary of the object at the initial configuration $\partial \bm{\Omega}$ can be represented as $\partial \bm{\Omega} = \partial \bm{\Omega}_D\cup \partial \bm{\Omega}_N \cup \partial \bm{\Omega}_C$, where $\partial \bm{\Omega}_D$ denotes the Dirichlet boundary where displacements are applied, $\partial \bm{\Omega}_N$ denotes the Neumann boundary where the pressure is applied, and $\partial \bm{\Omega}_C$ denotes the boundary where there is potential contact with other rigid obstacle. We consider the boundary value problem with the governing equations as follows:
\begin{align}
    \mathrm{div}\bm{P} + \bm{f}_b &= \bm{0} \text{ on } \bm{\Omega}\label{pde_1} \\
    \bm{u} &= \bm{u}^{*}\text{ on } \partial \bm{\Omega}_D\label{pde_2} \\
    \bm{P} \cdot \bm{N} &= \bm{t}^{*}\text{ on } \partial \bm{\Omega}_N\label{pde_3}
\end{align}
where $\bm{P}$ denotes the first Piola-Kirchhoff stress tensor, $\bm{f}_b$ denotes the density force on the body, $\bm{N}$ denotes the outward normal vector, and $u^*, t^*$ denotes the reference values on Dirichlet boundary and Neumann boundary, respectively. The deformation gradient tensor which measures the deformation of the body $\bm{\Omega}$ is defined as:
\begin{align*}
    \bm{F} = \mathbf{\nabla}_{\bm{X}} \Phi(\bm{X}) = \mathbf{I} +  \mathbf{\nabla}_{\bm{X}} \bm{u}
\end{align*}
% \textcolor{green}{Jean: ici $\bm{f}_b$ est une densité d'effort, une force divisée par un volume} \textcolor{blue}{Khoa: ah c'est pas 'body force' en anglais? Sinon, pourrais-tu le corriger directment s'il te plait?}\textcolor{green}{Jean: j'ai corrigé}
To describe the behavior of hyperelastic materials, the strain energy function $\Psi$ can be used to show a relation between the stress tensors and the displacement field. The first Piola-Kirchhoff stress tensor can be given as follows:
\begin{align*}
    \bm{P} = \dfrac{\partial \Psi}{\partial \bm{F}}
\end{align*}
In this work, we focus on the Saint-Venant Kirchhoff hyperelastic model, a generalization of linear elasticity. The strain energy in this model is given by:
\begin{align*}
    \Psi = \lambda tr^2(\bm{E})\mathbf{I} + \mu tr(\bm{E}^2)
\end{align*}
where $\lambda$ and $\mu$ are the Lamé constants, and $\bm{E} = \dfrac{1}{2}(\bm{F}^T\bm{F} - \mathbf{I})$ is Green-Lagrange tensor. 

The potential energy of the system, which is at minimum in the equilibrium state, is given by:
\begin{align}
    \Pi(\bm\phi) =  \int_{\bm{\Omega}}\Psi dV -\Big(\int_{\bm{\Omega}} \bm{f}_b.\bm \phi dV +\int_{\bm{\Omega}_N} \bm t^*.\bm \phi dA \Big)\label{weak_form}
\end{align}
where $\bm\phi \in \mathcal{H}$ is a trial function, $\mathcal{H}$ denotes the space of trial functions (admissible functions) \citep{fortin2011elements}. In (\ref{weak_form}), The term $\int_{\bm{\Omega}}\Psi dV$ represents the internal energy, and the term $\Big(\int_{\bm{\Omega}} \bm{f}_b.\bm \phi dV +\int_{\bm{\Omega}_N} \bm t^*.\bm \phi dA \Big)$ represents the external energy.
% \textcolor{green}{Jean: cette équation n'est pas "habituelle", peut-être ajouter une citation pour indiquer d'où elle provient ?} \textcolor{blue}{oui j'ai cité une référence}

Solving the system of PDEs (\ref{pde_1}),(\ref{pde_2}),(\ref{pde_3}) is referred to as solving the strong form of the system, while minimizing the potential energy (\ref{weak_form}) is referred to as solving the weak form of the system. The weak form has a clear advantage compared to the strong form in that it only requires first-order differentiation, and thus reduces a huge amount of computational resources. In this work, we focus on solving the mechanical problems using the weak formulation. The calculation of the integral is discussed in more detail in section \ref{DEM_section}. 

Next, we present the additional conditions when there is frictionless contact between the body $\bm \Omega$ and a rigid obstacle. On the potential contact boundary $\bm \Omega_C$, the three conditions representing unilateral contact (also known as Signorini conditions) are given by:
\begin{align}
    &u_n\le g\label{contact_1}\\ &P_n \le 0\label{contact_2} \\ &P_n (u_n-g) =0\label{contact_3}
\end{align}
The initial gap $g$ is calculated by $g=(\bm X_0-\bm X).\bm n$ where $\bm X_0$ is the corresponding projection of $\bm X$ onto the rigid obstacle. The normal component of $u$ is given by $u_n = \bm u . \bm n$ where $n$ is the outward normal vector of the object in the actual configuration, and the normal component of first Piola-Kirchhoff stress tensor is given by $P_n = (\bm P. \bm N). \bm n$. Inequation \ref{contact_1} indicates that no penetration can occur, that is, the gap between the body and the obstacle can only be positive or zero. Inequation \ref{contact_2} forces the non-adhesive stress in the contact zone. Equation \ref{contact_3} indicates that when there is contact (the gap is zero, $u_n=g$), the normal stress is different than zero ($P_n<0$), and when there is no contact (the gap is positive, $u_n<g$), the normal stress is identical to zero. The contact conditions can be integrated into the potential energy as follows:
\begin{align*}
    \Pi(\bm\phi) =  \int_{\bm{\Omega}}\Psi dV -\Big(\int_{\bm{\Omega}} \bm{f}_b.\bm \phi dV +\int_{\bm{\Omega}_N} \bm t^*.\bm \phi dA\Big) - \int_{\bm \Omega_C}P_n (u_n-g)
\end{align*}
where $P_n\in \mathcal{P}$, and $\mathcal{P}=\{q\in \mathcal{H}_C| q\le 0\}$ is the space of trial functions. Interested readers may refer to the work of \cite{papadopoulos1992mixed} and \cite{khenous2006hybrid} for further theoretical formulations. In this work, we choose to impose these contact constraints as a strong formulation. More details are described in section \ref{sec_pinns}.
% \textcolor{green}{Jean: attention, ici cette inégalité concerne la configuration déformée. Pour récupérer une "force" sur la déformée, il faut multiplier le premier tenseur de Piola-Kirchoff par la normale à la non-déformée. Puis multiplier cette quantité sur la déformée par la normale à la déformée. Donc $P_n = (P.N).n$ } \textcolor{blue}{Khoa: ah désolée, je n'ai pas précisé mais $n$ c'est le vector normal de l'obstacle et pas de l'objet, donc ça ne change pas avant et après la déformation je pense. Corrige-moi si je dis des betise @@ Je vais ajouter une phrase pour la définition de $n$ si tu es d'accord}\textcolor{green}{Jean: haaaa mais oui ! Je n'avais pas compris. Précise le oui ;-)} \textcolor{blue}{Khoa: Merci Jean :)) J'ai corrigé $P_n = (P.N).n$ et précisé que $n$ est le normal de l'objet après la déformation.}

\section{Methodology}
In this section, the framework of the vanilla Physics-informed neural networks (PINNs) for solving contact mechanics problems is presented. The Deep Energy Method (DEM) based on PINNs is then illustrated. Later in this section, the Geometry-aware Deep Energy Method (GADEM) is introduced. 

\subsection{Physics-informed neural networks}\label{sec_pinns}
The general framework of PINNs for solving problems involving PDE or systems of PDEs can be found in \cite{raissi2019physics, karniadakis2021physics}. Here we present the framework of PINNs for solving forward problems for contact mechanics, which involves solving the systems of equations (\ref{pde_1}),(\ref{pde_2}),(\ref{pde_3}) with the constraints (\ref{contact_1}),(\ref{contact_2}),(\ref{contact_3}).

In the conventional framework of PINNs \citep{raissi2019physics}, the displacement $\bm{u}$ is approximated by a fully connected feedforward neural network $\mathcal{NN}$, which takes the spatial coordinate $\bm{X}$ (positions of the object at initial configuration) as inputs and the solution $\bm{u}$ as output. This can be represented as follows:
\begin{align*}
    \bm{u} \approx \bm{\hat{u}} = \mathcal{NN}(\bm X, \bm{\theta})
\end{align*}
where $\mathcal{NN}$ denotes the neural networks, $\bm{\hat{u}}$ denotes the prediction value for the displacement and $\bm{\theta}$ denotes the trainable parameters of the neural network. The parameters of the neural network are trained by minimizing the cost function $L$:
\begin{align}
    L(\bm{\theta}) = L_{pde}(\bm{\theta}) + L_{bc_D}(\bm{\theta})+ L_{bc_N} (\bm{\theta}) + L_{contact}(\bm{\theta})\label{loss}
\end{align}
where the terms $L_{pde},L_{bc_D},L_{bc_N}$, and $L_{contact}$ penalize the loss in the residual of the PDE, the Dirichlet boundary condition, the Neumann boundary condition, and the contact conditions respectively, which can be represented as follow:
\begin{align*}
    &L_{pde}(\bm{\theta}) = \dfrac{1}{N_{pde}}\sum_{i=1}^{N_{pde}}||\mathrm{div}\hat{\bm P}(\bm X_i) + (\bm f_b)_i)||^2\quad\\
    &L_{bc_D}(\bm{\theta}) = \dfrac{w_{bc_D}}{N_{bc_D}}\sum_{i=1}^{N_{bc_D}}|\bm{\hat{u}}_i - \bm{u^*}_i|^2\\
    &L_{bc_N}(\bm{\theta}) = \dfrac{w_{bc_N}}{N_{bc_N}}\sum_{i=1}^{N_{bc_N}}|\hat{\bm P}(\bm X_i)\hat{\bm N}(\bm X_i) - \bm{t^*}_i|^2\\
\end{align*}
where $N_{bc_D}, N_{bc_N}$ denote the numbers of training points for the Dirichlet and Neumann boundary condition, and measurements, respectively, and $N_{pde}$ denotes the number of residual points (or collocation points or unsupervised points) of the PDE. $w_{bc_D}, w_{bc_N}$ are the weight coefficients for different loss terms, which are used to adjust the contribution of each term to the cost function. For the contact conditions which involve inequalities, there are different ways to enforce these constraints. \cite{lu2021physics} proposed to use soft constraints, penalty method, and augmented Lagrangian method to enforce the inequalities as hard constraints into the loss function. \cite{sahin2023solving} proposed the sign-based method, sigmoid-based method, and Fischer-Burmeister function to enforce these constraints. The details and performance of each method can be found in the cited references. In this work, we present the results of using the Fischer-Burmeister function as our preliminary results suggest that this method performs better than the others in terms of accuracy and robustness. We use the Fischer-Burmeister function as follows:
\begin{align*}
    \varphi(a,b) = a+b-\sqrt{a^2+b^2}
\end{align*}
which has the property that $\varphi(a,b)=0 \Leftrightarrow a \ge 0, b\ge 0, ab=0$. Then the contact constraints (\ref{contact_1}),(\ref{contact_2}),(\ref{contact_3}) can be implemented as follows:
\begin{align}
    L_{contact}(\bm{\theta}) =\dfrac{w_{c}}{N_{c}}\sum_{i=1}^{N_{c}}|g(\bm X_i)-u_n(\bm X_i) -P_n(\bm X_i) + \sqrt{\big(g(\bm X_i)-u_n(\bm X_i)\big)^2+P_n(\bm X_i)^2}|^2\label{loss_contact}
\end{align}
where $N_c$ denotes the number of training points on the potential contact zone, $w_{c}$ is the weight coefficient for the contact loss term. The weight coefficients $w_{bc_D}, w_{bc_N},w_c$ can be either pre-specified before the training or tuned during the training. To construct the residual loss $L_{pde}$, automatic differentiation (AD) \citep{baydin2018automatic} is used to compute the partial derivatives of the output $\bm{\hat{u}}$ of the network with respect to the inputs $\bm{X}$. 

\subsection{Deep energy method based on PINNs}\label{DEM_section}
\cite{samaniego2020energy,nguyen2021parametric} introduced the Deep Energy Method (DEM) to infer the solution of mechanical problems by minimizing the loss function which is not based on residual equations but on the potential energy of the system. More precisely, the loss function in DEM is formulated as follows:
\begin{align*}
    L(\bm{\theta}) =  \Pi(\hat{\bm{\phi}})+ L_{contact}(\bm{\theta}) =  \int_{\bm{\Omega}}\hat{\Psi} dV -\int_{\bm{\Omega}} \bm{f}_b.\hat{\bm{\phi}} dV -\int_{\bm{\Omega}_N} \bm t^*.\hat{\bm{\phi}} dA +  L_{contact}(\bm{\theta})
\end{align*}
where $\hat{\bm{\phi}}, \hat{\Psi}$ denote the predictions of the neural networks for $\bm\phi$ and $\Psi$, respectively. The trial function is expressed as $\hat{\bm\phi}=\hat{\bm u} + \bm X$, where the prediction for the displacement $\hat{\bm u}$ satisfies the Dirichlet condition \textit{a priori}. To this end, we can multiply the output of the neural networks by a distance function to the Dirichlet boundary $\bm \Omega_D$. The interested readers may refer to the work of \cite{berg2018unified, liu2019solving} for the general formulations. The detail of the distance function used in this work is given in section \ref{application_problem_section}. We note however that we can enforce the boundary conditions softly as in the classical PINNs by minimizing the mean square error (MSE) between the prediction and the solution on these boundaries. The formulation of the loss $L_{contact}$ for contact constraints remains the same as (\ref{loss_contact}).

To approximate the integrals, there exist various numerical methods such as the trapezoidal rule Simpson's rule, Gaussian quadrature, and the Monte-Carlo method. In this work, we employ the Monte Carlo method as this method allows the training points to be randomly generated over the entire domain, which is efficient in case of irregular geometries. The loss function now becomes as follows:
\begin{align*}
    L(\bm{\theta}) \approx \dfrac{1}{N}\sum_{i=1}^N \hat{\Psi}(\bm X_i)V - \dfrac{1}{N}\sum_{i=1}^N (\bm{f}_b)_i.\hat{\bm{\phi}}_iV - \dfrac{1}{N_N}\sum_{i=1}^{N_N}(\bm t^*)_i.\hat{\bm{\phi}}_iA+ L_{contact}(\bm{\theta})
\end{align*}
where $N, N_N$ denotes the number of training points in the domain and on the Neumann boundary, respectively, $V$ denotes the volume (in 3D) or area (in 2D) of the object, and $A$ denotes the area (in 3D) or length (in 2D) of the potential contact zone.

\subsection{Geometry-aware deep energy method}\label{sec_gapinns}
With the purpose of building a model that is able to predict the solution on different geometries, we need a model that takes into account the geometric information. \cite{oldenburg2022geometry} proposed to use a variational autoencoder \citep{kingma2019introduction} on boundary (spatial) coordinates of the objects to infer the latent representation of each geometric shape. Inspired by this work, we separate the training into two phases. In the first phase, we use geometric encoding methods to extract the geometric information into latent vectors. In the second phase, the latent vector is considered as the input of the DEM network. The loss function is now defined by the sum of the loss functions of the classical DEM in each geometry.  In this work, we employ either the
discretized spatial coordinates of the objects’ boundaries or images of the objects to represent the geometries. We
also investigate and compare the performance of the model using different geometric encoding methods: parametric
encoding (that is, the parameters that control the shapes of geometries, if available) and different reduction methods: Principal Component Analysis (PCA) \citep{jolliffe2002principal} and Variational Autoencoder (VAE).  In our framework, during the second phase, we rely on the Deep Energy Method to formulate the loss function, that is, the potential of the system is minimized. In the following, we refer to this framework as the Geometry-aware Deep Energy Method (GADEM).

For the training during the first phase, if the geometric parameters are available, we take them directly as input of the model for the second phase. If not, the spatial coordinates of the objects’ boundaries or the images of the objects are used to represent the geometries. More precisely, each geometry is represented by its boundary spatial coordinates or its black-and-white image represented by a matrix of 0s and 1s. These data are taken as the input of the PCA and VAE. For the VAE, we use 1D (for spatial coordinates) or 2D (for images) convolution layers, followed by fully connected layers to predict the mean $\bm\mu_{\bm X}$ and variance $\bm\sigma_{\bm X}$ of the posterior distribution. In the decoder, the latent vector is taken as input, and fully connected layers are employed to reconstruct the boundaries. The details of the architecture of the VAE are given in the appendix \ref{append_rec}. 

After the training of reduction methods, the geometric latent vector $\bm z$ (that is the geometric parameters (if available) or the latent space of the PCA, or the output of the encoder of the VAE) is taken as an input of GADEM. 
\begin{align*}
    \bm{\hat{u}} = \mathcal{NN}(\bm X,\bm z, \bm{\theta})
\end{align*}
The final loss function of GADEM is the sum of the PINNs loss of all the training geometries. Since we employ the Deep Energy Method, the loss function can be represented as follows:
\begin{align*}
    L_{GADEM}(\bm{\theta}) = \sum_{j=1}^{N_{\mathcal{G}}}\Bigg (\dfrac{1}{N^j}\sum_{i=1}^{N^j} \hat{\Psi}(\bm X^j_i)V^j - \dfrac{1}{N^j}\sum_{i=1}^{N^j} (\bm{f}_b)^j_i.\hat{\bm{\phi}}^j_iV^j - \dfrac{1}{N_N^j}\sum_{i=1}^{N_N^j}(\bm t^*)_i^j.\hat{\bm{\phi}}_i^jA^j+ L^j_{contact}(\bm{\theta}) \Bigg)
\end{align*}

The architecture of GADEM and the training configurations are given in detail in each application. In the following, for the sake of clarity, we denote different approaches of GADEM as follows:
\begin{itemize}
\setlength\itemsep{0.1em}
    \item Parametric: explicit parametric encoding. This approach is straightforward to implement in cases where the parameters that control the shape of geometry are available.
    \item PCA-Coord: spatial coordinates for the geometric representation and PCA for the encoding. This approach can be employed when we dispose of the spatial coordinates of the objects' boundaries. The training of PCA is fast (compared to the VAE), however, the PCA can only learn the linear representations. The performance using the PCA may decrease when dealing with complex geometries.
    \item VAE-Coord: spatial coordinates for the geometric representation and VAE for the encoding. As PCA-Coord, this approach can be employed when the spatial coordinates of the objects' boundaries are available. As the VAE can learn non-linear representation, it offers greater flexibility than PCA when dealing with complex geometries.
    \item PCA-Image: images for the geometric representation and PCA for the encoding. This approach can be employed when the objects' images are available. The advantages and drawbacks of using PCA are discussed above. 
    \item VAE-Image: images for the geometric representation and VAE for the encoding. This approach can be employed when the objects' images are available. The advantages and drawbacks of using the VAE are discussed above. 
\end{itemize}
The configuration and architecture of the neural networks are specified in each test case. The detailed configuration and performance of the PCA and VAE can be found in the appendix.

\subsection{Adaptive learning strategy for collocation points for energy-based methods}
In PINNs and DEM, the loss for PDE residuals or the potential energy is computed on a set of collocation points (non-supervised points). In the literature on vanilla PINNs, the training collocation points are often fixed during the training process \citep{raissi2019physics, karniadakis2021physics}. It has been shown that the location of these collocation points has a great impact on the performance of PINNs \citep{lu2021deepxde,daw2022rethinking,nguyen2022physics}. Recently, a considerable amount of work has also developed adaptive re-sampling techniques for the collocation points during the training \citep{peng2022rang,daw2022rethinking, zeng2022adaptive,wu2023comprehensive}. \cite{nguyen2023fixed} proposed the Fixed-Budget Online Adaptive Learning (FBOAL) strategy that decomposes the domain into smaller sub-domains, then adds and removes the collocation points based on the PDEs residuals so that the local extremum of the PDEs residuals can be quickly captured by the method given a constant budget of collocation points. The authors also show that the method can be utilized in parametric cases, where the PDE parameters may vary. In this work, we employ FBOAL to improve the performance of GADEM. The main idea of FBOAL is to add and remove collocation points that yield the largest and the smallest PDE residuals on the sub-domains during the training. After the training, the number of training points remains the same as the initial training configuration. We note that, when parametric encoding or latent vectors are considered as additional inputs of the training, FBOAL can relocate the training points adaptively following the parametric encoding or the latent vectors. The details of the method can be found in the reference \citep{nguyen2023fixed}. In this work, we use a simplified version of FBOAL (i.e., we do not effectuate the domain decomposition step) to infer adaptively the position of collocation points for the energy-based methods (DEM or GADEM). We recall the loss function of DEM:
\begin{align*}
    L(\bm{\theta}) =\int_{\bm{\Omega}}\hat{\Psi} dV -\int_{\bm{\Omega}} \bm{f}_b.\hat{\bm{\phi}} dV -\int_{\bm{\Omega}_N} \bm t^*.\hat{\bm{\phi}} dA +  L_{contact}(\bm{\theta})
\end{align*}
In this loss function, the training points are from three different sets $\bm \Omega, \bm \Omega_N,\bm \Omega_C$, and on each set, different independent quantities are minimized. Thus we can treat these training sets independently. For the sake of simplicity, in the following, we describe the FBOAL algorithm on the domain $\bm\Omega$. Our goal is to minimize the following quantity on $\bm\Omega$:
\begin{align*}
    \int_{\bm{\Omega}}\hat{\Psi} dV -\int_{\bm{\Omega}} \bm{f}_b.\hat{\bm{\phi}} dV = \int_{\bm{\Omega}}(\hat{\Psi}- \bm{f}_b.\hat{\bm{\phi}})dV \approx \dfrac{1}{N}\sum_{i=1}^N \big(\hat{\Psi}(\bm X_i) -(\bm{f}_b)_i.\hat{\bm{\phi}}_i\big)V = \dfrac{1}{N}\sum_{i=1}^N e(\bm X_i) V
\end{align*}
where we denote the quantity $e(\bm X_i)=\hat{\Psi}(\bm X_i) -(\bm{f}_b)_i.\hat{\bm{\phi}}_i$. The simplified FBOAL algorithm effectuated on $\bm \Omega$ is described as follows:
\begin{algorithm}[H]
\begin{algorithmic}[1]
\caption{Simplified Fixed-Budget Online Adaptive Learning (FBOAL)}
\label{fboam_algo}
\REQUIRE{The number of added and removed points $m$, the period of resampling $k$}
\STATE Generate the set $\mathcal{C}$ of collocation points inside the domain $\bm \Omega$.
\STATE Train the networks for $K$ iterations.
\REPEAT
\STATE Generate a new set $\mathcal{C'}$ of random points inside the domain $\bm\Omega$.
\STATE Compute the quantity $e(\bm X'_i)$ at all points in the set $\mathcal{C'}$ and sort $\bm X'_i$ in decreasing order.
\STATE Compute the quantity $e(\bm X_i)$ at all points in the set $\mathcal{C}$ and sort $\bm X_i$ in decreasing order.
\STATE Update $\mathcal{C}\leftarrow \mathcal{C}\backslash \{\bm X_i\}_{i=m}^{N}\cup \{\bm X_i'\}_{i=1}^m$
\STATE Train the networks for $k$ iterations.
\UNTIL The maximum number of iterations is reached.
\end{algorithmic}
\end{algorithm}
A similar approach can be effectuated on $\bm \Omega_N,\bm \Omega_C$. For GADEM, we apply the same approach as described above, except that now we calculate the quantity $e(\bm X_i)$ over all the training geometries.

\section{Numerical results on an academic test case}

Before tackling a complex problem involving hyperelastic material and contact constraints, we validate  GADEM in an academic case of linear elasticity without contact. The goal is to assess and compare the performance of different approaches for geometric representation and encoding in GADEM. More precisely, we investigate the displacement of a beam whose geometry depends on five parameters $l, d, l_1, p_1, p_2$, and the right side of the beam is shaped by two parameters $p_1,p_2$. The left side of the beam is clamped and the right side is subjected to a traction $\Vec{P}=(0, -1)N$. The beam is made of a homogeneous, isotropic material with Young's modulus $E=1000N/m^2$, and Poisson's ratio $\nu=0.3$. The reference solution is calculated using the Finite Element Method (FEM) and is obtained by Fenics \citep{logg2012automated}. To evaluate the accuracy of the prediction, we calculate the relative $\mathcal{L}^2$ error defined as follows:
\begin{align*}
    \epsilon_{\bm u} = \dfrac{||\bm u-\hat{\bm u}||_2}{||\bm u||_2}
\end{align*}
where $\bm u$ denotes the reference solution for the displacement and $\hat{\bm u}$ is the corresponding prediction.

\begin{figure}[H]
    \centering
    \subfloat[Training set 1]{{\includegraphics[width=4cm]{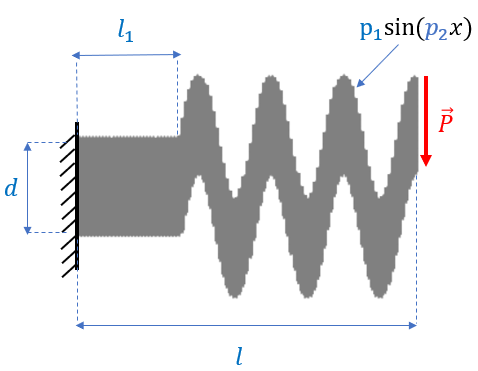} \label{paper_gadem_2d_geo_train1}}}%%
    \subfloat[Training set 2]{{\includegraphics[width=4cm]{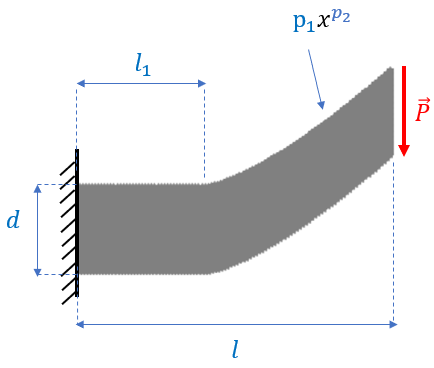}\label{paper_gadem_2d_geo_train2}}}
    \subfloat[Testing set 3]{{\includegraphics[width=4cm]{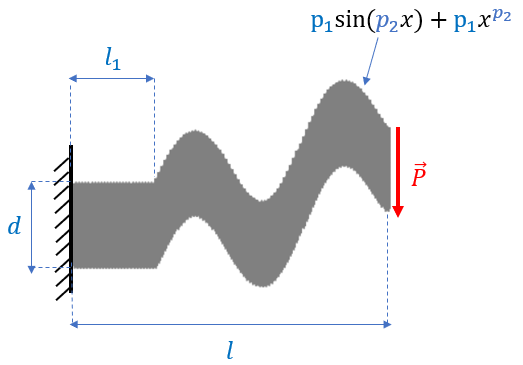} \label{paper_gadem_2d_geo_test3}}}%%
    \subfloat[Testing set 4]{{\includegraphics[width=4cm]{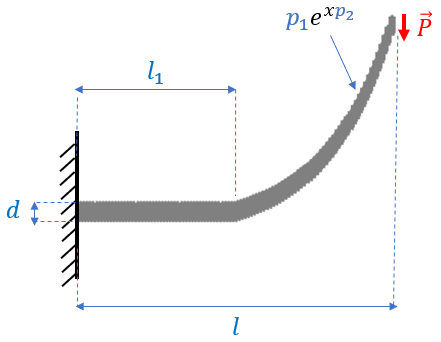}\label{paper_gadem_2d_geo_test4}}}
    \caption{\textit{Visualization of the training and testing geometries and problem setup.}}%
    \label{2d_gadem_config}%
\end{figure}

For the training geometry data set, we use a Latin Hypercube Sampling (LHS) to generate $50$ geometries where the right side of the beam is of the shape $p_1\sin(p_2x)$ and 50 geometries where the right side of the beam is of the shape $p_1x^{p_2}$. Figures \ref{paper_gadem_2d_geo_train1} and \ref{paper_gadem_2d_geo_train2} show an example of geometries in each training set together with the problem setup. The training values of the parameters are detailed in Table \ref{config_geo}.

With the purpose of testing the capability of GADEM in various types of geometries, for the testing geometry set, we use LHS to generate four types of testing. In the first testing set, we generate the geometries with the same shapes as the training set, and the parameters in the same intervals as the training. In the second testing set, we generate the geometries with the same shapes as the training set, however, the parameters are in different intervals than the ones used for training. In the third testing set, we generate the parameters in the same intervals as the training, however, the geometries are now of the shape $p_1\sin(p_2x) + p_1x^{p_2}$. In the fourth set, we generate the parameters in the same intervals as the training, however, the geometries are now of the shape $p_1\exp(p_2x)$. We see that the geometries in the third set are new, but still a linear combination of the training set. The geometries in the fourth set are completely new to the training. Figures \ref{paper_gadem_2d_geo_test3} and \ref{paper_gadem_2d_geo_test4} show an example of geometries in the third and the fourth testing set. The values of the parameters are detailed in Table \ref{config_geo}. 
% # l_bounds = [4, 1, 1, 0.5, 0.25]
% # u_bounds = [12, 3, 2, 1.5, 0.5]
\begin{table}[H]
\centering
\begin{tabular}{|c|c|c|c|}
\hline
 & Nb. geometries & Shape & Interval of parameters $l,d,p_1,p_2,p$ \\
\hline
Train & $50$ &$p_1sin(p_2 x)$ & $[4,12], [1, 3], [1, 2], [0.5, 1.5], [0.25, 0.5]$ \\
              & $50$ & $p_1x^{p_2}$& $[4,12], [1, 3], [0.1, 1], [1, 1.5], [0.25, 0.5]$  \\
\hline
Test 1 & $10$ &$p_1sin(p_2 x)$ & $[4,12], [1, 3], [1, 2], [0.5, 1.5], [0.25, 0.5]$ \\
              & $10$ & $p_1x^{p_2}$& $[4,12], [1, 3], [0.1, 1], [1, 1.5], [0.25, 0.5]$  \\
\hline
Test 2 & $10$ &$p_1sin(p_2 x)$ & $[12,14], [3, 4], [2, 2.5], [0.25, 0.5], [0.5, 0.6]$ \\
              & $10$ & $p_1x^{p_2}$& $[2,4], [0.5, 1], [0.05, 0.1], [1.5, 2], [0.1, 0.25]$  \\
\hline
Test 3 & $20$ &$p_1sin(p_2 x) +p_1x^{p_2} $&  $[4,12], [1, 3], [1, 2], [0.5, 1.5], [0.25, 0.5]$\\
\hline 
Test 4 & $20$ & $p_1\exp(p_2 x)$&  $[4,12], [1, 3], [1, 2], [0.5, 1.5], [0.25, 0.5]$\\
\hline 
\end{tabular}
\caption{\textit{Configuration for the geometries in training and testing sets.}}
\label{config_geo}
\end{table}

\textbf{First training phase} (i.e. extracting the geometric latent vector): For the training of GADEM during the first phase, to represent the geometries using spatial coordinates, we dispose of $N_b={400}$ regular points on the boundaries of each geometry. To represent the geometries using images, we dispose of an image of resolution $200\times200$. To compare fairly different approaches of geometric encoding, the number of kept components in the PCA and the size of the latent vectors in the VAE are fixed $k_{\bm z}=5$, which is equal to the number of geometric parameters ($l, d, l_1, p_1, p_2$). The architecture of the VAE is specified in the appendix \ref{append_rec_1}. The visualization of the latent vectors is also provided in the appendix \ref{append_rec_2}.

\textbf{Second training phase:} During the second phase of GADEM, for each geometry, we randomly generate $N_j=5000$ collocation points. Fully connected layers are employed with 5 layers and 100 neurons per layer, and the hyperbolic tangent is used as an activation function. This architecture is inspired by the work of \cite{raissi2019physics}, allowing a good balance between the network representation capacity and the computational costs based on our preliminary results. To minimize the loss function, we adopt L-BFGS optimizer with the number of maximum epochs equal to $5\times 10^4$. A distance function is used so that the prediction for the displacement satisfies the Dirichlet boundary conditions \textit{a priori}. As the left side of the beams is clamped, that is $\bm u = \bm 0$ on $\partial \bm \Omega_D=\{\bm X=(x,y) | x=0\}$, the output of the neural networks is of the form: $\bm{\hat{u}} = x\mathcal{NN}(\bm X,\bm z, \bm{\theta})$. 
% It is worth noting that as our goal is to compare the performance of different approaches of GADEM, we use a very small number of training collocation points ($N_j=1000$ for each geometry). The accuracy can be further improved by increasing this number of training points.

\begin{figure}[H]
    \centering
    \subfloat[Training set]{{\includegraphics[width=3.2cm]{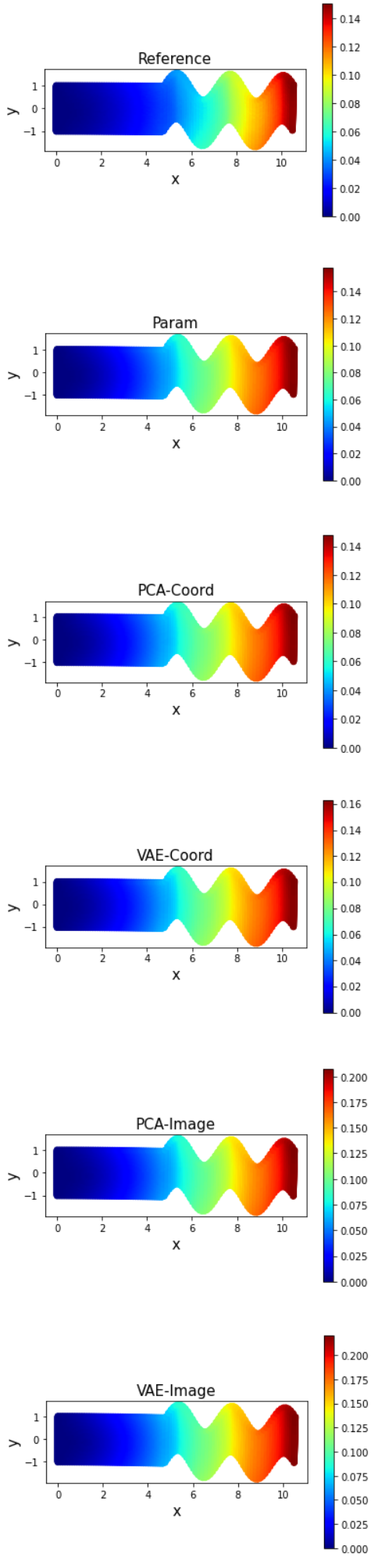}\label{2d_ref_train}}}
    \subfloat[Test 1]{{\includegraphics[width=3.2cm]{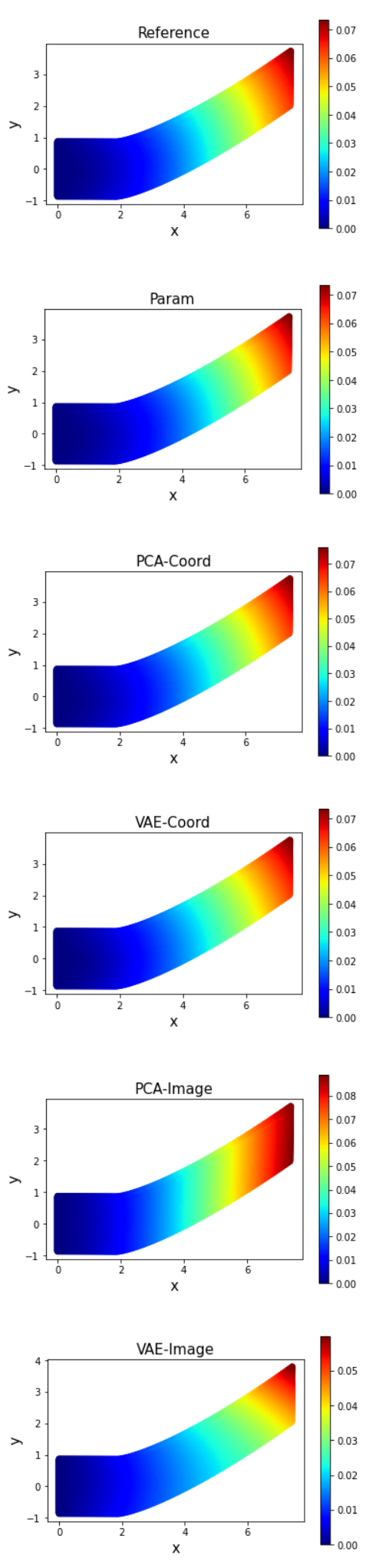}\label{2d_ref_test1}}}
    \subfloat[Test 2]{{\includegraphics[width=3.2cm]{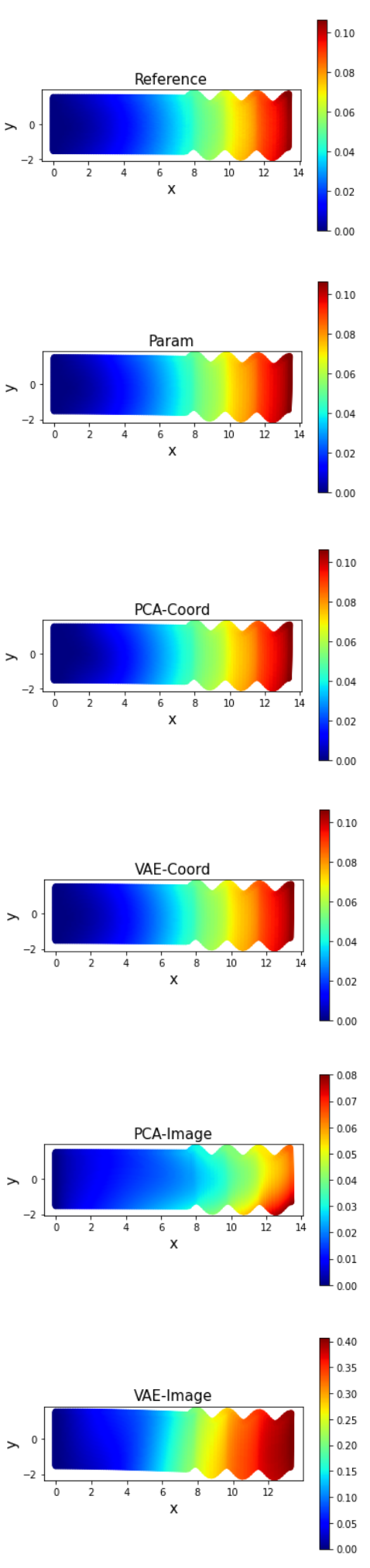}\label{2d_ref_test2}}}
    \subfloat[Test 3]{{\includegraphics[width=3.2cm]{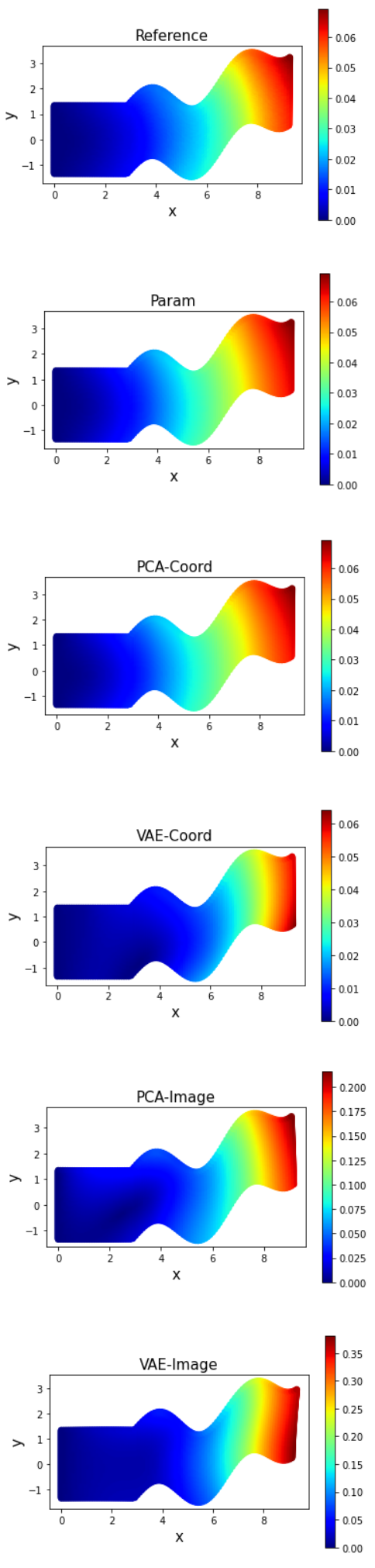}\label{2d_ref_test3}}}
    \subfloat[Test 4]{{\includegraphics[width=3.2cm]{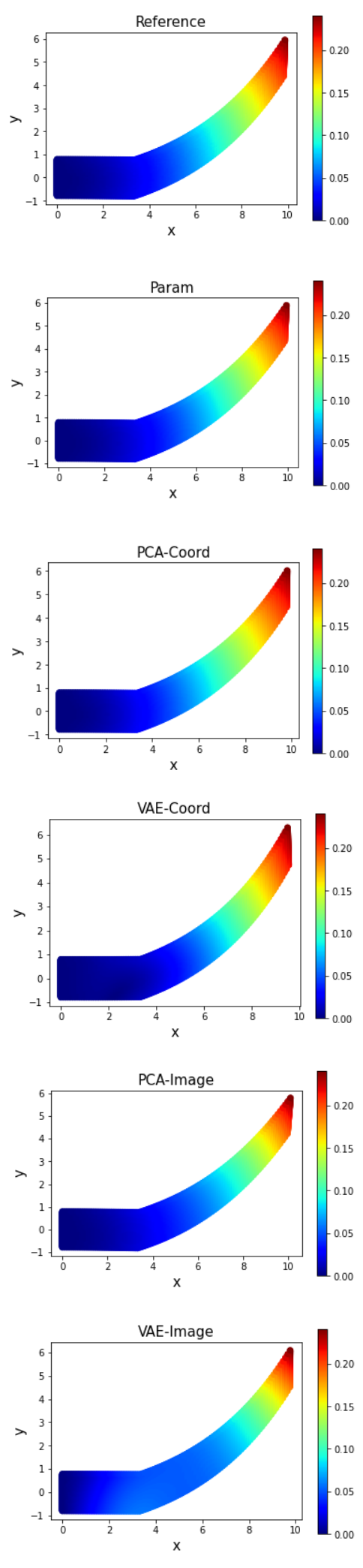}\label{2d_ref_test4}}}
    \caption{\textit{Reference solution and prediction by GADEM on training and testing sets.}}%
    \label{visu_2d_all}%
\end{figure}

Figure \ref{visu_2d_all} shows the visualization of the reference solution and the predictions of GADEM for some examples on the training set and testing sets. We see that all approaches of GADEM give good predictions for the displacements in the training set. For the testing sets, the parametric encoding, PCA-Coord, and VAE-Coord are able to provide accurate predictions, while PCA-Image and VAE-Coord give less accurate predictions.

% Figure \ref{gadem_2d_loss} visualizes the loss function during the training of GADEM approaches. We see that the methods using spatial coordinates (PCA-Coord and VAE-Coord) can better minimize the potential energy of the systems than the ones using images (PCA-Image and VAE-image). 

\begin{figure}[H]
\centering
\includegraphics[width=8cm]{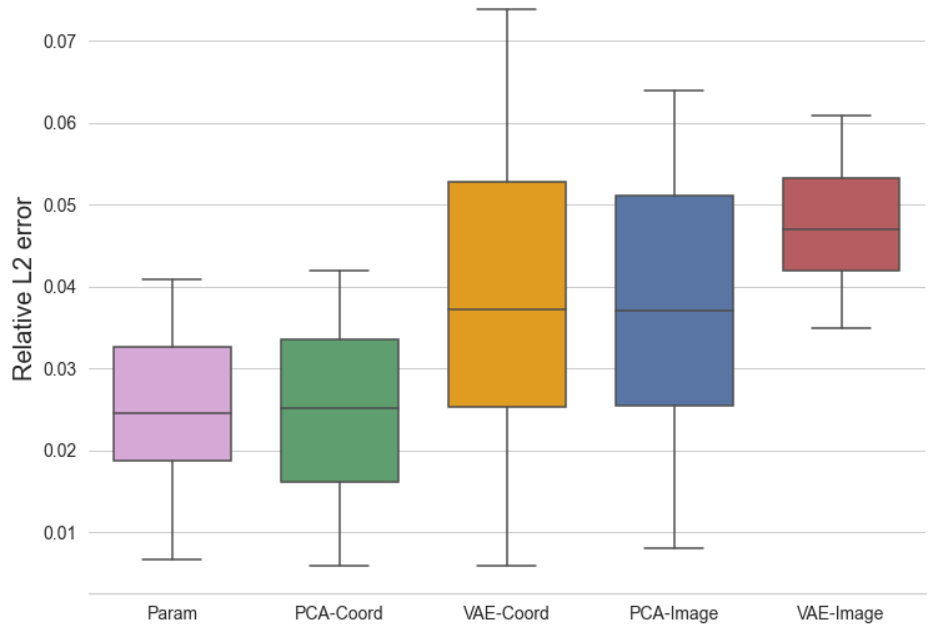}
\caption{\textit{Performance for all geometries in the training data set.}}
\label{perf_train_2d}
\end{figure}

\begin{figure}[H]
    \centering
    \subfloat[Test 1]{{\includegraphics[width=8cm]{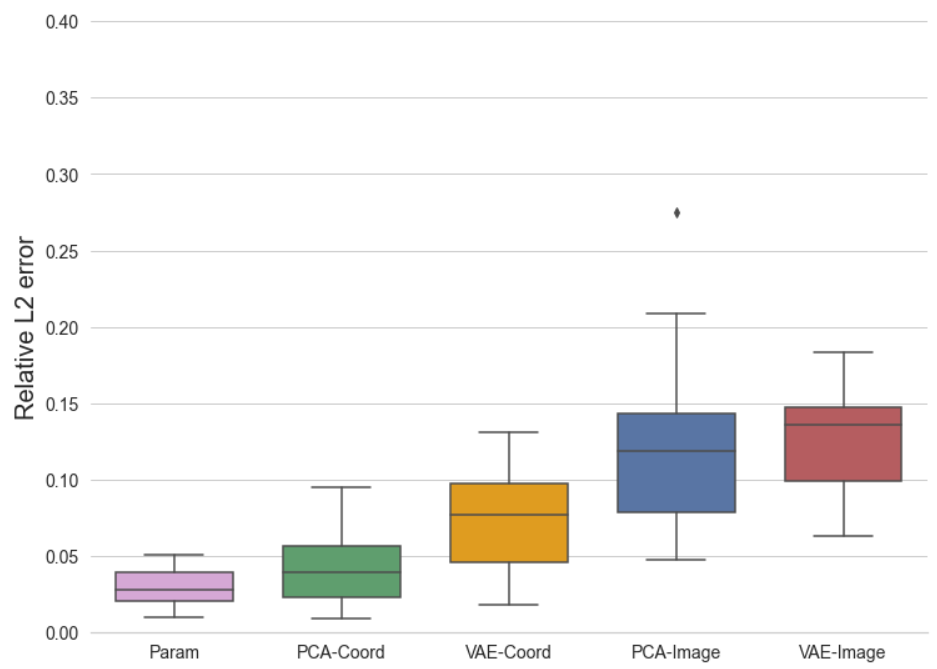}\label{boxplot_test1}}}
    \quad
    \subfloat[Test 2]{{\includegraphics[width=8cm]{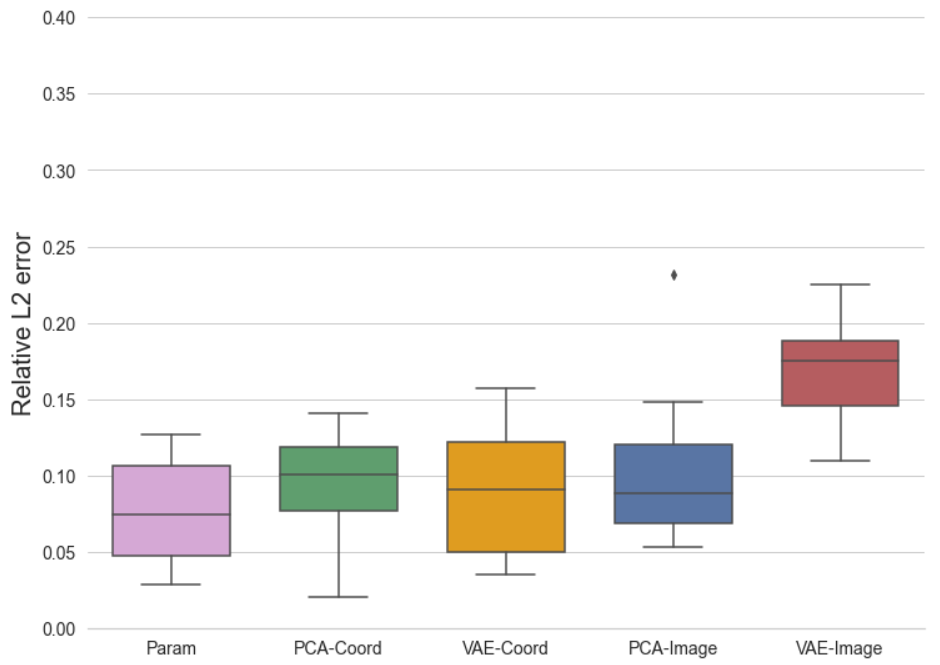}\label{boxplot_test2}}}
    \quad
    \subfloat[Test 3]{{\includegraphics[width=8cm]{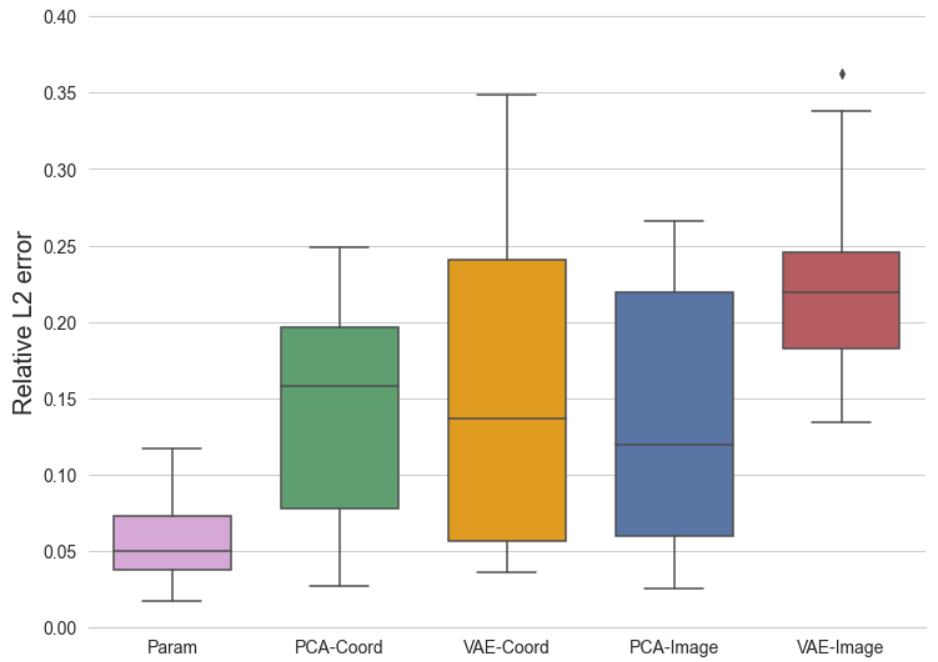}\label{boxplot_test3}}}
    \quad
    \subfloat[Test 4]{{\includegraphics[width=8cm]{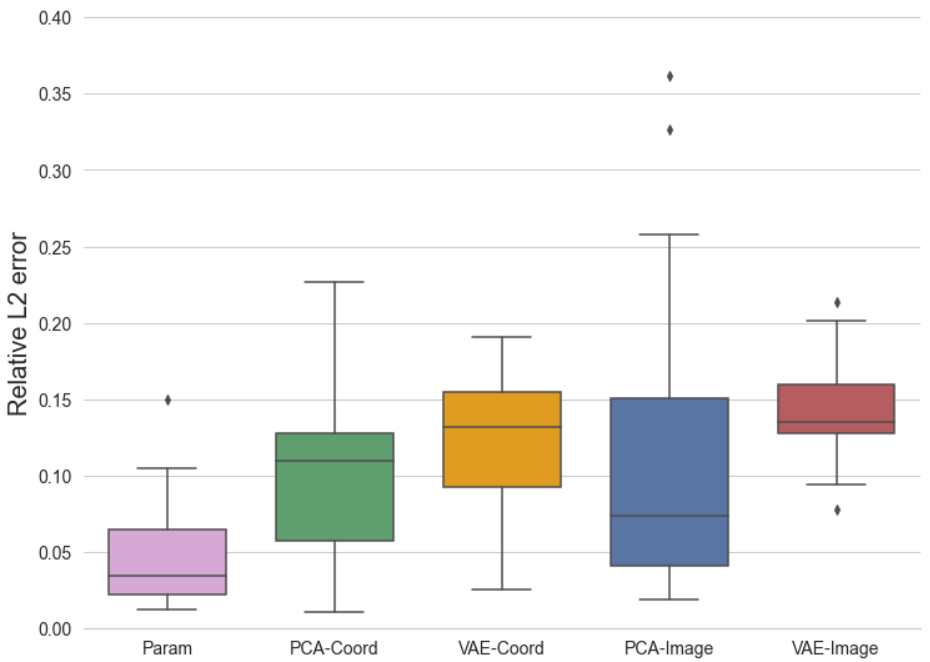}\label{boxplot_test4}}}
    \caption{\textit{Performance on the testing data sets.}}%
    \label{boxplot_2d_test}%
\end{figure}

% \begin{figure}[H]
% \centering
% \includegraphics[width=6.5cm]{paper_gadem_2d_loss.png}
% \caption{\textit{Loss function during the second training phase of GADEM approaches.}}
% \label{gadem_2d_loss}
% \end{figure}
To evaluate the accuracy of GADEM, we choose testing points for each geometry. More precisely, for each geometry, we choose $N_{test}=10000$ points and evaluate the error of the prediction on these testing points. Figures \ref{perf_train_2d} and \ref{boxplot_2d_test} illustrate more precisely the performance of all approaches in terms of accuracy evaluated on testing points. Figure \ref{perf_train_2d} shows the box plots of relative $\mathcal{L}^2$ errors of the prediction of GADEM for all the geometries in the training geometry data set. We see that all the approaches of GADEM are able to give good accuracies for the prediction. The PCA-Coord and VAE-Coord approaches give nearly the same performance and even better than the explicit parametric encodings, whereas the PCA-Image and VAE-Image give the least accurate predictions. Figure \ref{boxplot_2d_test} shows the box plots of relative $\mathcal{L}^2$ errors of the prediction of GADEM for all the geometries in the testing geometry data sets. In general, we see that the parametric encoding always gives good performance for the predictions. The methods that use spatial coordinates (PCA-Coord and VAE-Coord) outperform the ones using images (PCA-Image and VAE-Image). More precisely, in the first test (where the geometries have the same shapes and learning intervals as the training set), the VAE-Coord approach provides the most accurate and robust predictions, followed by PCA-Coord and parametric encoding. In the second test (where the geometries have the same shapes as the training set but different learning intervals), the parametric encoding and VAE-Coord provide good and nearly the same performances.  In the third and fourth tests (where the geometries have different shapes from the training), the parametric encoding provides the best results, followed by PCA-Coord and VAE-Coord approaches.

Table \ref{err_gadem_2d_draft_beam} shows the relative $\mathcal{L}^2$ errors (average results of all the geometries) of all considered approaches on the training and testing sets. The good results (errors smaller than $10\%$) are highlighted. The same conclusion can be drawn from the training and testing sets: parametric encoding always provides good performance in terms of accuracy, and using the spatial coordinates to represent the geometries is better than using the images.  We note that the accuracy of GADEM can be further improved by increasing the number of training collocation points, or using adaptive training methods for the activation function or the weights in the loss function \cite{jagtap2020adaptive,mcclenny2020self}. In this work, we use FBOAL to infer adaptively the training collocation points to enhance the performance of GADEM. 
\begin{table}[H]
\centering
\begin{tabular}{|c|c|c|c|c|c|c|}
\hline
& & Params & PCA-Coord & VAE-Coord & PCA-Image & VAE-Image \\
\hline
$\epsilon_{train}$ & W/o (*) & \textbf{2.52e-2}$\pm$\textbf{9.37e-3} & \textbf{2.46e-2}$\pm$\textbf{1.00e-2} & \textbf{3.85e-2}$\pm$\textbf{1.83e-2} &\textbf{3.79e-2}$\pm$\textbf{1.59e-2} & \textbf{4.74e-2}$\pm$\textbf{6.85e-3} \\
      & With (*) & \textbf{1.67e-2}$\pm$\textbf{1.16e-2} & \textbf{1.83e-2}$\pm$\textbf{1.48e-2} & \textbf{2.78e-2}$\pm$\textbf{2.13e-2}& \textbf{2.54e-2}$\pm$\textbf{2.91e-2} & \textbf{3.89e-2}$\pm$\textbf{2.61e-2} \\
\hline
$\epsilon_{test_1}$ & W/o (*) & \textbf{2.99e-2}$\pm$\textbf{1.26e-2} & \textbf{4.30e-2}$\pm$\textbf{2.62e-2} & \textbf{7.31e-2}$\pm$\textbf{3.51e-2} & 1.27e-1$\pm$6.71e-2& 1.26e-1$\pm$3.05e-2\\
      & With (*)& \textbf{2.21e-2}$\pm$\textbf{1.75e-2} & \textbf{3.72e-2}$\pm$\textbf{3.88e-2} & \textbf{6.58e-2}$\pm$\textbf{4.67e-2} & 1.16e-1$\pm$7.63e-2 & 1.19e-1$\pm$4.12e-2 \\
\hline
$\epsilon_{test_2}$ & W/o (*)& \textbf{7.63e-2}$\pm$\textbf{2.24e-2} & \textbf{9.43e-2}$\pm$\textbf{3.59e-2}& \textbf{9.03e-2}$\pm$\textbf{4.36e-2} & 1.05e-1$\pm$5.09e-2 & 1.71e-1$\pm$3.45e-2\\
      & With (*) & \textbf{7.08e-2}$\pm$\textbf{2.31e-2} & \textbf{8.65e-2}$\pm$\textbf{4.38e-2} & \textbf{8.42e-2}$\pm$\textbf{5.73e-2} & 9.81e-1$\pm$6.96e-2 & 1.64e-1$\pm$3.88e-2\\
\hline
$\epsilon_{test_3}$ & W/o (*) & \textbf{5.68e-2}$\pm$\textbf{2.75e-2} & 1.43e-1$\pm$7.03e-2 & 1.55e-1$\pm$1.08e-1 & 1.36e-1$\pm$8.36e-2 & 2.24e-1$\pm$6.09e-2\\
      & With (*)& \textbf{4.97e-2}$\pm$\textbf{3.28e-2} & 1.35e-1$\pm$7.47e-2 & 1.48e-1$\pm$1.39e-2 & 1.28e-1$\pm$8.93e-2 & 2.19e-1$\pm$6.51e-2\\
\hline
$\epsilon_{test_4}$ & W/o (*)& \textbf{4.99e-2}$\pm$\textbf{3.74e-2} & 1.00e-1$\pm$5.73e-2 & 1.21e-1$\pm$4.61e-2 & 1.19e-1$\pm$1.08e-1 & 1.43e-1$\pm$3.56e-2\\
      & With (*)& \textbf{4.31e-2}$\pm$\textbf{3.45e-2} & \textbf{9.52e-2}$\pm$\textbf{4.19e-2} & 1.15e-1$\pm$5.10e-2 & 1.10e-1$\pm$3.40e-2 & 1.38e-1$\pm$4.34e-1\\
\hline
\end{tabular}
\caption{\textit{Relative $\mathcal{L}^2$ error of different approaches of GADEM with and without using FBOAL (*).}}
\label{err_gadem_2d_draft_beam}
\end{table}

\begin{figure}[H]
    \centering
    \subfloat[Without FBOAL]{{\includegraphics[width=3.5cm]{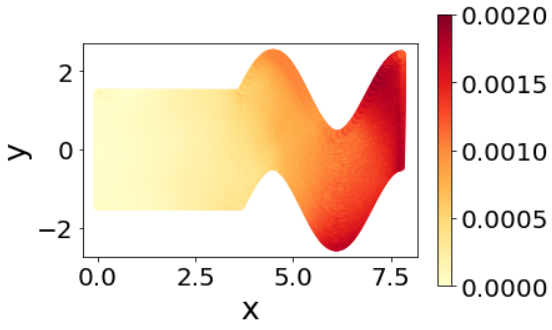}}}
    \subfloat[Without FBOAL]{{\includegraphics[width=3.5cm]{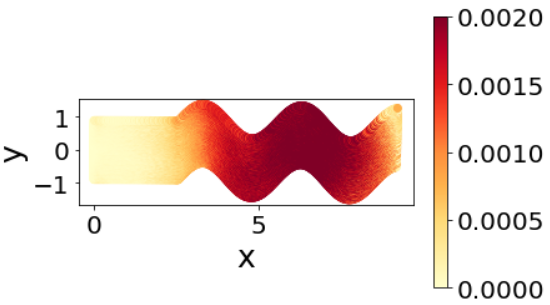}}}
    \subfloat[Without FBOAL]{{\includegraphics[width=3.5cm]{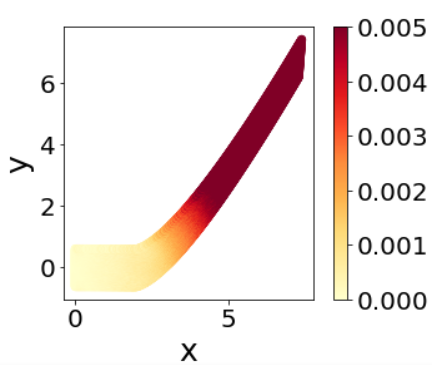}}}
    \subfloat[Without FBOAL]{{\includegraphics[width=3.5cm]{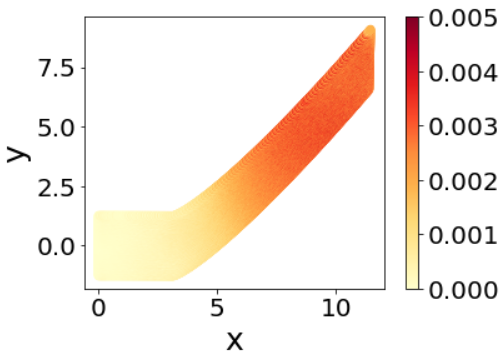}}}
    \quad 
    \subfloat[With FBOAL]{{\includegraphics[width=3.5cm]{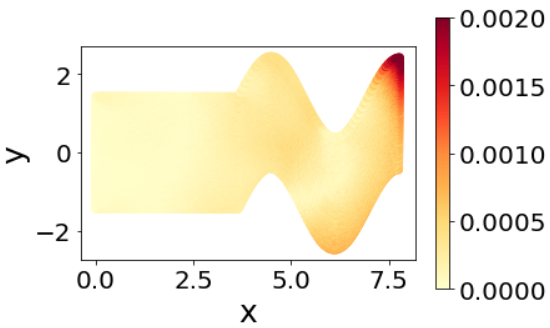}}}
    \subfloat[With FBOAL]{{\includegraphics[width=3.5cm]{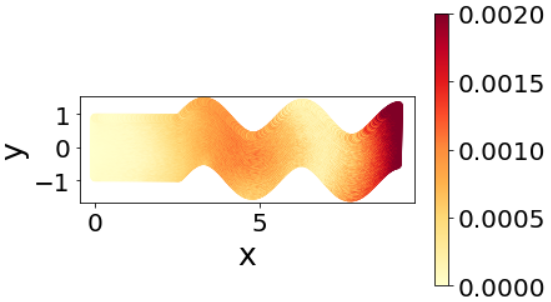}}}
    \subfloat[With FBOAL]{{\includegraphics[width=3.5cm]{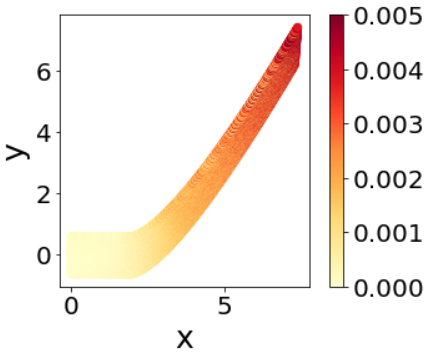}}}
    \subfloat[With FBOAL]{{\includegraphics[width=3.5cm]{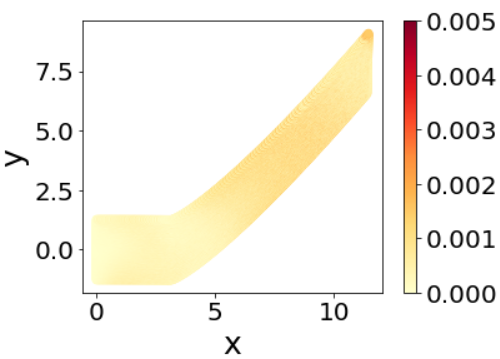}}}
    \caption{\textit{Visualization of the absolute error between the reference solution and GADEM with parametric encoding for beams in the training set: (a-d) without using FBOAL, (e-h) with using FBOAL on corresponding beams.}}%
    \label{fboal_2d_perf}%
\end{figure}

\begin{figure}[H]
    \centering
    \subfloat[Loss function]{{\includegraphics[width=7cm]{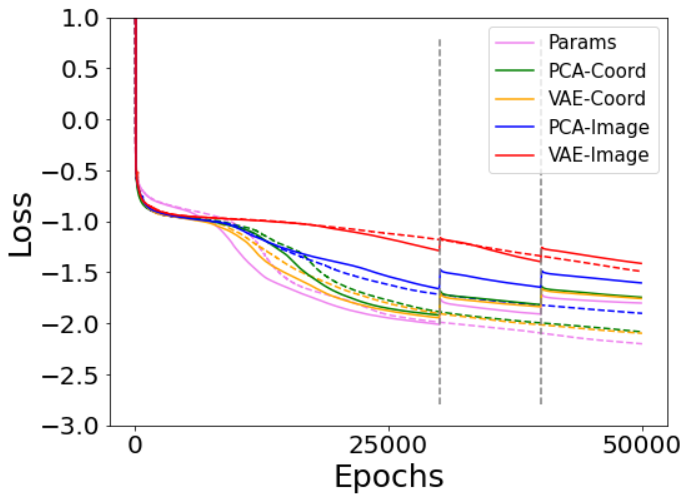}\label{2d_ref_train_loss}}}
    \subfloat[Error on testing points]{{\includegraphics[width=7cm]{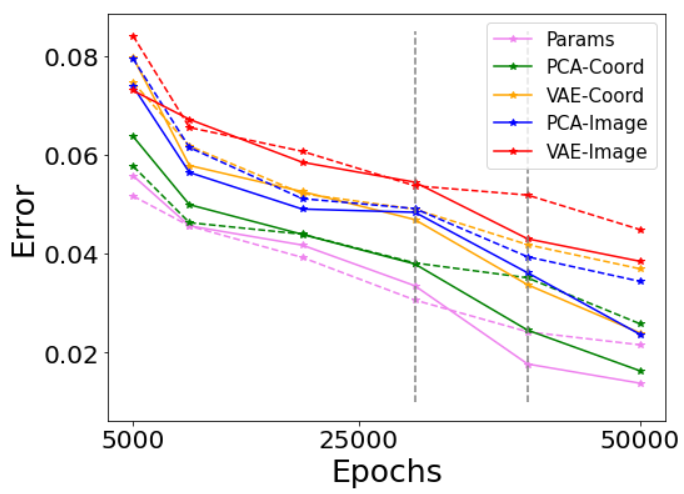}\label{2d_ref_test1_loss}}}
    \caption{\textit{Loss function during the training process and the relative $\mathcal{L}^2$ errors on the testing points (average results of all geometries) produced by GADEM without FBOAL (dashed lines) and GADEM with FBOAL (solid lines). The gray lines indicate the instances where FBOAL effectuates the resampling.}}%
    \label{visu_2d_all_loss}%
\end{figure}

For the sake of comparison, we use the same initial training configuration as described above, that is, for each geometry, we dispose of $N_j=5000$ randomly generated collocation points. This leads to the number of collocation points $N=100\times5000=500000$ points. For the training of FBOAL, we first train the network for $K=30000$ iterations, and then after every $k=10000$ iterations, we add and remove $m=1\%\times N=5000$ points based on the energy quantity to minimize for all the training geometries. We note that the hyperparameters $m$ and $k$ can be tuned to obtain the best performance of FBOAL. The study of these hyperparameters is presented in the work of \cite{nguyen2023fixed}. Here we choose the values of $m$ and $k$ based on the conclusion of this work.

Figure \ref{fboal_2d_perf} illustrates the absolute error between the reference solution and GADEM with parametric encoding for tires in the training set with and without using FBOAL. From this figure and Table \ref{err_gadem_2d_draft_beam}, we can see that FBOAL helps to improve significantly the accuracy of GADEM on the training geometry set. The same conclusion can be drawn for the testing geometry sets. Figure \ref{visu_2d_all_loss} shows the loss function during the training by all GADEM approaches and the relative $\mathcal{L}^2$ errors of the prediction over all geometries on the testing points. We see that when FBOAL effectuates the resampling (\textit{i.e.} the positions of the training collocations points are modified), there are jumps in the loss function. Thus the loss functions produced by GADEM with FBOAL are less well minimized than the ones produced by GADEM without FBOAL (Figure \ref{2d_ref_train_loss}). However, the prediction errors on the testing points are significantly reduced by using GADEM with FBOAL (Figure \ref{2d_ref_test1_loss}).

\section{Application to toy tire loading simulation}

\subsection{Problem configuration and formulation}\label{application_problem_section}

\begin{figure}[H]
\centering
\includegraphics[width=10cm]{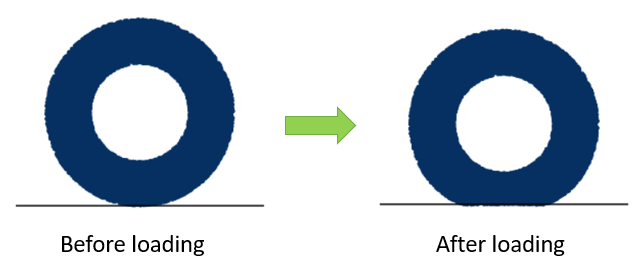}
\caption{\textit{Sketch of a tire before and after loading.}}
\label{crushing}
\end{figure}

During the initial phase of tire design, static loading simulations are a crucial step for the designer to determine the basic properties of the tire. The modeling of the tire loading step is a challenging task especially when dealing with different shapes of tire geometries, different materials, and potentially 3D effects. For the sake of simplicity, the goal of the present study is only to model the 2D displacement of toy tires composed of hyperelastic materials (rubber in our case). An illustration of the tire before and after loading is shown in Figure \ref{crushing}. 

\begin{figure}[H]
\centering
\includegraphics[width=12cm]{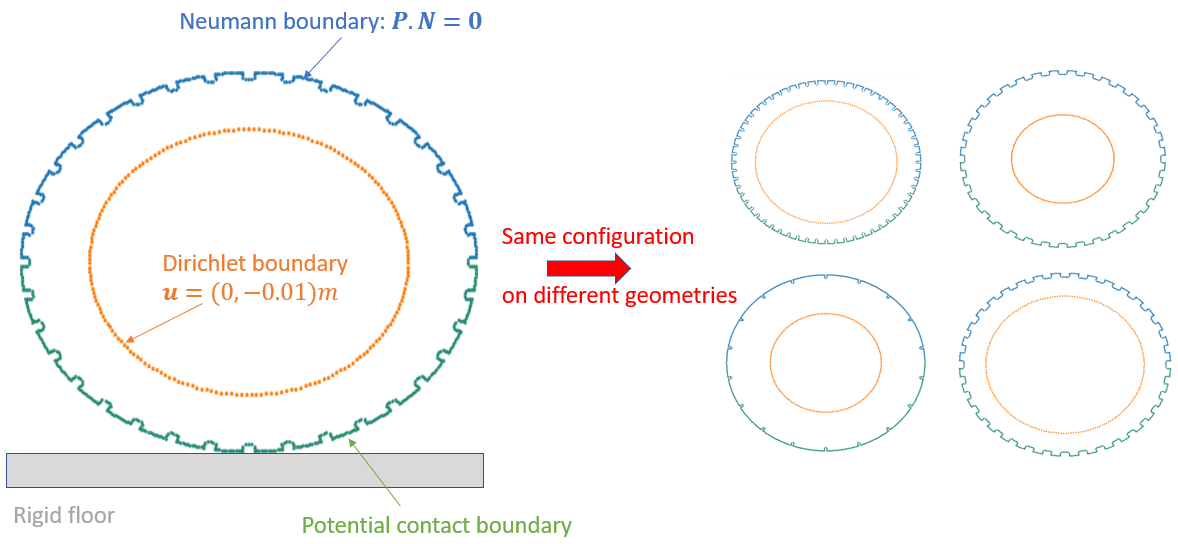}
\caption{\textit{Sketch of the 2D configuration and example of some learning geometries considered in the present study.}}
\label{config}
\end{figure}

The tires are made of rubber with a Young modulus $E=21.10^6 Pa$ and a Poisson's ratio $\nu=0.3$. We assume that the tires follow the Saint-Venant Kirchhoff hyperelastic model. The physical domain of interest together with the geometrical configuration of the study are presented in Figure \ref{config}. The tires are supposed to be in contact with rigid ground. A Dirichlet boundary condition $\bm u_D=(0, -0.01)m$ is imposed on the rim of the tires (the inner circles). A Neumann boundary $\bm P. \bm N=\bm 0$ is applied on the outer boundary of the tires. The potential contact boundaries with the floor can be taken as the lower outer boundaries (shown as green lines in Figure \ref{config}). We suppose that there is no body force applied to the tires. 
% \textcolor{red}{TIBO: ah c'est étrange ça je pensais qu'on appliquant justement seulement une force volumique correspondant à la force de (pesanteur + masse d'un véhicule) par exemple... Mais j'imagine que c'était plus compliqué à gérer car dans ce cas, on ne sait pas quoi mettre comme condition sur la jante ? Bref, laissons comme ça mais note que c'est très bizarre à mon avis.} \textcolor{blue}{Khoa: Ah je ne savais pas que c'etait étrange comme configuration. C'est la configuration que Raph m'a donné initialement je crois. On a déjà essayé le cas où il y a une force volumique et DEM marche aussi, je n'ai pas essayé GADEM sur ce cas.}\textcolor{green}{Jean: cette configuration est la bonne, on néglige les forces volumiques, et le poids véhicule est représenté par le déplacement imposé sur la jante} The potential energy on each tire then becomes:
\begin{align*}
    \Pi(\bm\phi) =  \int_{\bm{\Omega}}\Psi dV = \int_{\bm{\Omega}}\lambda tr^2(\bm{E})\mathbf{I} + \mu tr(\bm{E}^2) dV
\end{align*}
where the Lamé constants are given as $\lambda=\dfrac{\nu E}{(1+\nu)(1-2\nu)}$, and $\mu=\dfrac{E}{2(1+\nu)}$.
To generate reference High-Fidelity (HF) solutions of the displacement field, we use the finite element solver GetFEM \cite{renard2007getfem}. Again, to evaluate the performance of the model in predicting the displacement, we calculate the relative $\mathcal{L}^2$ error defined as follows:
\begin{align*}
    \epsilon_{\bm u} = \dfrac{||\bm u-\hat{\bm u}||_2}{||\bm u||_2}
\end{align*}
where $\bm u$ denotes the reference solution for the displacement and $\hat{\bm u}$ is the corresponding prediction. In our work, the errors are evaluated on the finite element mesh. The number of nodes on the mesh varies with different geometries, from $11934$ points to $32636$ points. The GetFEM solver gives us high-fidelity solutions on the finite-element mesh. 

% To evaluate the performance of the reconstruction of tire geometries, we calculate the distance between the points on the boundary of the geometries and the corresponding reconstruction.
% \begin{align*}
%     \epsilon_{rec} = ||\bm X-\hat{\bm X}||_2
% \end{align*}

\subsection{Numerical results}
% All of the training in this work is effectuated on a GPU card NVIDIA A100. Blabla...

\subsubsection{Extracting the geometric latent vector}\label{sec_geo_vector}
\begin{figure}[H]
\centering
\includegraphics[width=16cm]{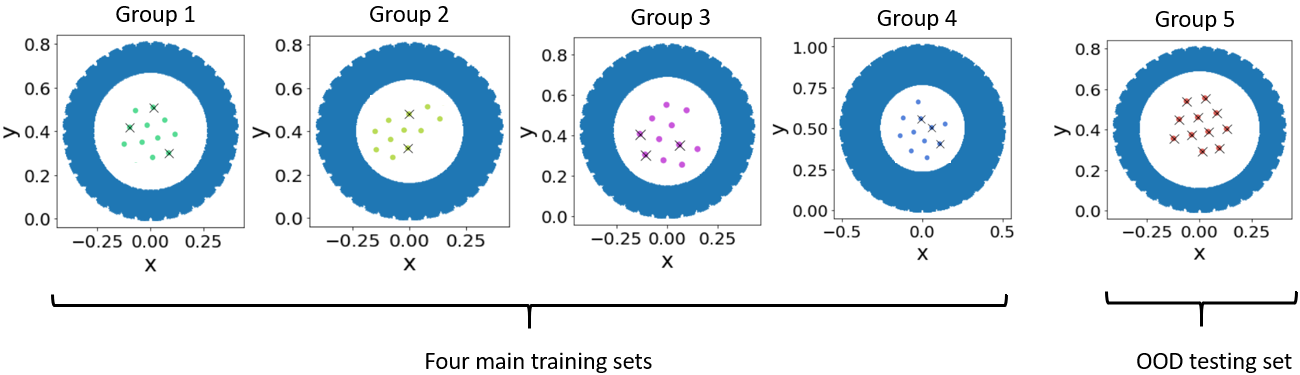}
\caption{\textit{Illustration of the training and OOD testing sets. Inside each training set, we also choose some tires for testing. The testing tires are denoted by the '$\times$' symbol.}}
\label{scheme_train_test}
\end{figure}

\begin{figure}[H]
\centering
\includegraphics[width=12cm]{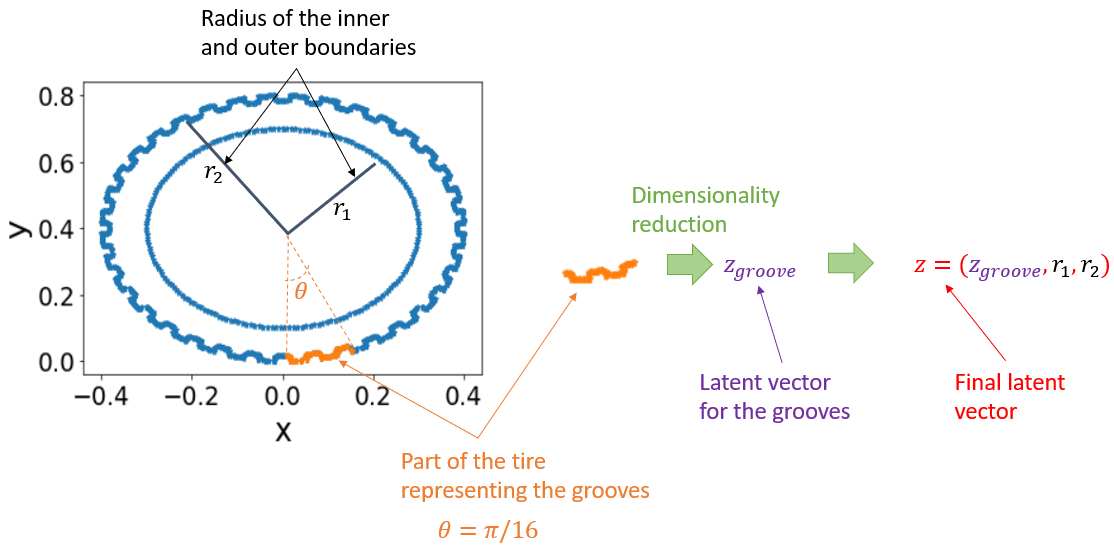}
\caption{\textit{Illustration of the procedure that employs the reduction methods on a part of the tires representing the grooves.} $r_1, r_2$ denotes the radius of the inner and outer boundaries, respectively. Part of the tire representing the grooves is controlled by an angle $\theta=\pi/16$.}
\label{groove}
\end{figure}

% \textcolor{red}{un petit schéma pour expliquer le découpage des sets? Perso je n'ai pas tout compris...} \textcolor{blue}{Khoa: Oui j'ai ajouté une image comme tu as proposé la dernière fois (dis-moi si j'ai mal compris ton idée.} 
We generate a set of $55$ geometries for the tires. In this set, there are $5$ main groups of geometries, which are distinguished by the radius of the inner and the outer boundaries. In each group, the geometries are distinguished by the grooves. The data set is divided into a training set of $33$ geometries in four groups, and two testing sets of $22$ geometries: the first testing set includes $11$ geometries in the same groups as the training set, and the second one includes $11$ geometries in a different group from the training groups (out-of-distribution (OOD) testing set). A visualization for the training set and testing set in a reduced space can be found in Figure \ref{scheme_train_test}. Since our goal is to compare different approaches to represent and encode the geometries, in the following, we choose the fifth group of tires as the OOD testing set. The impact of choosing another group as the OOD testing is presented in the appendix \ref{append_ood_test}.
To represent the geometries using spatial coordinates, we dispose of $N_{b}=1400$ points on the boundaries, which are composed of $1200$ points of the outer boundary, and $200$ points of the inner boundary. To further improve the performance of the reduction methods, we consider employing the reduction methods on only a small part of the tires, which represents the grooves. Then the final latent vector will be composed of the latent vectors produced by the reduction methods on the grooves, and the radius of the inner and outer boundary of the tires. Figure \ref{groove} illustrates explicitly this procedure on a training tire. 

To represent the geometries using images, we dispose of an image of resolution $200\times200$ that represents the boundaries for each geometry. The PCA and VAE are employed to build the latent vectors that represent the geometric features. 
\begin{itemize}
    \item PCA-Coord: the latent vector is of the form $z=(z_{groove}, r_1, r_2)$. We only effectuate PCA on the part that represents the grooves. The number of kept components of the PCA is fixed as $k_{\bm z_{groove}}=3$, or $k_{\bm z}=5$. Before effectuating the PCA, the input data are centered. 
    \item VAE-Coord: the latent vector is of the form $z=(z_{groove}, r_1, r_2)$. We only effectuate VAE on the part that represents the grooves. The size of the latent vector of the VAE is fixed as $k_{\bm z_{groove}}=3$, or $k_{\bm z}=5$.
    \item PCA-Image: the latent vector is of the form $z=z_{tire}$. We effectuate the PCA on the images that represent the boundaries of the tires. The number of kept components of the PCA is fixed as $k_{\bm z}=5$. Before effectuating the PCA, the input data are centered. 
    \item VAE-Image: the latent vector is of the form $z=z_{tire}$. We effectuate the VAE on the images that represent the boundaries of the tires. The size of the latent vector of the VAE is fixed as $k_{\bm z}=5$.
\end{itemize}

More details of the configuration and the performance of the PCA and VAE, and also the visualization of the latent vectors can be found in the appendix \ref{append_rec_1} and \ref{append_rec_3}.

\begin{table}[H]
\centering
\begin{tabular}{ |c|c|c|c|c| } 
\hline
 & PCA-Coord & VAE-Coord & PCA-Image & VAE-Image \\
\hline
Training time & \textbf{3.54e-3} $\pm$ \textbf{8.59e-4} & 1.93e+2 $\pm$ 6.07e+1 & \textbf{1.01e-1} $\pm$ \textbf{7.74e-3}  & 7.66e+2 $\pm$ 5.14e+1 \\
\hline
\end{tabular}
\caption{\textit{Training time (or calculation time in second) during the first phase of GADEM approaches.}}
\label{time_rom}
\end{table}

Table \ref{time_rom} shows the training time during the first phase of all GADEM approaches, \textit{i.e.} the calculation time for the reduction methods to extract geometric latent vectors. We see that with PCA, we gain a huge amount of calculation time compared to the VAE. When these methods are effectuated on the spatial coordinates that represent the grooves, we save even better training time.

\subsubsection{Performance on tire loading simulation}\label{sec_perf_tire}

The extracted geometric latent vector $\bm z$ is taken into the input of GADEM. As we use the Deep Energy Method in the loss function, it is preferred that the prediction of the displacement satisfies the Dirichlet boundary condition \textit{a priori} $\bm u_D=(0, -0.01)m$.  To do this, we multiply the output of the neural network by a distance function to this boundary (which is in our case the inner boundary of the tire), that is:
\begin{align*}
    \hat{\bm u} = \mathcal{NN}(\bm X, \bm z_r, \bm \theta) \cdot \mathrm{dist}(\bm X, bc_D)+\bm u_D
\end{align*}
where $bc_D$ represents the Dirichlet boundary. As the inner boundary of the tires is perfect circles, the distance function can be formulated as:
\begin{align*}
     \mathrm{dist}(\bm X, bc_D) = (X_x - c_x)^2/r^2 + (X_y - c_y)^2/r^2 -1
\end{align*}
where $X_x, X_y$ are the horizontal and vertical components of $\bm X$, $(c_x, c_y)$ represents the coordinates of the center of the circle, and $r$ is the radius of the circle. In the loss function, the potential energy is minimized with the contact constraints for all geometries:
\begin{align*}
    L_{GADEM}(\bm{\theta}) &= \sum_{j=1}^{N_{\mathcal{G}}}\Big (\Pi(\hat{\bm{\phi^j}})+ L_{contact}^j(\bm{\theta})\Big) =  \sum_{j=1}^{N_{\mathcal{G}}}\Big (\int_{\bm{\Omega^j}}\hat{\Psi}^j dV^j +  L^j_{contact}(\bm{\theta}) \Big)\\ &=\sum_{j=1}^{N_{\mathcal{G}}}\Bigg (\dfrac{1}{N^j}\sum_{i=1}^{N^j} \hat{\Psi}(\bm X^j_i)V^j + L^j_{contact}(\bm{\theta}) \Bigg)
\end{align*}
where the loss contact $L^j_{contact}$ is defined as (\ref{loss_contact}).

For the training, on each geometry, we take $N_j=2500$ collocation points, which are randomly generated inside the domain, to minimize the energy. For the contact constraints, we take $N_c=600$ points on the potential contact boundary of each geometry. For the network architecture, fully connected layers are employed with 5 layers and 100 neurons per layer, and the hyperbolic tangent is used for activation functions \citep{raissi2019physics}. To minimize the loss function, we adopt L-BFGS optimizer with the number of maximum epochs equal to $5\times10^5$. To evaluate the models, for each geometry, we choose $N_{test}=20000$ points and calculate the error of the prediction on these testing points.

\begin{figure}[H]
    \centering
    \subfloat[PCA-Coord]{{\includegraphics[width=4cm]{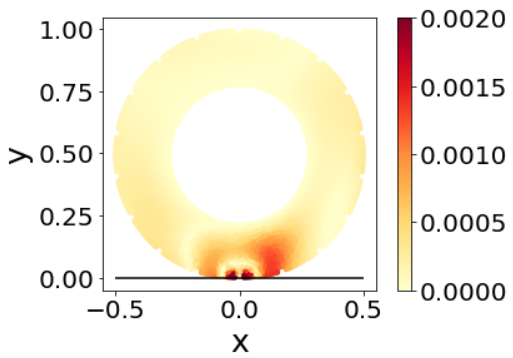}\label{pca_error_pneu_visu}}}
    \subfloat[VAE-Coord]{{\includegraphics[width=4cm]{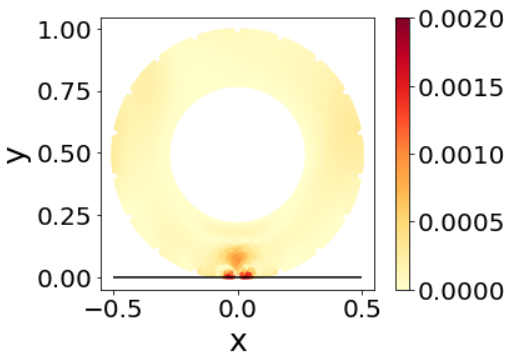}\label{vae_error_pneu_visu}}}
    \subfloat[PCA-Image]{{\includegraphics[width=4cm]{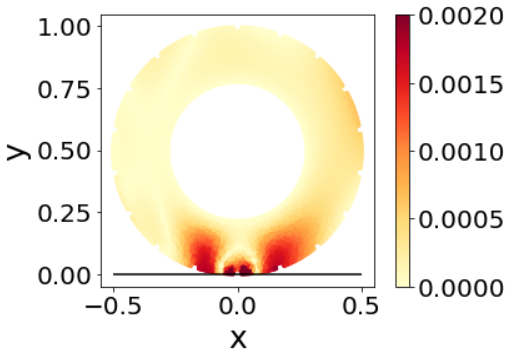}\label{image_pca_error_pneu_visu}}}
     \subfloat[VAE-Image]{{\includegraphics[width=4cm]{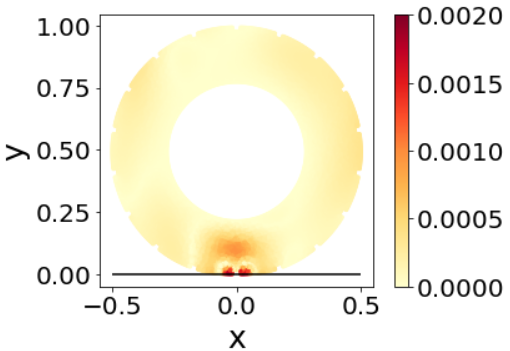}\label{image_vae_error_pneu_visu}}}
    \quad
   \subfloat[PCA-Coord]{{\includegraphics[width=4cm]{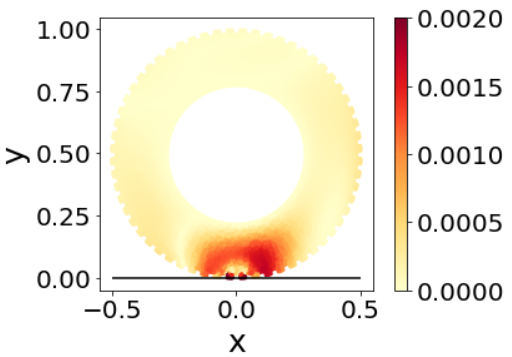}\label{pca_error_pneu_visu_test}}}
    \subfloat[VAE-Coord]{{\includegraphics[width=4cm]{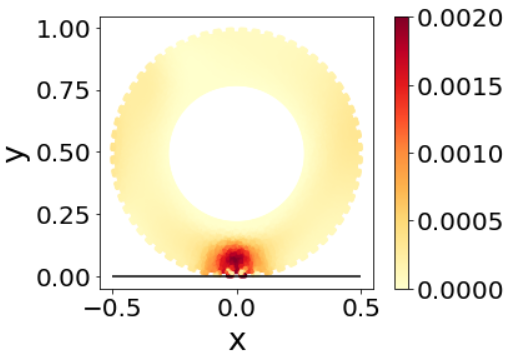}\label{vae_error_pneu_visu_test}}}
    \subfloat[PCA-Image]{{\includegraphics[width=4cm]{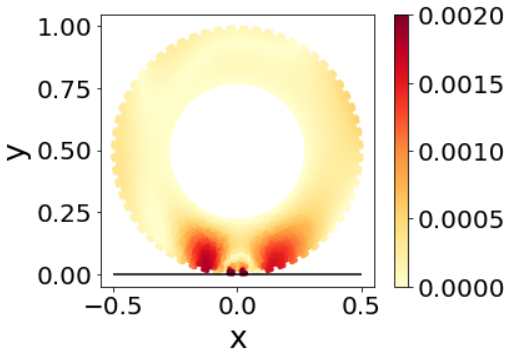}\label{image_pca_error_pneu_visu_test}}}
     \subfloat[VAE-Image]{{\includegraphics[width=4cm]{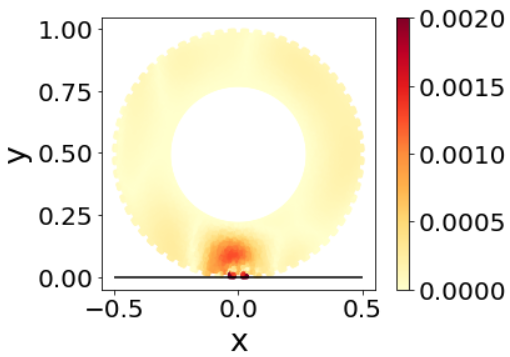}\label{image_vae_error_pneu_visu_test}}}
     \quad
   \subfloat[PCA-Coord]{{\includegraphics[width=4cm]{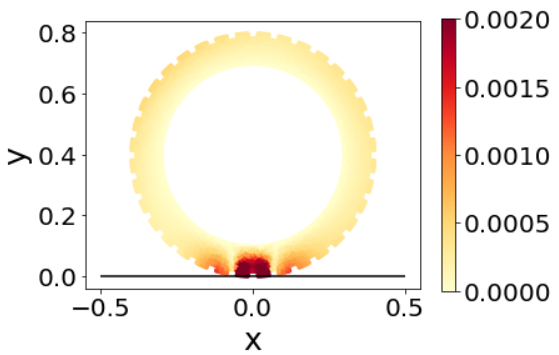}\label{pca_error_pneu_visu_test1}}}
    \subfloat[VAE-Coord]{{\includegraphics[width=4cm]{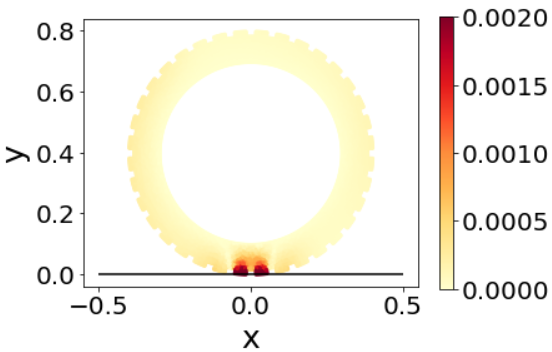}\label{vae_error_pneu_visu_test1}}}
    \subfloat[PCA-Image]{{\includegraphics[width=4cm]{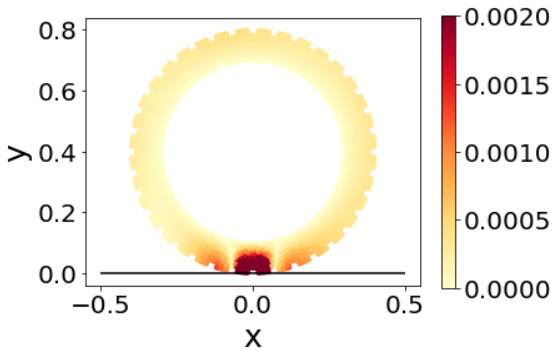}\label{image_pca_error_pneu_visu_test1}}}
     \subfloat[VAE-Image]{{\includegraphics[width=4cm]{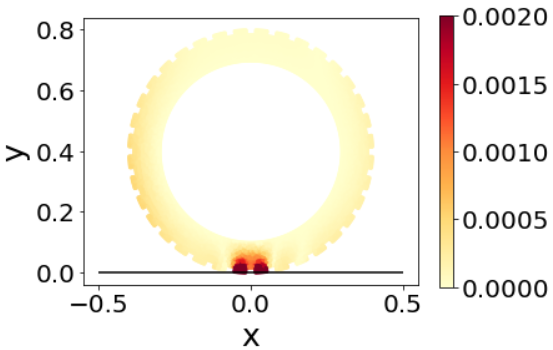}\label{image_vae_error_pneu_visu_test1}}}
    \caption{\textit{Visualization of the absolute error between the reference and the prediction GADEM approaches for a tire in the training set (a-d), first testing set (e-h), and second testing set (i-l).}}%
    \label{pneu_visu_error}%
\end{figure}

\begin{figure}[H]
    \centering
    \subfloat[Reference solution]{{\includegraphics[width=5cm]{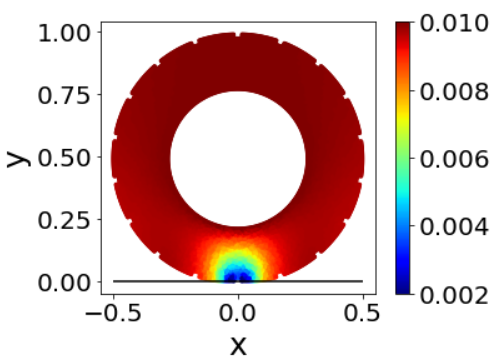}\label{sol_train}}}
    \subfloat[Reference solution]{{\includegraphics[width=5cm]{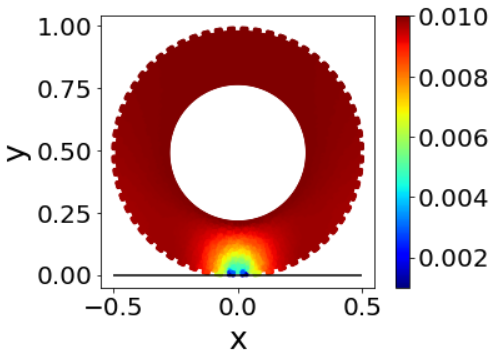}\label{sol_test1}}}
    \subfloat[Reference solution]{{\includegraphics[width=5cm]{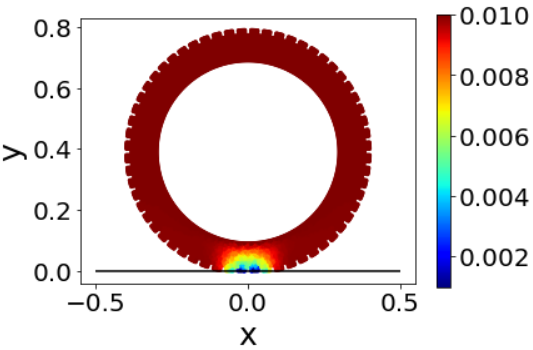}\label{sol_test2}}}
    \quad
    \subfloat[Prediction]{{\includegraphics[width=5cm]{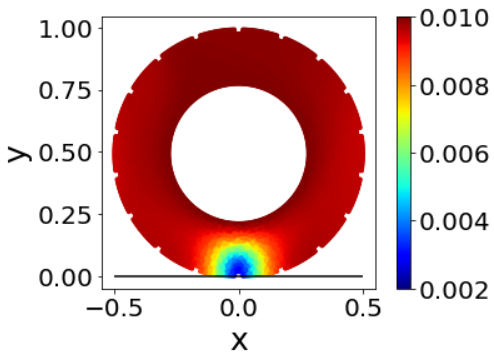}\label{pred_train}}}
    \subfloat[Prediction]{{\includegraphics[width=5cm]{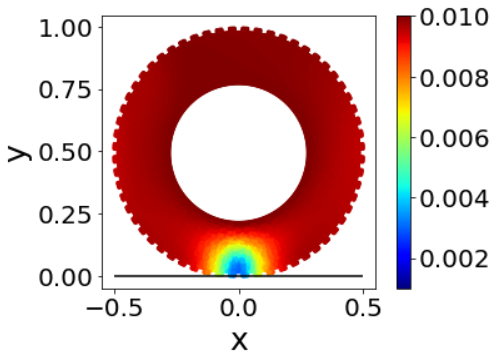}\label{pred_test1}}}
    \subfloat[Prediction]{{\includegraphics[width=5cm]{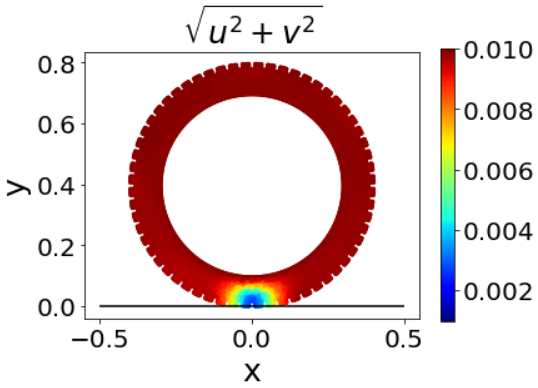}\label{pred_test2}}}
    \caption{\textit{Visualization of the displacement $||\bm u||$ for a tire in the training set (a)-(d), the first testing set (b)-(e), and the second testing set (c)-(f). The prediction is obtained by using GADEM with the VAE-Coord approach.}}%
    \label{visu_train_test}%
\end{figure}

\begin{figure}[H]
\centering
\includegraphics[width=8cm]{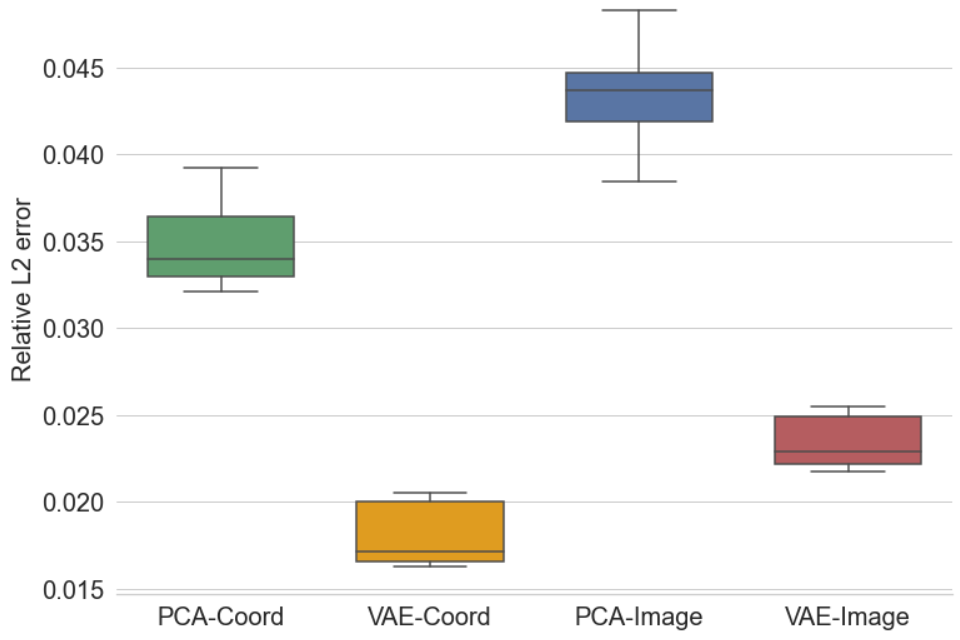}
\caption{\textit{Performance for all geometries in the training data set.}}
\label{perf_train_pneu_train}
\end{figure}

\begin{figure}[H]
    \centering
    \subfloat[Test 1]{{\includegraphics[width=8cm]{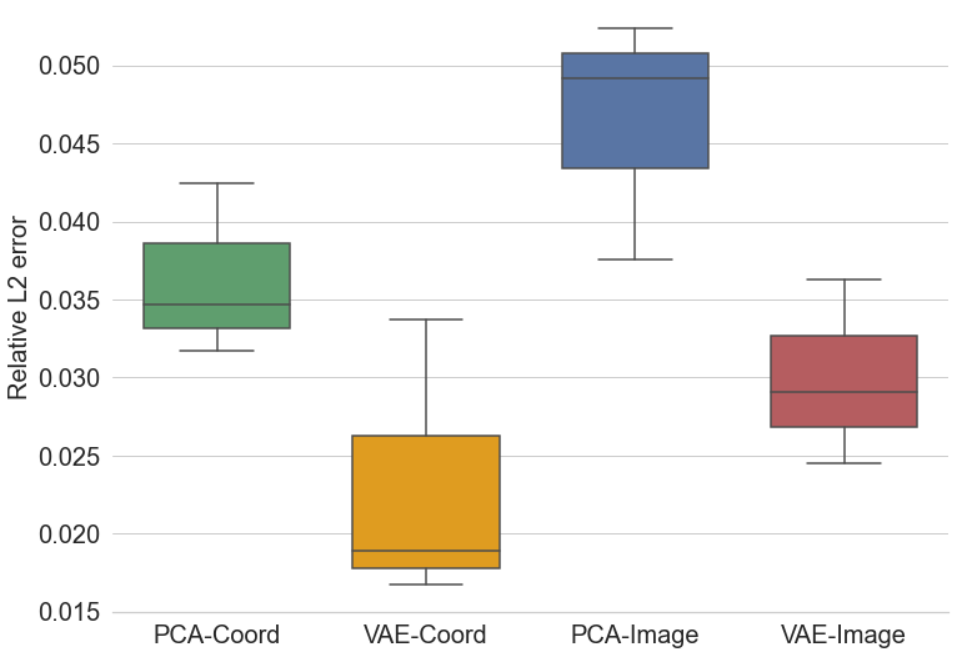}\label{boxplot_pneu_test}}}
    \quad
    \subfloat[Test 2 (OOD testing set)]{{\includegraphics[width=8cm]{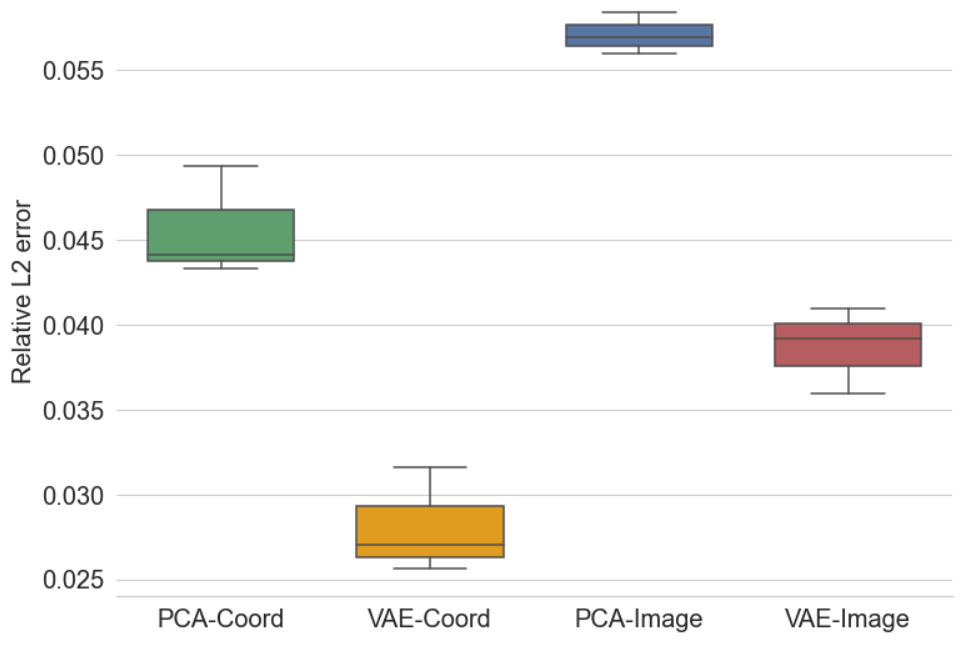}\label{boxplot_pneu_tes2}}}
    \caption{\textit{Performance on the testing data sets.}}%
    \label{pneu_boxplot_2d_train_test}%
\end{figure}

Figure \ref{pneu_visu_error} illustrates the errors produced by different approaches of GADEM for a tire in the training, the fist testing set, and the second testing set (ODD testing set). We see that the VAE-Coord provides the smallest errors for the displacement among all considered approaches. The prediction of the VAE-Coord approach for the displacement on different tires in the training and testing set is shown in Figure \ref{visu_train_test}.  We see that GADEM is able to predict accurately the solution for the displacement on different shapes of tire geometries. Figures \ref{perf_train_pneu_train} and \ref{pneu_boxplot_2d_train_test} show the accuracy of GADEM for all geometries in the training and testing sets. Again, the methods using spatial coordinates (PCA-Coord and VAE-Coord) outperform the methods using images to represent the geometries. The VAE's approaches provide better accuracy than the PCA's approaches. Among all methods, the VAE-Coord approach gives the best performance in terms of accuracy in both the training and testing sets. 

% Figure \ref{gadem_2d_pneu_loss} shows the loss function during the second training phase of all GADEM approaches. 

\subsubsection{Enhance the accuracy with FBOAL}
As the displacement of all the tires is only important when there is contact, we use FBOAL to infer the location of the training points to enhance the accuracy of GADEM at the contact zone. For the sake of comparison, we use the same initial training configuration as described in section \ref{sec_perf_tire}. For the training, on each geometry, we take $N_j=2500$ collocation points, which are randomly generated inside the domain, to minimize the energy. This leads to the total number of collocation points $N=44\times 2500=110000$. For the contact constraints, we take $N^j_c=600$ points on the potential contact boundary of each geometry, which leads to the number of training points on contact zone $N_c=6000$. For the training of FBOAL, we first train the network for $K=500000$ iterations, and then after every $k=100000$ iterations, we add and remove $m= 1\%\times N=1100$ collocation points based on the quantity to minimize on all the training geometries $\{\bm\Omega^j\}_{j=1}^\mathcal{G}$. We also add and remove $m_c= 1\%\times N_C=600$ points based on the quantity to minimize on all potential contact zones $\{\bm\Omega^j_C\}_{j=1}^\mathcal{G}$. Again, the values of these hyperparameters are chosen based on the conclusion of the work of \cite{nguyen2023fixed}.

\begin{figure}[H]
    \centering
    \subfloat[Loss function]{{\includegraphics[width=7cm]{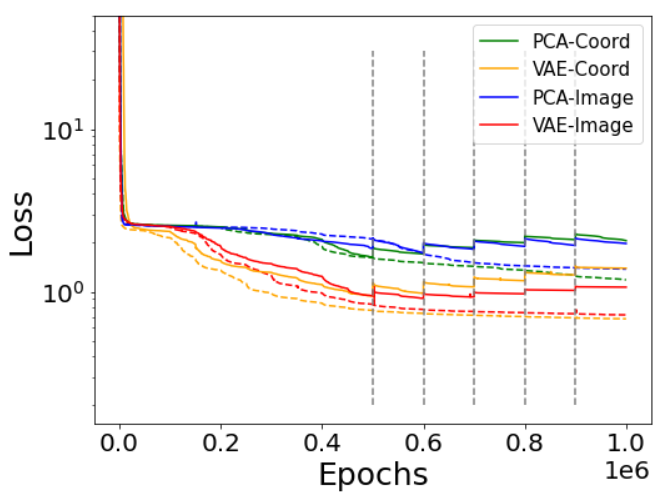}\label{pneu_ref_train_loss}}}
    \subfloat[Error on testing points]{{\includegraphics[width=7cm]{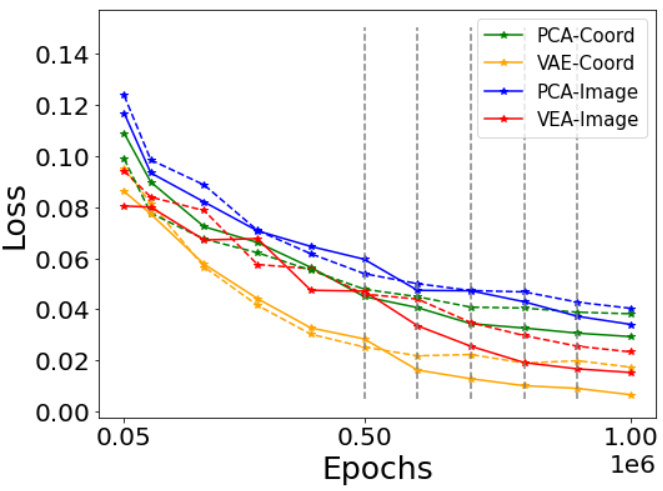}\label{pneu_ref_test1_loss}}}
    \caption{\textit{Loss function during the training process and the relative $\mathcal{L}^2$ errors on the testing points (average results of all geometries) produced by GADEM without FBOAL (dashed lines) and GADEM with FBOAL (solid lines). The gray lines indicate the instances where FBOAL effectuates the resampling.}}%
    \label{visu_pneu_all_loss}%
\end{figure}

Figure \ref{visu_pneu_all_loss} shows the loss function during the training process and the relative $\mathcal{L}^2$ errors on the testing points produced by GADEM with and without FBOAL. Again, when FBOAL effectuates the resampling, we observe the jumps in the loss function. Thus the loss functions produced by GADEM with FBOAL are less well minimized than the loss produced by GADEM without FBOAL (Figure \ref{pneu_ref_train_loss}). However, the errors of the prediction on the testing points are significantly reduced by using GADEM with FBOAL (Figure \ref{pneu_ref_test1_loss}).

\begin{table}[H]
\centering
\begin{tabular}{|c|c|c|c|c|c|}
\hline
 & & PCA-Coord & VAE-Coord & PCA-Image & VAE-Image \\
\hline
Train & W/o FBOAL & 3.56e-2$\pm$2.28e-3 & \textbf{1.95e-2}$\pm$\textbf{1.17e-3} & 4.43e-2$\pm$2.55e-3 & \textbf{2.33e-2}$\pm$\textbf{1.35e-3}\\
      & With FBOAL & \textbf{1.41e-2}$\pm$\textbf{4.19e-3} & \textbf{1.02e-2}$\pm$\textbf{3.76e-3} & 3.13e-2$\pm$5.55e-3 & \textbf{1.63e-2}$\pm$\textbf{5.31e-3} \\
\hline
Test 1 & W/o FBOAL & 3.63e-2$\pm$4.55e-3 & \textbf{2.31e-2}$\pm$\textbf{7.55e-3} & 4.63e-2$\pm$8.89e-3 & \textbf{2.99e-2}$\pm$\textbf{4.41e-3} \\
      & With FBOAL & \textbf{2.66e-2}$\pm$\textbf{5.27e-3} & \textbf{1.94e-2}$\pm$\textbf{4.98e-3} & 3.44e-2$\pm$8.06e-3 & \textbf{2.17e-2}$\pm$\textbf{4.65e-3}\\
\hline
Test 2 & W/o FBOAL & 4.55e-2$\pm$2.03e-3 & \textbf{2.80e-2}$\pm$\textbf{1.67e-3} & 5.71e-2$\pm$1.24e-3 & 3.86e-2$\pm$1.19e-3 \\
      & With FBOAL & 4.28e-2$\pm$2.11e-3 & \textbf{2.49e-2}$\pm$\textbf{1.75e-3} & 5.56e-2$\pm$2.06e-3 & 3.49e-2$\pm$1.74e-3\\
\hline
\end{tabular}
\caption{\textit{Relative $\mathcal{L}^2$ errors of all the approaches with and without using FBOAL.}}
\label{error_pneu_fboal}
\end{table}

\begin{figure}[H]
    \centering
    \subfloat[W/o FBOAL]{{\includegraphics[width=5.5cm]{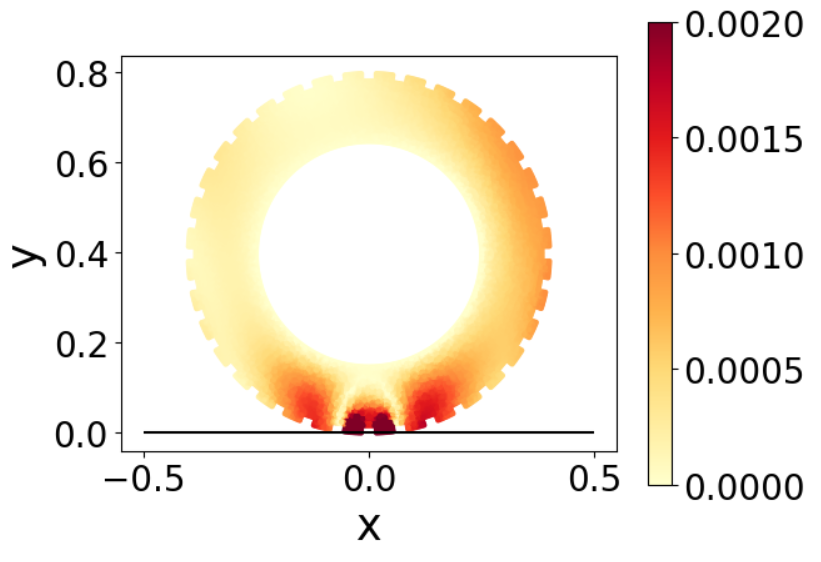}\label{fboal_error_wo_1}}}
    \subfloat[W/o FBOAL]{{\includegraphics[width=5.8cm]{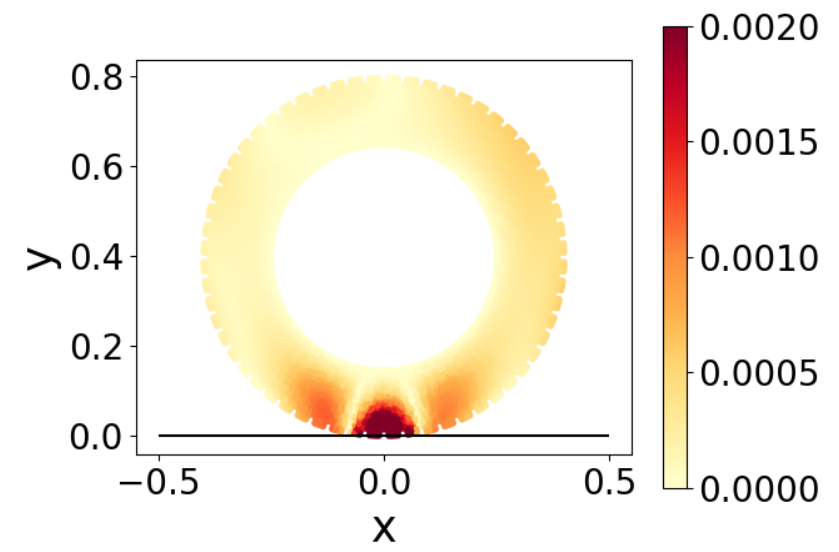}\label{fboal_error_wo_2}}}
    \subfloat[W/o FBOAL]{{\includegraphics[width=5.5cm]{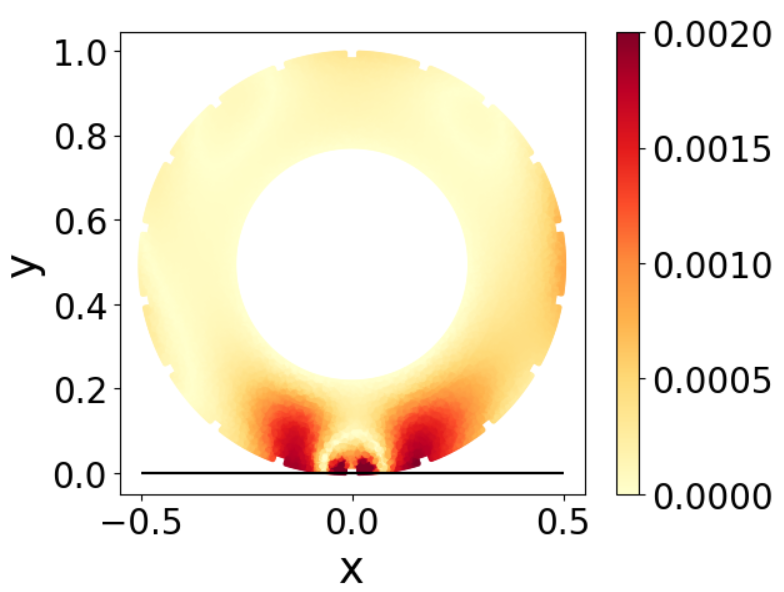}\label{fboal_error_wo_3}}}
    \quad
    \subfloat[With FBOAL]{{\includegraphics[width=5.5cm]{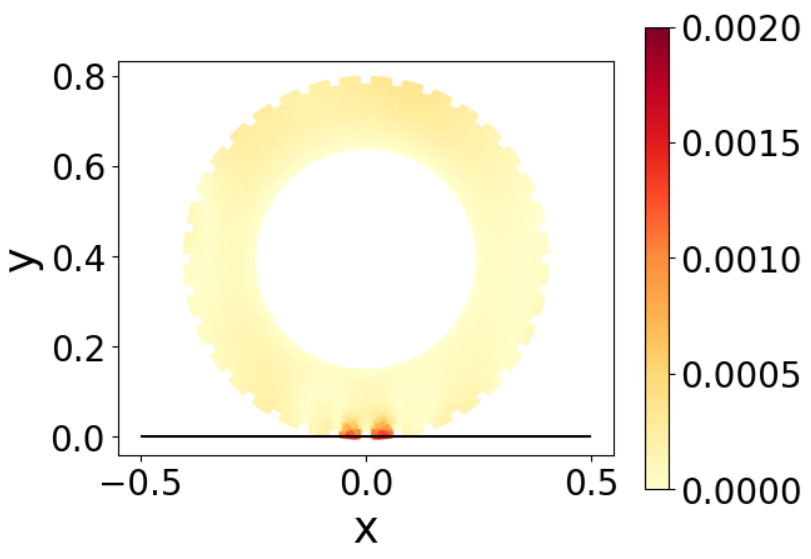}\label{fboal_error_w_1}}}
    \subfloat[With FBOAL]{{\includegraphics[width=5.7cm]{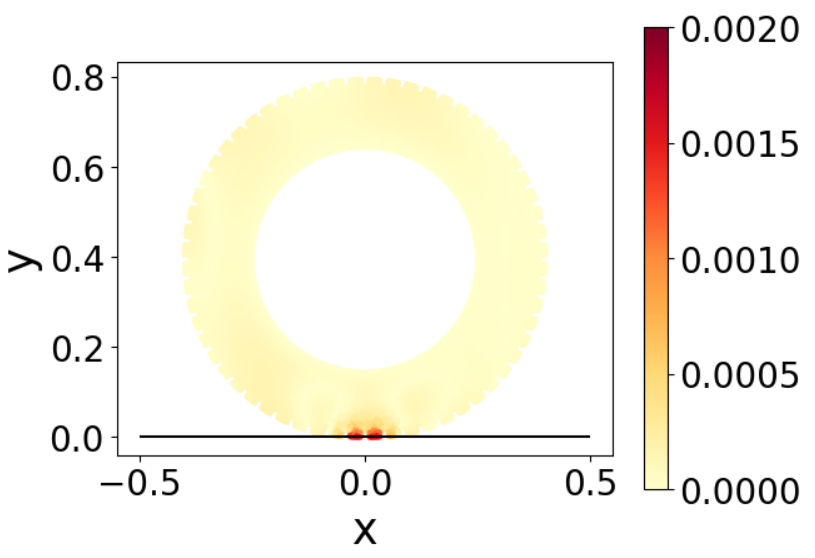}\label{fboal_error_w_2}}}
    \subfloat[With FBOAL]{{\includegraphics[width=5.5cm]{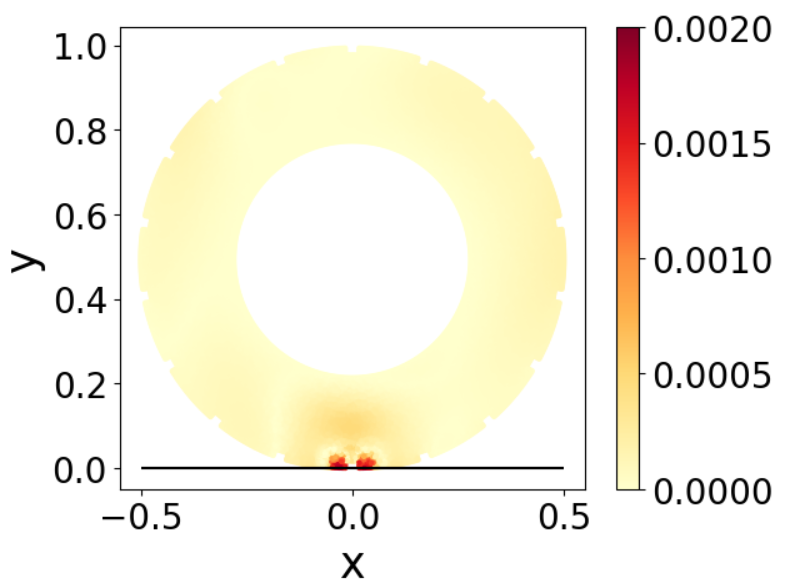}\label{fboal_error_w_3}}}
    \quad
    \subfloat[Density by FBOAL]{{\includegraphics[width=5.3cm]{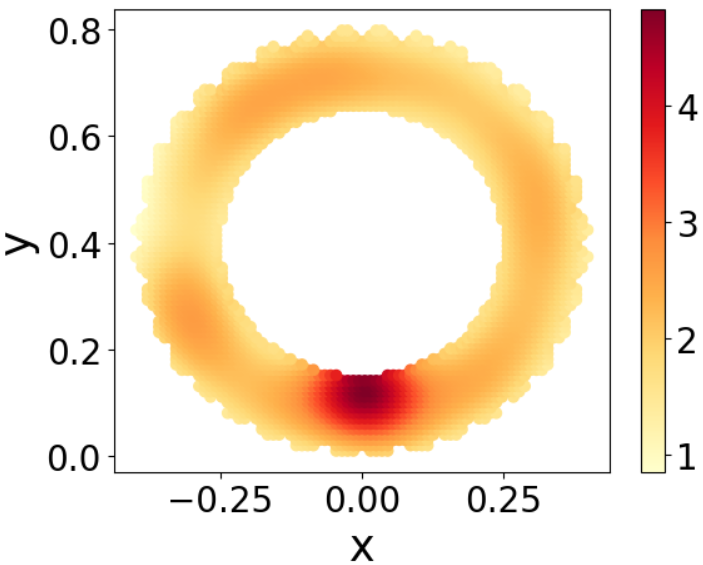}\label{fboal_dens_1}}}
    \subfloat[Density by FBOAL]{{\includegraphics[width=5.3cm]{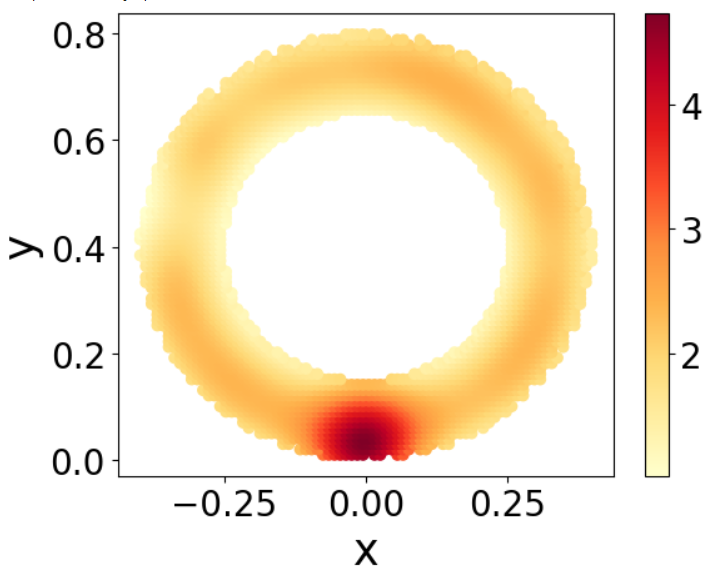}\label{fboal_dens_2}}}
    \subfloat[Density by FBOAL]{{\includegraphics[width=5.5cm]{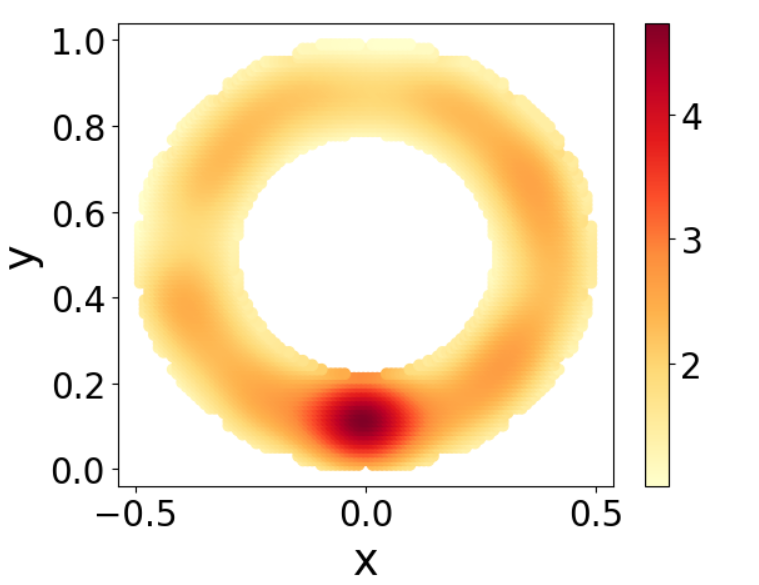}\label{fboal_dens_3}}}
    \caption{\textit{Visualization of the absolute errors produced by GADEM with the PCA-Coord approaches without using FBOAL (a-c), with using FBOAL on the corresponding tires (d-f), and the density of collocation points after the training with FBOAL on the corresponding tires (g-i).}}%
    \label{fboal_error}%
\end{figure}

Figure \ref{fboal_error} visualizes the errors with and without FBOAL, the density of collocation points after the training with FBOAL on different training geometries. We see that FBOAL adds many points close to the contact positions for all geometries. This is reasonable as the potential energy is important in this area. The errors near the contact area are significantly reduced by using FBOAL. Table \ref{error_pneu_fboal} shows the relative $\mathcal{L}^2$ errors (average results of all the geometries) between the reference solution and the prediction of GADEM approaches. The good results (errors smaller than $3\%$) are highlighted. Again, with the adaptive learning strategy FBOAL, we gain a huge accuracy compared to the classical approaches that fix the collocation points for the training data set. The same conclusion can be drawn for the testing sets.

\section{Conclusion}
In this study, we introduced the Geometry-Aware Deep Energy Method (GADEM) based on PINNs for the modeling of structural mechanics problems, which is capable of inferring the solution on different shapes of geometries. To represent the geometries, either boundary spatial coordinates or object images can be used. In the first phase of training, GADEM employs either parametric encoding or dimensionality reduction methods (PCA or VAE) to extract the geometric latent vectors. In the second phase, these latent vectors are considered as additional inputs of the model, and we minimize the potential energy of the systems over all the geometries in the loss function. We validate our proposed framework in an academic test case involving linear elasticity with various types of geometries. We then investigate the use of GADEM in a more realistic industrial problem of tire loading simulation. As far as our knowledge, this is the first time a physics-informed deep learning framework has been successfully applied to solve a contact mechanics problem involving hyperelastic materials. Through this work, GADEM has shown a great capability to predict the solution for different shapes of geometries using only one trained model. Consistently, using spatial coordinates to represent the geometries provides better accuracy than using images, and using the VAE for the encoding provides more accurate results when dealing with complex geometries than using the PCA. We also employed an adaptive learning strategy to infer the best location for the training collocation points in GADEM. With this method, it has been shown that the accuracy of GADEM can be further improved. Besides that, the idea of GADEM can be explored for solving other problems involving PDEs on different geometries.

It can be seen that there are many challenges and opportunities for further research. In this work, we only consider the cases where the boundary conditions are the same for all geometries. In the case where different boundary conditions are applied to different geometries, GADEM can also be employed. However, there will be a need to incorporate the geometries and boundary conditions to improve the performance of the model. Furthermore, one could employ GADEM for new design generation, that is, VAE could be used to generate new latent vectors corresponding to new geometries, and GADEM could be used to predict the solution on these whole new geometries. We aim to purchase these lines of research for future studies. 

\section*{Acknowledgements}
T.N.K. Nguyen is funded by Michelin and CEA through the Industrial Data Analytics and Machine Learning chair of ENS Paris-Saclay. 

% \textcolor{red}{TIBO: j'aimerais vous proposer de rajouter Jean Di Stasio à la liste des co-auteurs de cet article pour le remercier de l'aide qu'il a apporté à Khoa et pour profiter aussi de sa relecture de la partie où le problème mécanique est décrit (section 2 principalement)}. \textcolor{blue}{Khoa: Oui je suis d'accord.}

\appendix
\section{About the of the PCA and VAE}\label{append_rec}
\subsection{Configuration and architecture}\label{append_rec_1}
The architecture of the VAE models used in this work is described as follows:
\begin{itemize}
    \item For the VAE on the spatial coordinates: in the encoder, we use five 1D convolutional layers of size $[32,64,128,256,512]$, followed by three fully connected layers of size $[256,256,256]$ with ReLU activation. In the decoder, we use three fully connected layers of size $[256,256,256]$. 
    \item For the VAE on the images: in the encoder, we use two 2D convolutional layers in sequential order, ReLU activation with 3x3 kernel sizes, and 32, 64 and filters, followed by two fully connected layers of size $[16,5]$ (where $5$ represents the size of latent vector). In the decoder, 
\end{itemize}
To minimize the loss function in the VAE, we adopt Adam optimizer with the learning rate decay strategy, which is proven to be very efficient in training deep learning models \citep{you2019does}. The results are obtained with $1,000$ epochs with the learning rate $lr=10^{-3}$, followed by $1,000$ epochs with the learning rate $lr=10^{-4}$ and $1,000$ epochs with $lr=10^{-5}$.

% The output is then split into the prediction of mean and variance the posterior distribution (which is in our case, a standard Gaussian distribution).

\subsection{Performance on the academic case}\label{append_rec_2}
\begin{figure}[H]
    \centering
    \subfloat[On spatial coordinates]{{\includegraphics[width=4.5cm]{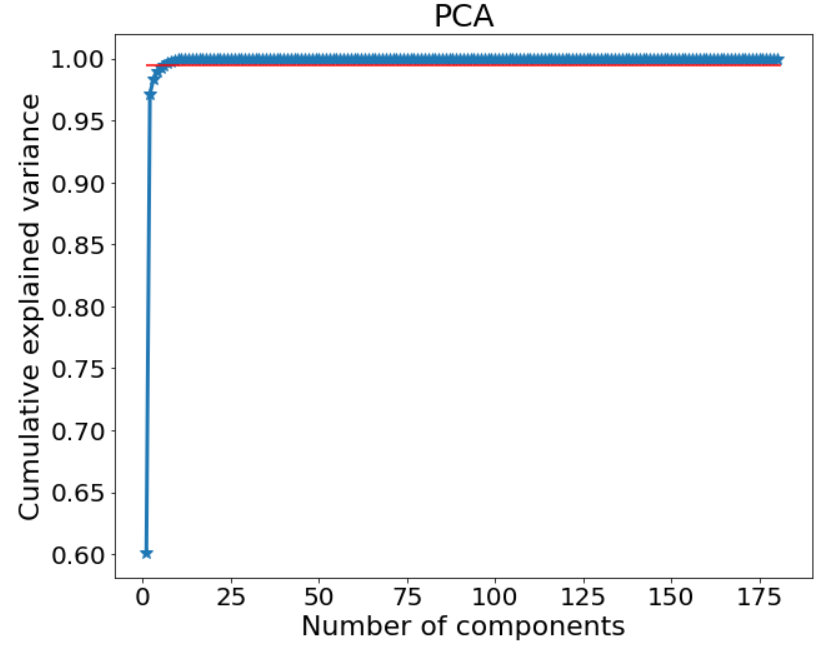} \label{pca_cum}}}%%
    \subfloat[On spatial coordinates]{{\includegraphics[width=4.5cm]{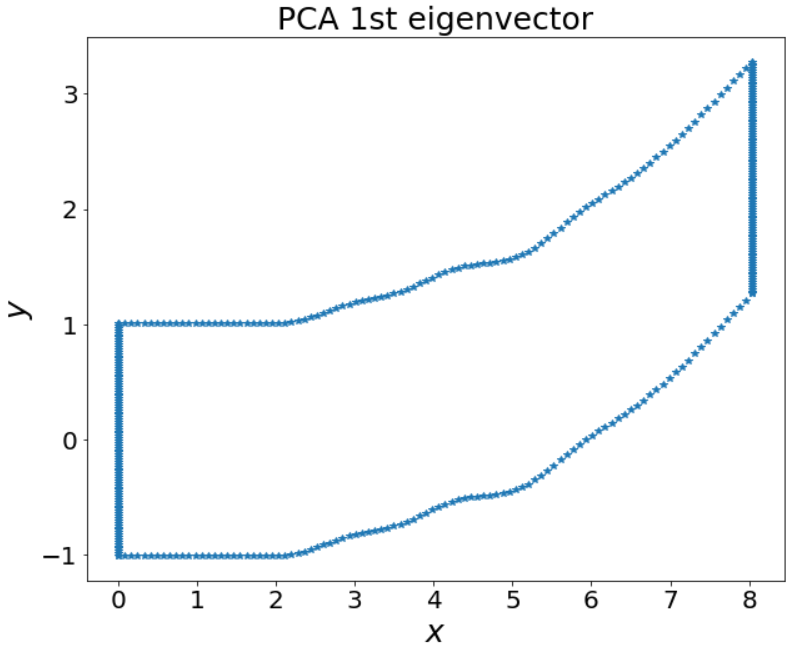}\label{pca_axe1}}}
    \subfloat[On images]{{\includegraphics[width=4.5cm]{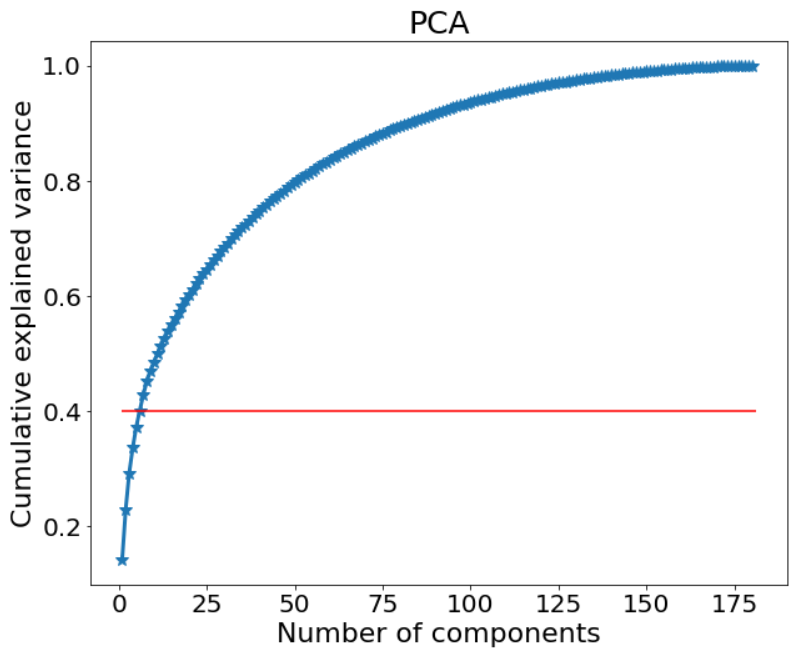}\label{pca_cum_image}}}
    \subfloat[On images]{{\includegraphics[width=4.5cm]{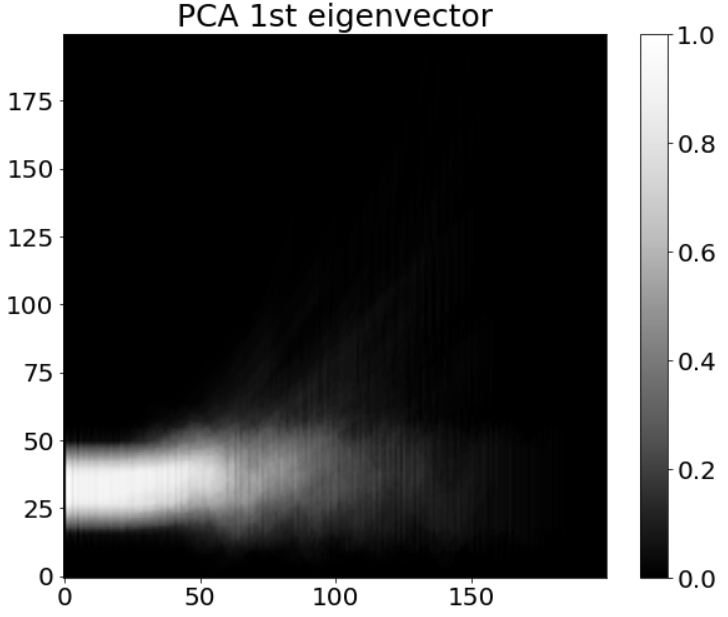}\label{pca_axe1_image}}}
    \caption{\textit{Visualization for the PCA: Cumulative explained variance and first eigenvector.}}%
    \label{reduction_pca}%
\end{figure}

\begin{figure}[H]
    \centering
    \subfloat[On spatial coordinates]{{\includegraphics[width=5cm]{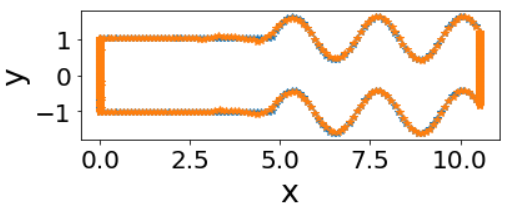} \label{vae_2d_train}}}%%
    \subfloat[On spatial coordinates]{{\includegraphics[width=5cm]{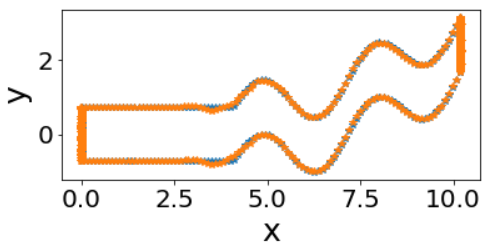}\label{vae_2d_test1}}}
    \subfloat[On spatial coordinates]{{\includegraphics[width=3cm]{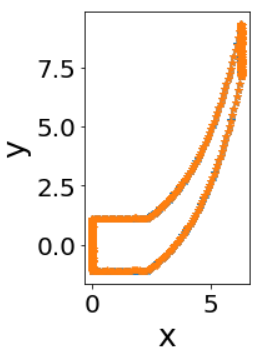}\label{vae_2d_test2}}}
    \quad
    \subfloat[On images]{{\includegraphics[width=5cm]{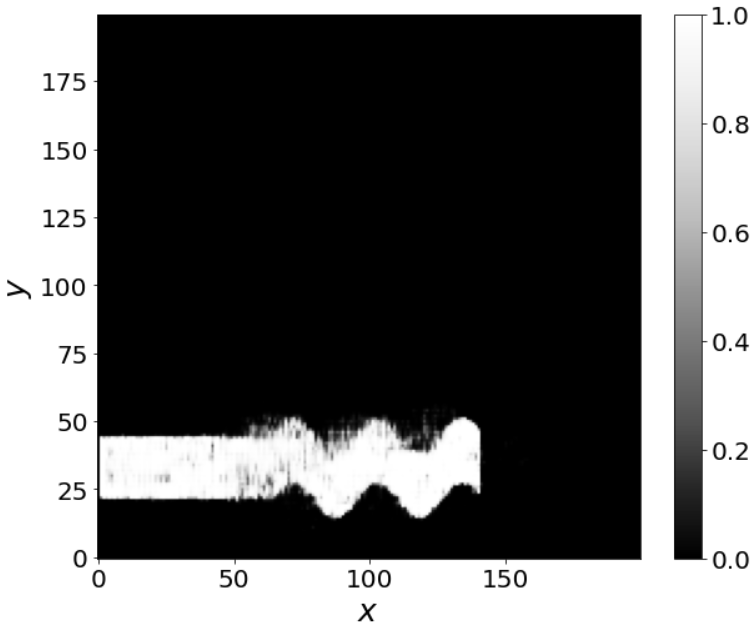} \label{vae_2d_train_image}}}%%
    \subfloat[On images]{{\includegraphics[width=5cm]{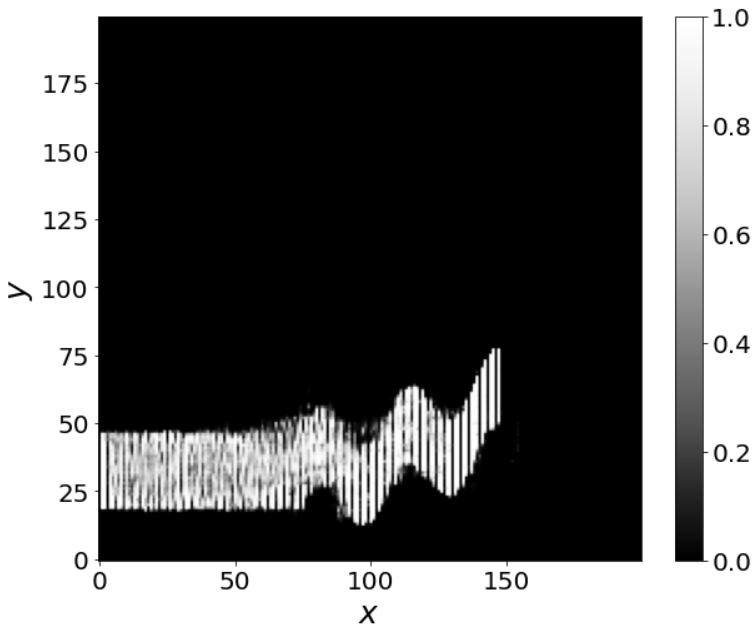}\label{vae_2d_test1_image}}}
    \subfloat[On images]{{\includegraphics[width=5cm]{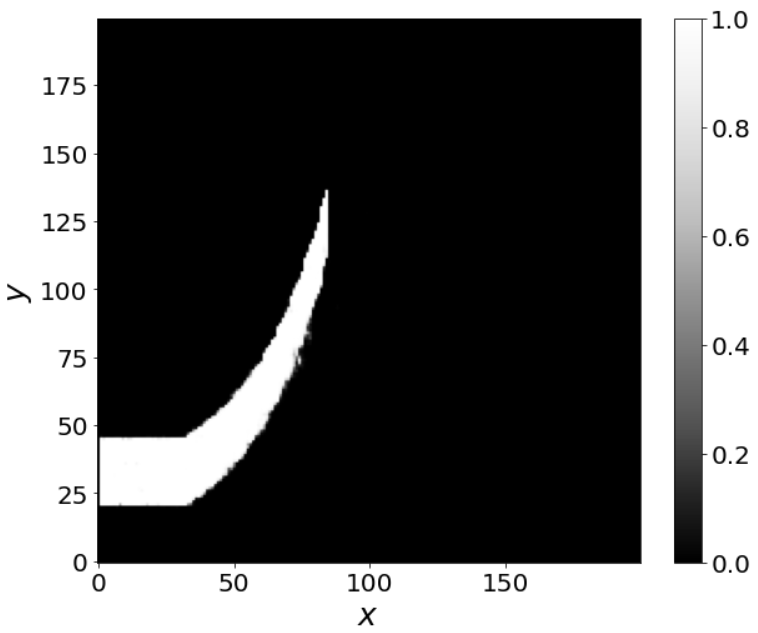}\label{vae_2d_test2_image}}}
    \caption{\textit{Reconstruction of the VAE for different beams.}}%
    \label{reduction_vae}%
\end{figure}
\begin{figure}[H]
    \centering
    \subfloat[Parametric]{{\includegraphics[width=5cm]{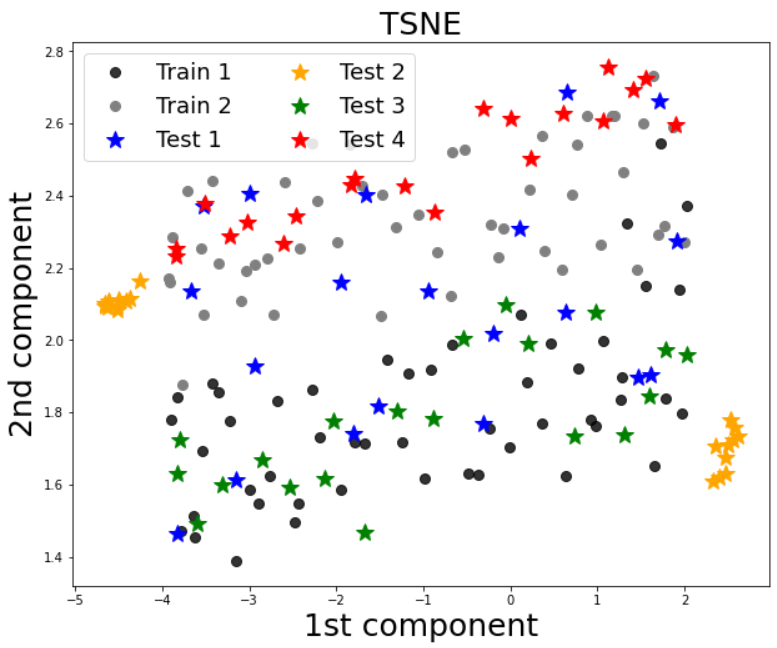} \label{tsne_params}}}%%
    \quad
    \subfloat[PCA-Coord]{{\includegraphics[width=5cm]{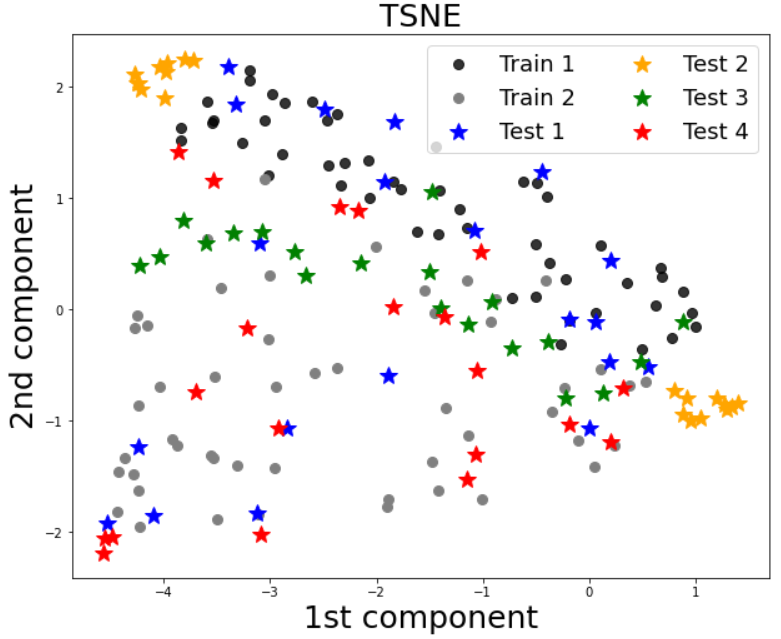}\label{tsne_pca}}}
    \quad
    \subfloat[VAE-Coord]{{\includegraphics[width=5cm]{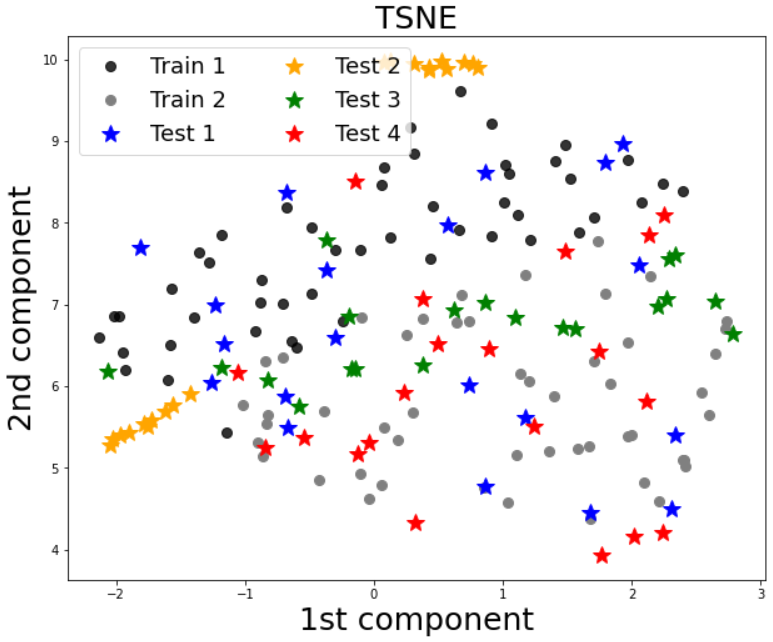}\label{tsne_vae}}}
    \quad
    \subfloat[PCA-Image]{{\includegraphics[width=5cm]{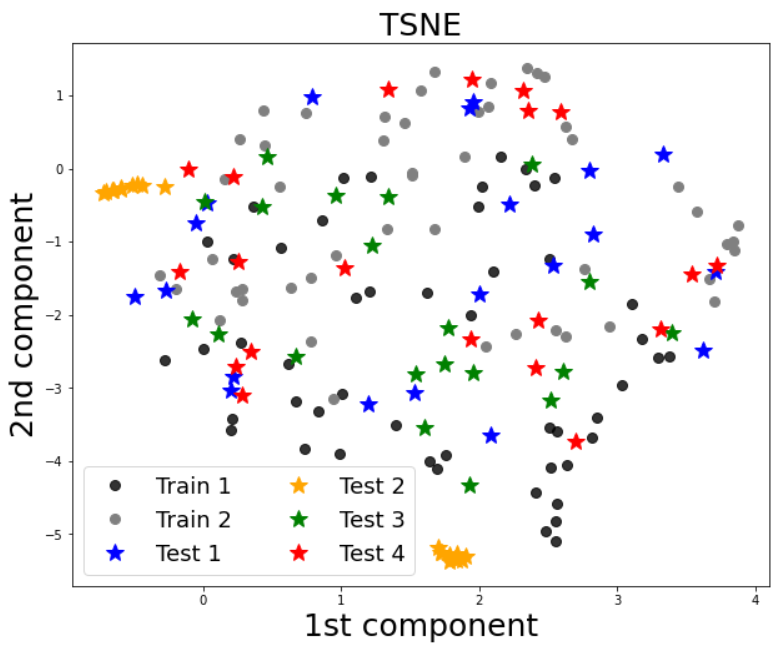}\label{tsne_image_pca}}}
    \quad
    \subfloat[VAE-Image]{{\includegraphics[width=5cm]{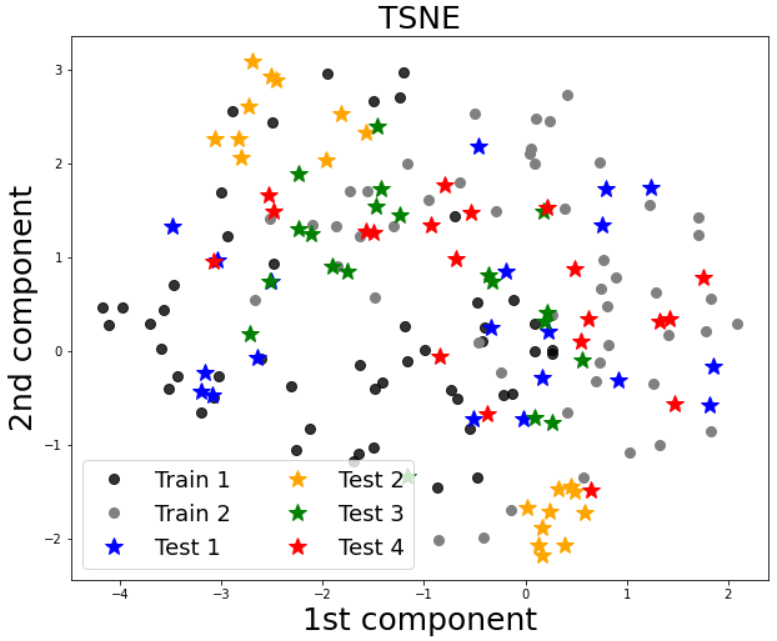}\label{tsne_image_vae}}}
    \caption{\textit{Visualization of the latent vectors in the reduced space by TSNE.}}%
    \label{tsne_2d}%
\end{figure}

To visualize the geometric latent vectors in a 2D space, we employ the TSNE technique (t-distributed Stochastic Neighbor Embedding \citep{van2008visualizing}), which is a nonlinear dimensionality reduction method for embedding high-dimensional data in a 2D or 3D space for visualization. Figure \ref{tsne_2d} shows the latent vectors in the reduced TSNE space. For the parametric encoding (Figure \ref{tsne_params}), as this method only encodes the parameters $l, d, l_1,p_1, p_2$, not the sinus, polynomial, or exponential information, we see that this approach only distinguishes well the training and testing sets when the learning intervals of the parameters are different (e.g. the latent vector of the second test located differently from other sets). For the PCA-Coord and VAE-Coord approaches (Figures \ref{tsne_pca} and \ref{tsne_vae}), we see that the latent vectors of the first and second training sets are well-distinguished. The latent vectors of the first test (where the geometries have the same shapes and learning intervals as the training set) are located within the space between the training sets, while the second test (where the geometries have the same shapes but different learning intervals as the training set) are located outside this space. For the third set, as the geometry shapes in this set are a combination of the two training sets, the latent vector of this set lies between the space of the first training set and the second training set. For the fourth test, although this test set composes new geometries of exponential shapes, these geometries have very similar shapes to the polynomial training set (second training set). Thus we see that the latent vector of this set lies close to the second training set. A similar interpretation can be obtained for the PCA-Image and VAE-Image approaches (Figures \ref{tsne_image_pca} and \ref{tsne_image_vae}), however, the difference between the training and testing sets can not clearly observed. 

\subsection{Performance on the tire loading simulation}\label{append_rec_3}
\begin{figure}[H]
    \centering
    \subfloat[On spatial coordinates]{{\includegraphics[width=4.5cm]{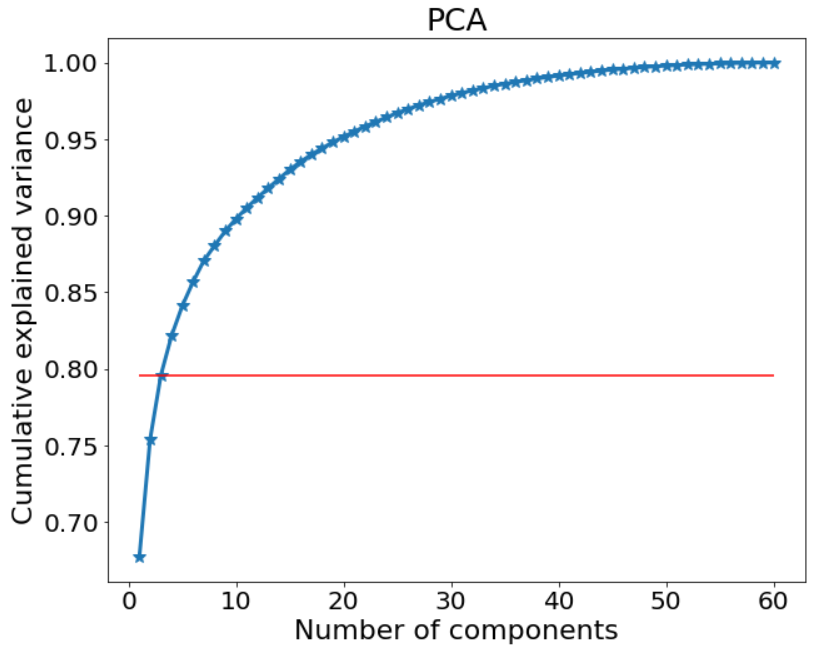} \label{pca_cum_pneu}}}%%
    \subfloat[On spatial coordinates]{{\includegraphics[width=4.5cm]{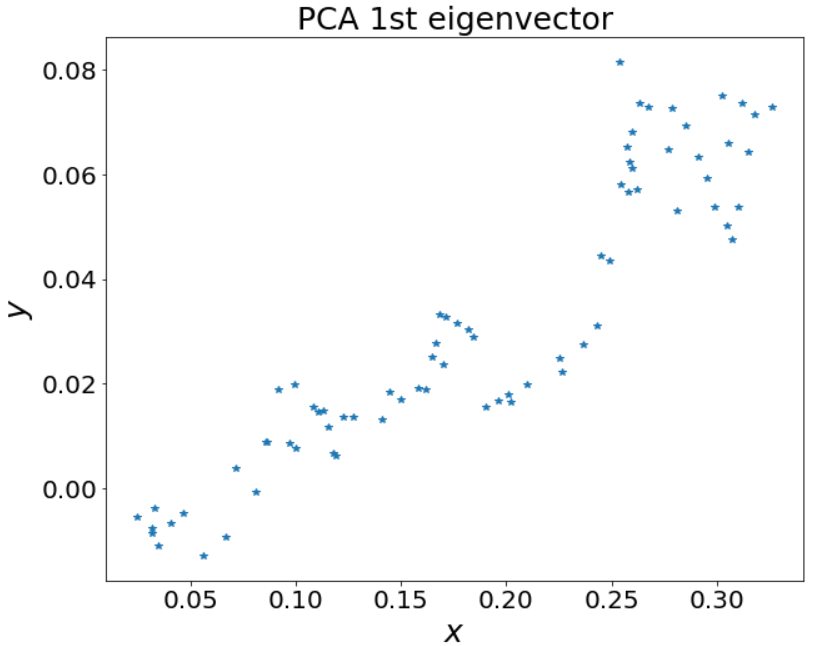}\label{pca_axe1_pneu}}}
    \subfloat[On images]{{\includegraphics[width=4.5cm]{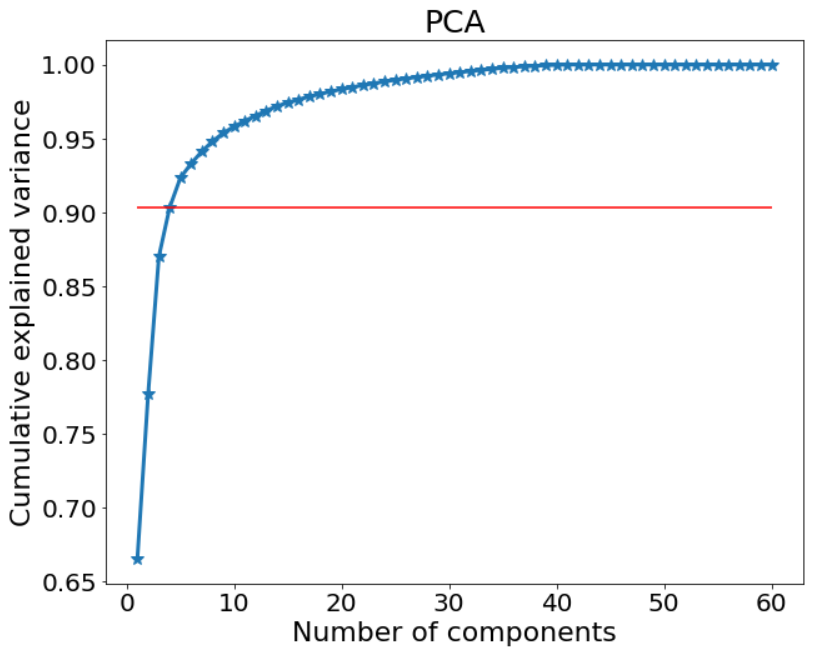}\label{pca_cum_image_pneu}}}
    \subfloat[On images]{{\includegraphics[width=4.5cm]{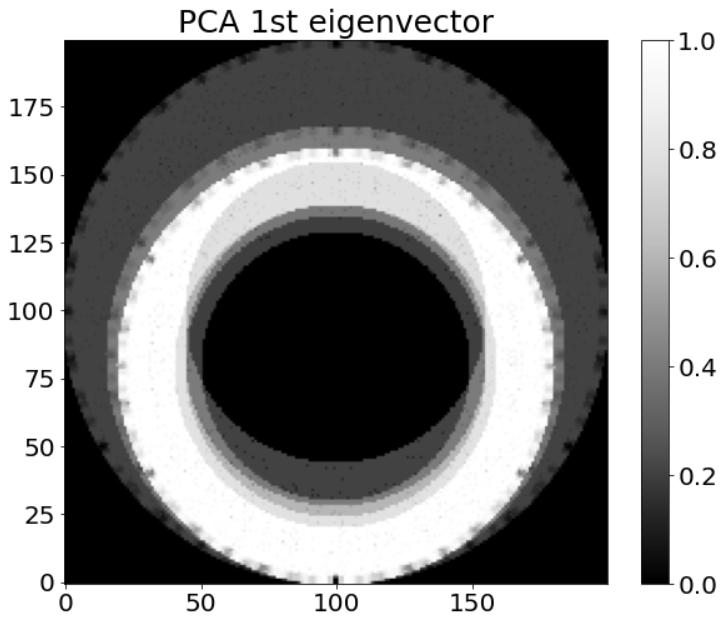}\label{pca_axe1_image_pneu}}}
    \caption{\textit{Visualization for the PCA: Cumulative explained variance and first eigenvector.}}%
    \label{reduction_pca_pneu}%
\end{figure}

\begin{figure}[H]
    \centering
    \subfloat[On spatial coordinates]{{\includegraphics[width=5cm]{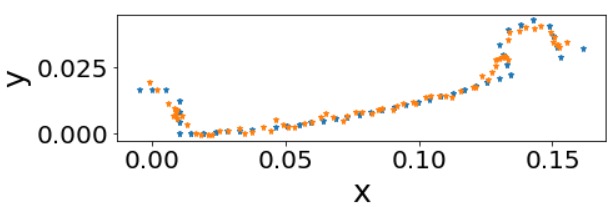} \label{vae_2d_train_pneu}}}%%
    \subfloat[On spatial coordinates]{{\includegraphics[width=5cm]{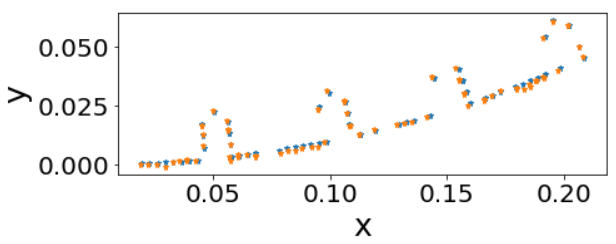}\label{vae_2d_test1_pneu}}}
    \subfloat[On spatial coordinates]{{\includegraphics[width=5cm]{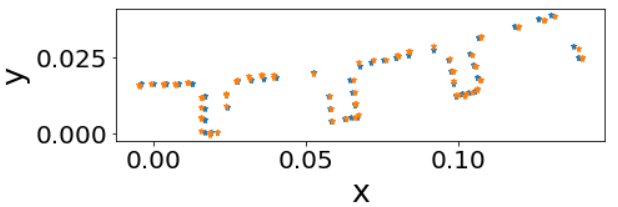}\label{vae_2d_test2_pneu}}}
    \quad
    \subfloat[On images]{{\includegraphics[width=5cm]{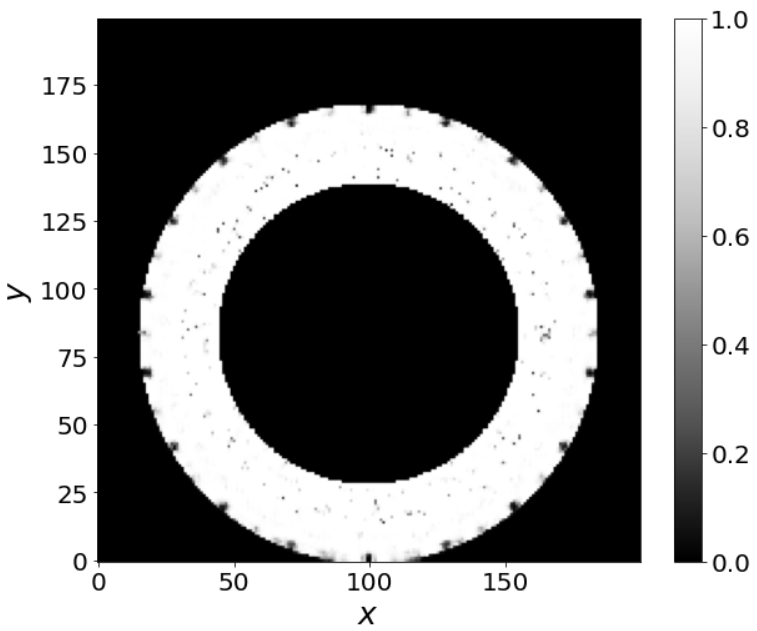} \label{vae_2d_train_image_pneu}}}%%
    \subfloat[On images]{{\includegraphics[width=5cm]{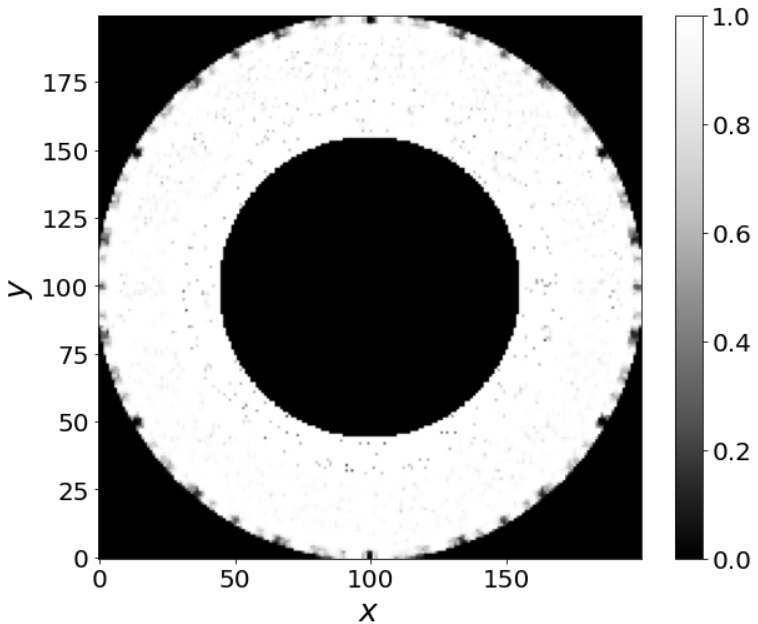}\label{vae_2d_test1_image_pneu}}}
    \subfloat[On images]{{\includegraphics[width=5cm]{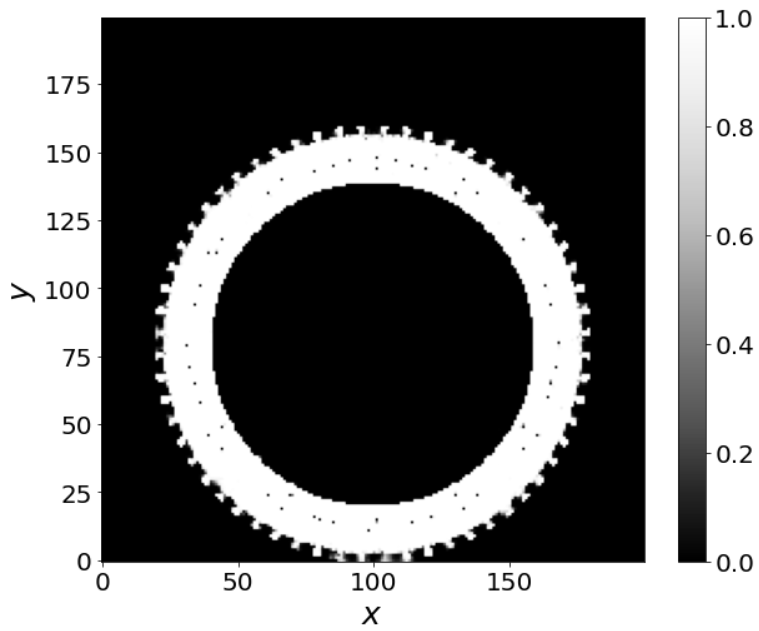}\label{vae_2d_test2_image_pneu}}}
    \caption{\textit{Reconstruction of the VAE for different tires.}}%
    \label{reduction_vae_pneu}%
\end{figure}

\begin{figure}[H]
    \centering
    \subfloat[PCA-Coord]{{\includegraphics[width=5.5cm]{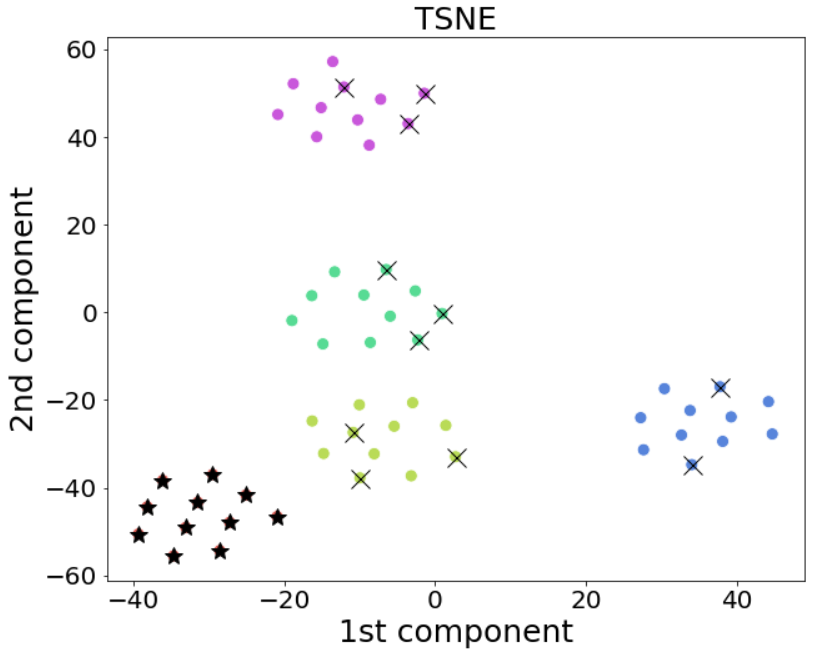} \label{pneu_tsne_pca}}}%%
    \quad
    \subfloat[VAE-Coord]{{\includegraphics[width=5.5cm]{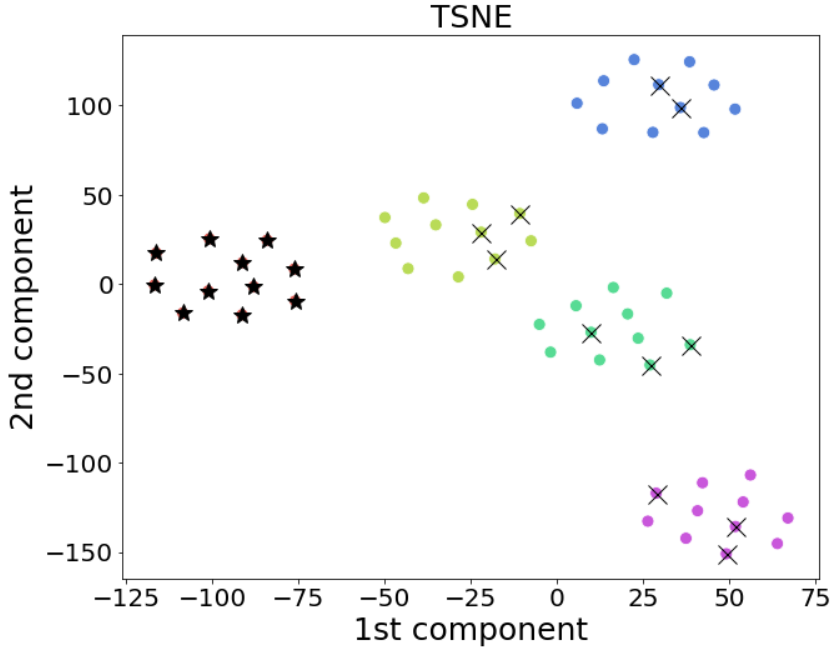}\label{pneu_tsne_vae}}}
    \quad
     \subfloat[PCA-Image]{{\includegraphics[width=5.5cm]{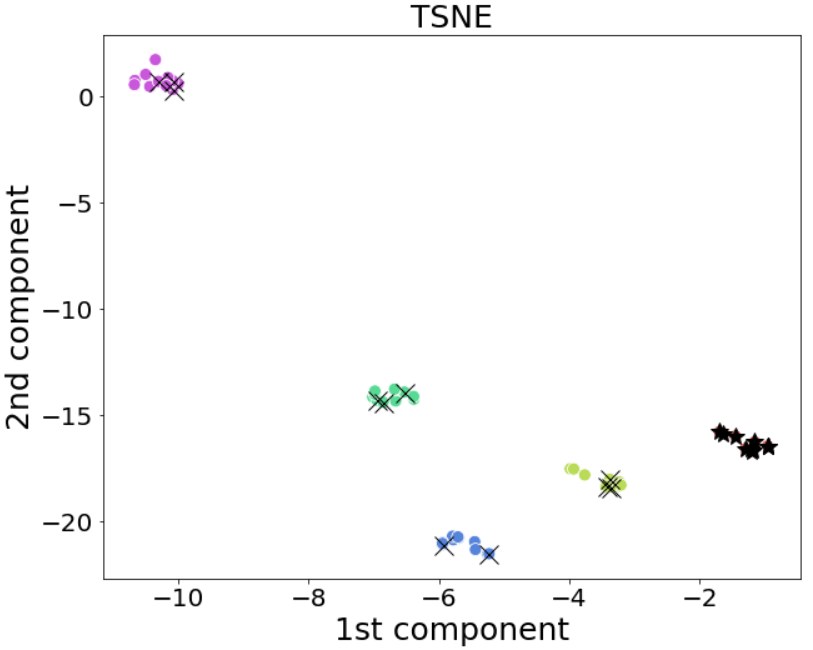}\label{pneu_tsne_pca_image}}}%%
     \quad
      \subfloat[VAE-Image]{{\includegraphics[width=5.5cm]{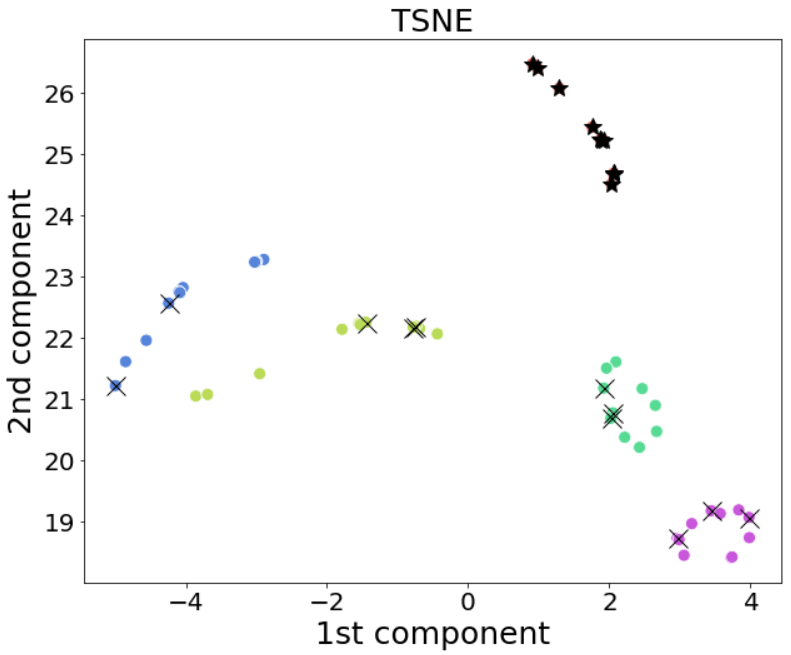}\label{pneu_tsne_vae_image}}}%%
    \caption{\textit{Visualization of latent vectors with $k_{\bm z}=5$ of different dimensionality reduction methods by T-SNE technique.} Each color represents each main group of geometries. The '$\times$' symbol denotes the tires on the first testing data set. The '$\star$' symbol denotes the tires on the second testing data set.}%
    \label{reduction}%
\end{figure}
Figure \ref{reduction} illustrates the latent vectors of the PCA and VAE with $k_{\bm z}=5$ in reduced space produced by the T-SNE for the visualization. We see that the latent vectors of five main classes of tire geometries are grouped well by all reduction methods, and the differences between each class are well-observed. 
% \textcolor{red}{TIBO: j'ai la même remarque ici, je ne sais pas quoi dire de ces visus T-SNE...}

\section{Results in tire loading simulation when choosing another OOD testing set}\label{append_ood_test}

We provide the error of GADEM approaches when choosing another group rather than the fifth group as the OOD testing set. The following results are obtained with the vanilla GADEM (\textit{i.e.} without using FBOAL to enhance the accuracy). 

\begin{figure}[H]
    \centering
    \subfloat[Train set]{{\includegraphics[width=5.5cm]{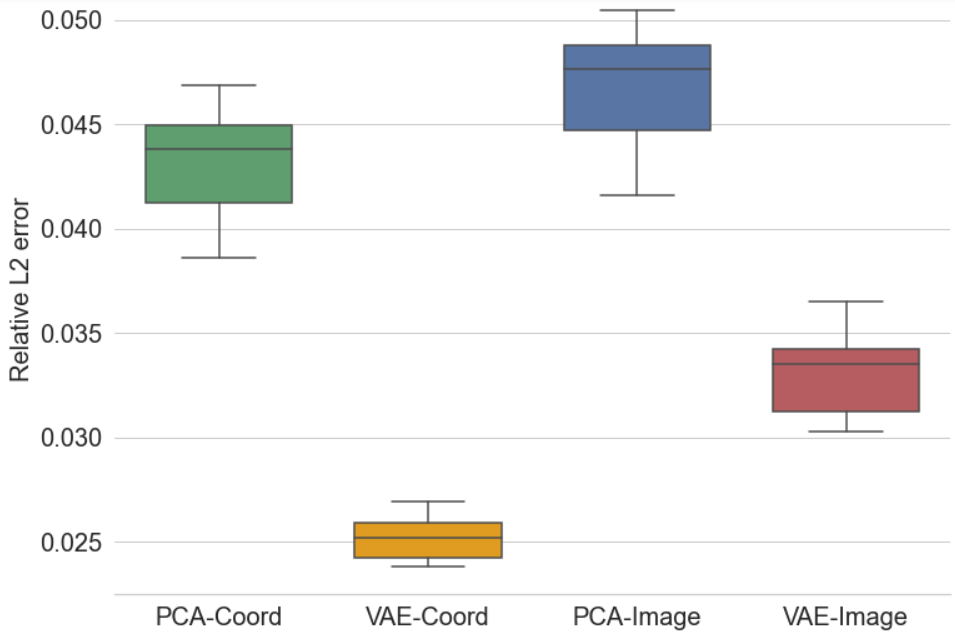} \label{ood_train_1}}}%%
    % \quad
    \subfloat[Test 1]{{\includegraphics[width=5.5cm]{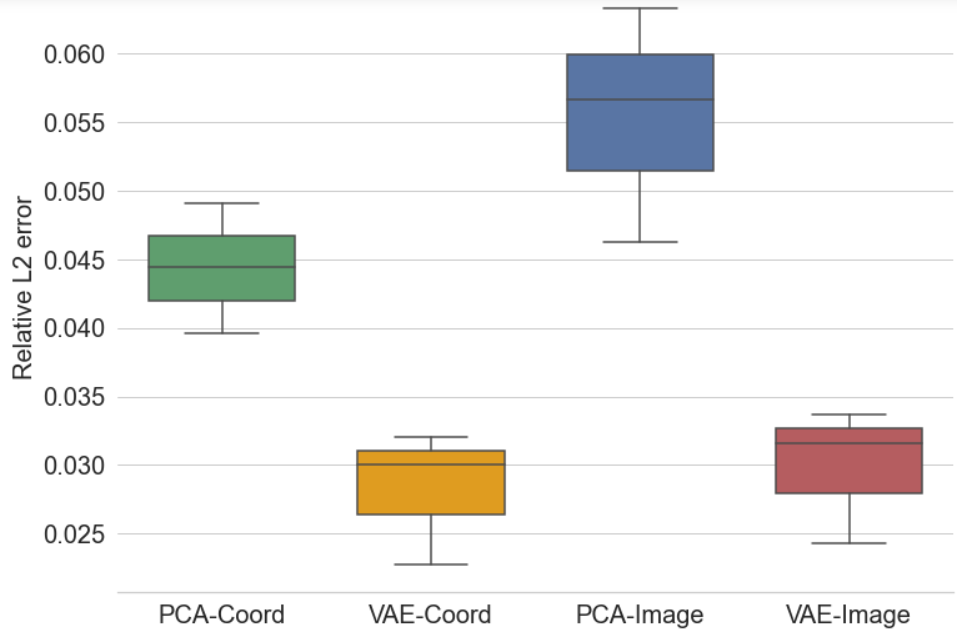}\label{ood_test1_1}}}
    % \quad
     \subfloat[Test 2]{{\includegraphics[width=5.5cm]{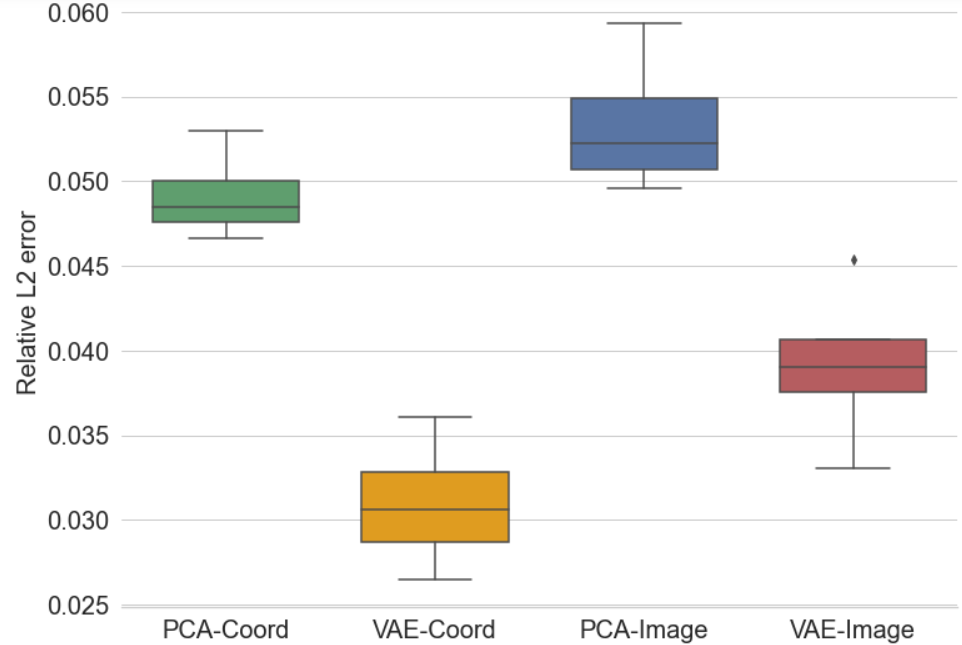}\label{ood_test2_1}}}%%
    \caption{\textit{Group 1 as OOD set: Performance on training and testing data sets.}}%
    \label{ood_1}%
\end{figure}

\begin{figure}[H]
    \centering
    \subfloat[Train set]{{\includegraphics[width=5.5cm]{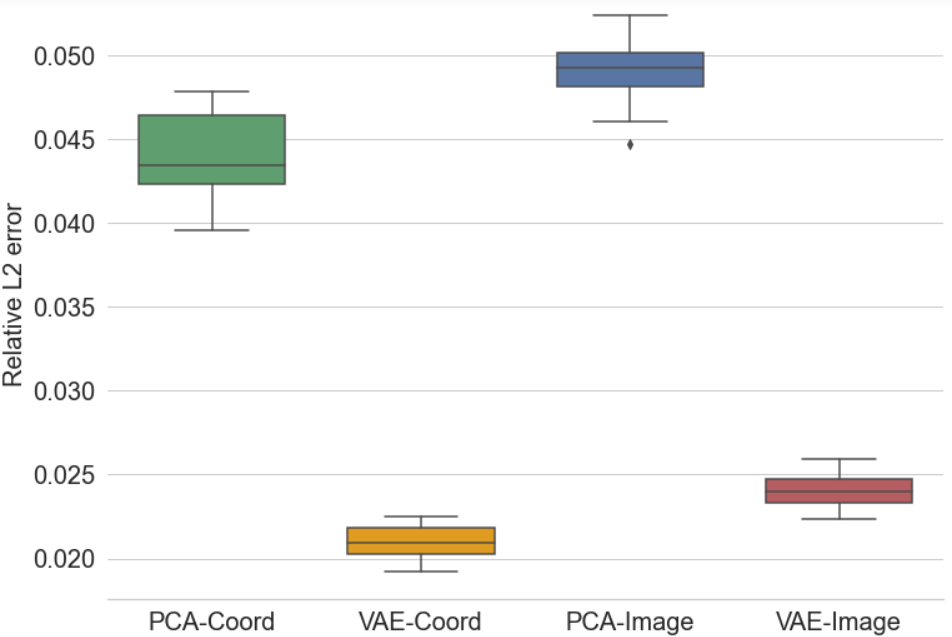} \label{ood_train_2}}}%%
    % \quad
    \subfloat[Test 1]{{\includegraphics[width=5.5cm]{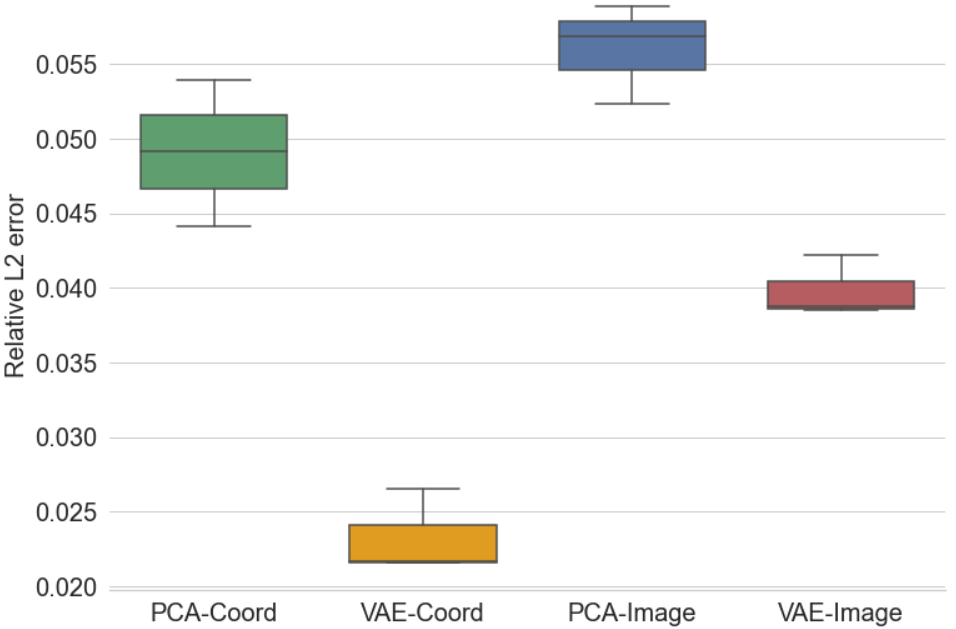}\label{ood_test1_2}}}
    % \quad
     \subfloat[Test 2]{{\includegraphics[width=5.5cm]{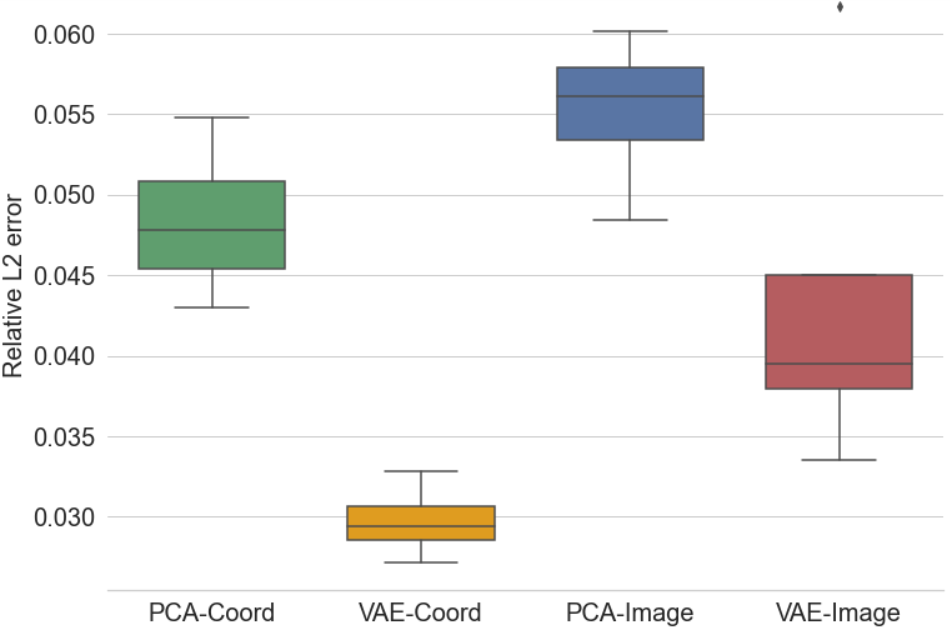}\label{ood_test2_2}}}%%
    \caption{\textit{Group 2 as OOD set: Performance on training and testing data sets.}}%
    \label{ood_2}%
\end{figure}

\begin{figure}[H]
    \centering
    \subfloat[Train set]{{\includegraphics[width=5.5cm]{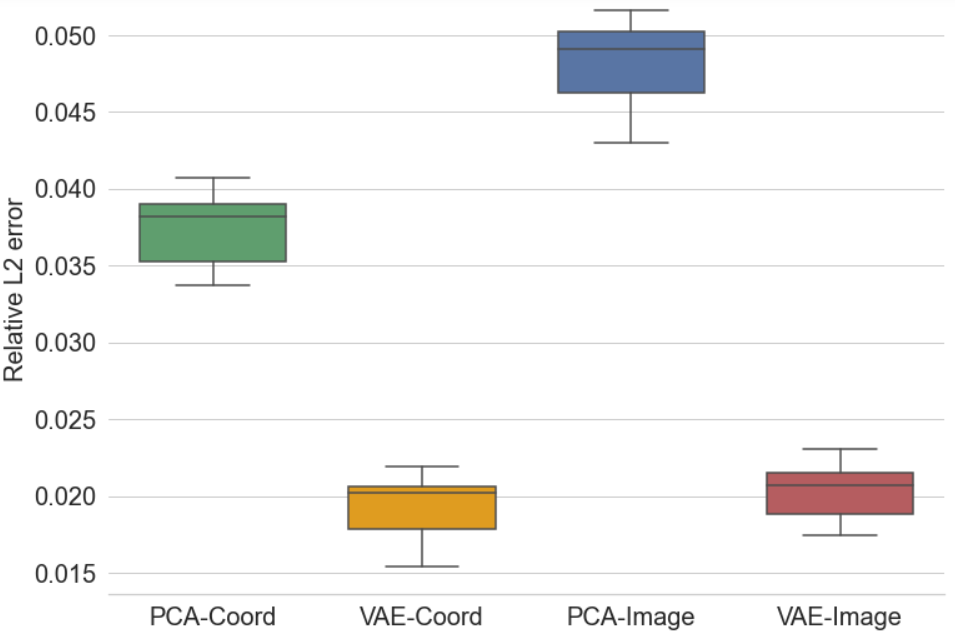} \label{ood_train_3}}}%%
    % \quad
    \subfloat[Test 1]{{\includegraphics[width=5.5cm]{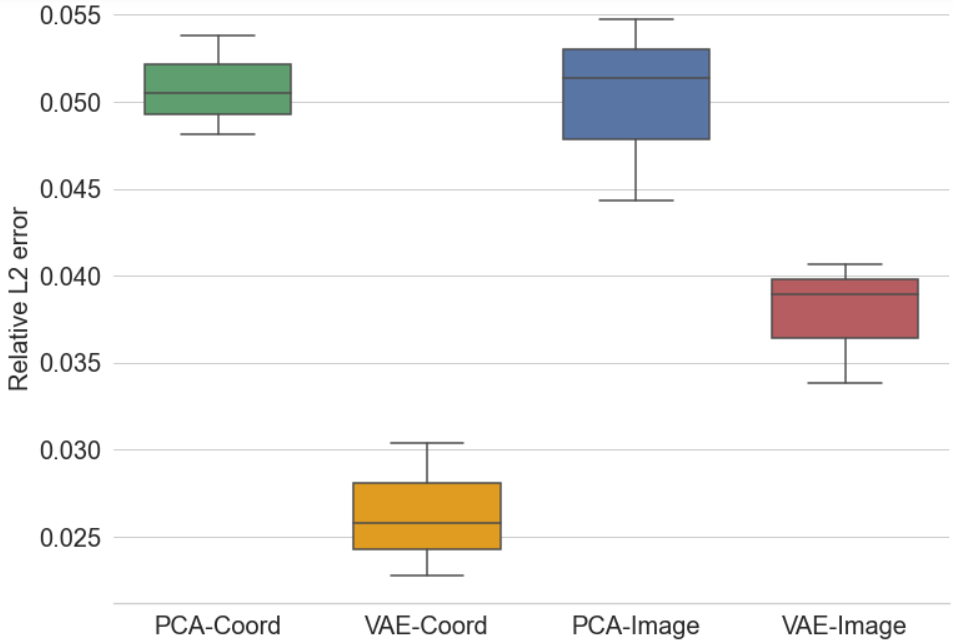}\label{ood_test1_3}}}
    % \quad
     \subfloat[Test 2]{{\includegraphics[width=5.5cm]{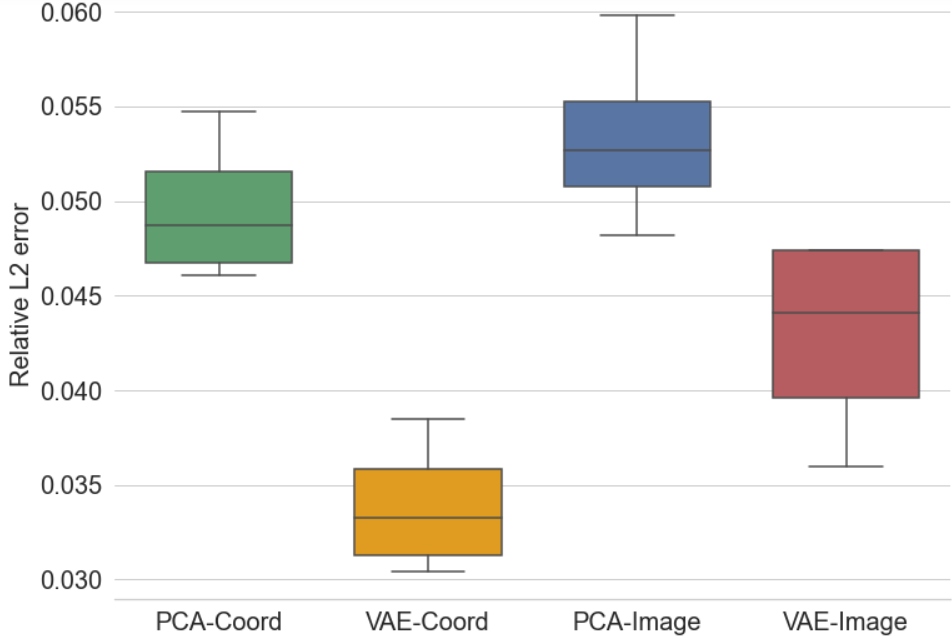}\label{ood_test2_3}}}%%
    \caption{\textit{Group 3 as OOD set: Performance on training and testing data sets.}}%
    \label{ood_3}%
\end{figure}

\begin{figure}[H]
    \centering
    \subfloat[Train set]{{\includegraphics[width=5.5cm]{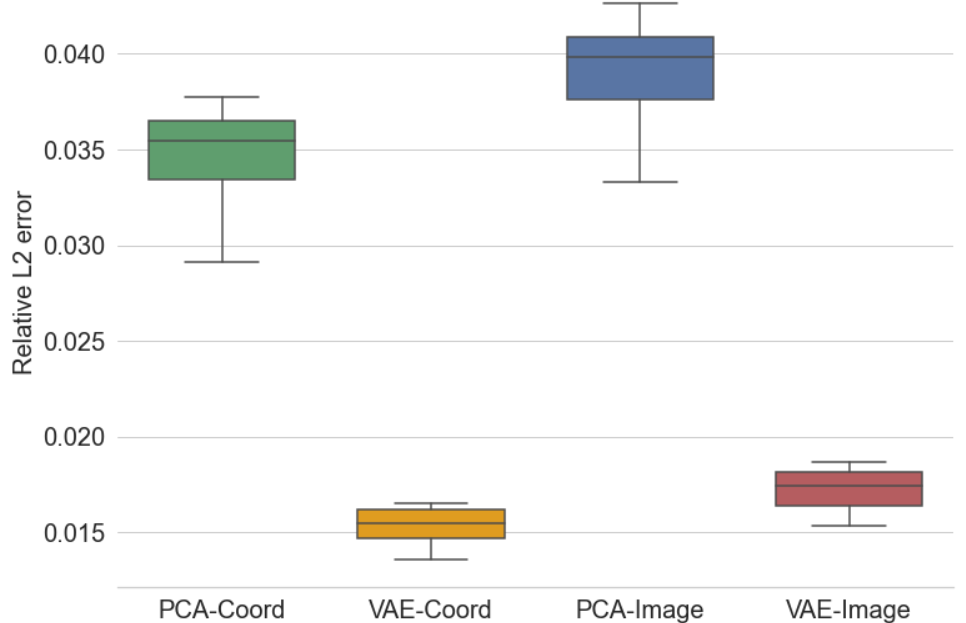} \label{ood_train_4}}}%%
    % \quad
    \subfloat[Test 1]{{\includegraphics[width=5.5cm]{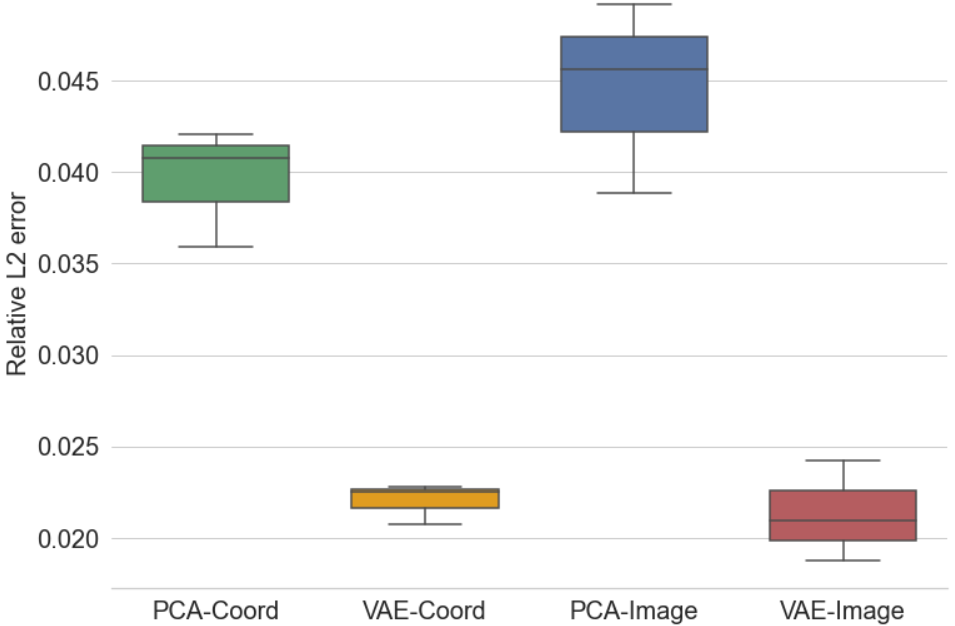}\label{ood_test1_4}}}
    % \quad
     \subfloat[Test 2]{{\includegraphics[width=5.5cm]{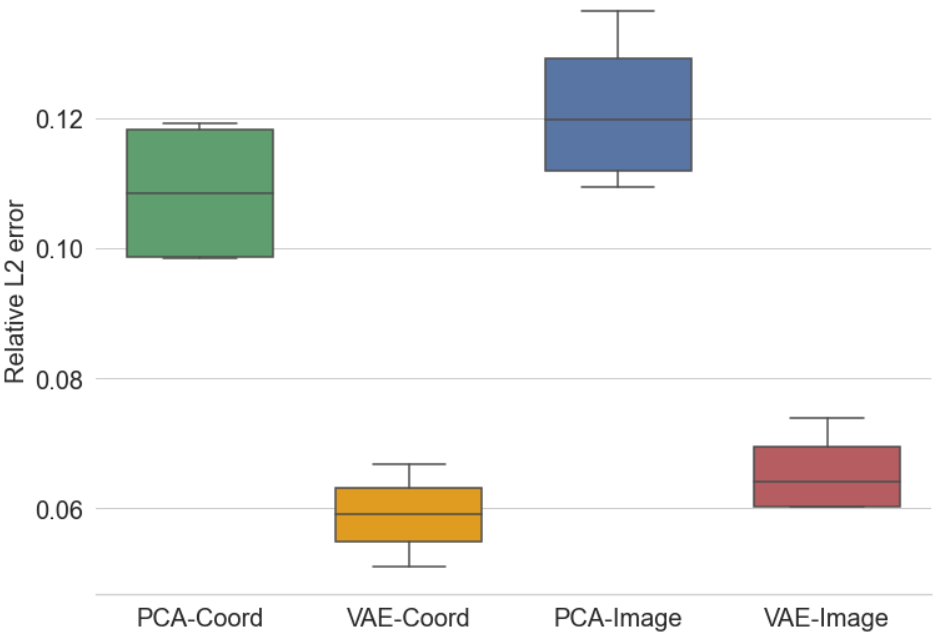}\label{ood_test2_4}}}%%
    \caption{\textit{Group 4 as OOD set: Performance on training and testing data sets.}}%
    \label{ood_4}%
\end{figure}

Figures \ref{ood_1}, \ref{ood_2}, \ref{ood_3}, \ref{ood_4} represent the box plots of relative errors between the prediction of GADEM and the reference solution when taking the group 1, 2, 3, 4 as the OOD testing set, respectively. We see that, with all cases, we obtain the same conclusion, i.e., the GADEM approaches with VAE always perform better than GADEM approaches with PCA. And using the spatial coordinates provides better accuracy than using images.

\bibliographystyle{plainnat} %unsrtnat
\bibliography{main} 

\end{document}